\documentclass[smallextended]{svjour3}       

\smartqed  

 \usepackage[utf8]{inputenc} 

\usepackage{mathptmx}       
\usepackage{helvet}         
\usepackage{courier}        
\usepackage{type1cm}        
\usepackage[ruled,vlined,linesnumbered]{algorithm2e}

\usepackage{makeidx}         
\usepackage{graphicx}        
\usepackage{multicol}        
\usepackage[bottom]{footmisc}
\usepackage{adjustbox}
\usepackage{array,multirow,makecell}
\setcellgapes{1pt}
\usepackage{epsfig}
\usepackage{epstopdf}
\usepackage{hyperref}
\hypersetup{colorlinks=true, pdfstartview=FitV, linkcolor=blue, citecolor=black, plainpages=false, pdfpagelabels=true, urlcolor=blue}
\usepackage{caption}
\usepackage{amssymb}
\usepackage{amsmath}
\usepackage{todonotes}
\usepackage{booktabs}  
\usepackage{multirow}  
\usepackage{float}
\usepackage{subcaption}
\usepackage{mwe}
\usepackage{hhline}

\begin{document}

\title{Movienet: A Movie Multilayer Network Model using Visual and Textual Semantic Cues}

\author{Youssef Mourchid         \and
Benjamin Renoust         \and
Olivier Roupin        \and
Lê Văn         \and
        Hocine Cherifi        \and
        Mohammed El Hassouni 
}


\institute{Youssef Mourchid \at
              LRIT - CNRST URAC 29, Rabat IT Center, Faculty of Sciences, Mohammed V University in Rabat, Morocco\\
              \email{youssefmour@gmail.com}  
           \and
Benjamin Renoust \at
              Institute for Datability Science, Osaka University, Osaka, Japan
           \and 
           Olivier Roupin \at
              Institute for Datability Science, Osaka University, Osaka, Japan
              \and
           Lê Văn             \at
              Institute for Datability Science, Osaka University, Osaka, Japan
           \and   Hocine Cherifi \at 
              LE2I UMR 6306 CNRS, University of Burgundy, Dijon, France.
              \and       
           Mohammed El Hassouni \at
              LRIT - CNRST URAC 29, Rabat IT Center, FLSH, Mohammed V University in Rabat, Morocco. LRIT - CNRST URAC 29, Rabat IT Center, Faculty of Sciences, Mohammed V University in Rabat, Morocco.\\                     
}

\date{Received: date / Accepted: date}

\maketitle

\begin{abstract} 

Discovering content and stories in movies is one of the most important concepts in multimedia content research studies. Network models have proven to be an efficient choice for this purpose. When an audience watches a movie, they usually compare the characters and the relationships between them. For this reason, most of the models developed so far are based on social networks analysis. They focus essentially on the characters at play. By analyzing characters interactions, we can obtain a broad picture of the narration’s content. Other works have proposed to exploit semantic elements such as scenes, dialogues, \emph{etc.}. However, they are always captured from a single facet. Motivated by these limitations, we introduce in this work a multilayer network model to capture the narration of a movie based on its script, its subtitles, and the movie content. After introducing the model and the extraction process from the raw data, we perform a comparative analysis of the whole 6-movie cycle of the Star Wars saga. Results demonstrate the effectiveness of the proposed framework for video content representation and analysis.
\end{abstract}

\section{Introduction}
\label{sec:introduction}

Since ancient times, humans have been telling stories, putting on scene different characters in their own rich world. Each story forms a small universe, sometimes intertwining with one another. The creation of a story is a careful recipe that brings together characters, location, and other elements so that it catches a reader, a viewer, or a listener's full attention. To collect these stories, books present and structure these elements such that any reader would assemble them in their mind, building their own vision of the story.

Movies follow the same narrative principles, but stimulate viewers differently by providing a fully constructed visual world that is the product of movie director's and its team's vision. Viewers' perception can be manipulated, motivating in them the elicitation of different emotions, and their progression into some unknown universe, such as it is done is science-fiction movies. The articulation of the story elements can be the hallmarks of a director's fingerprint, characterizing genre and stories or even movie rating prediction.

Network modelling puts into relation different entities, therefore it has naturally become a powerful tool to capture the elements articulation in stories~\cite{rital2005weighted,park2012social,waumans2015topology,tan2014character,renoust2015social,renoust2016visual,mish2016game,mourchid_multilayer18,viard2018movie,markovivc2018applying}. Such network models have been applied to many different types of stories, starting with written stories in books \cite{waumans2015topology,markovivc2018applying}, in news events from news papers and TV~\cite{renoust2015social}, in television series~\cite{tan2014character}, and eventually in the target medium of this paper: movies~\cite{park2012social,mourchid_multilayer18}. The topology and structure of these networks have been investigated both visually~\cite{renoust2015social,renoust2016visual} and analytically~\cite{waumans2015topology,rital2005weighted}, and may in turn be used for prediction tasks~\cite{viard2018movie}. These narrative networks built from large scale archives can be automatically created~\cite{waumans2015topology,renoust2015social,renoust2016visual} or use manual annotations~\cite{mish2016game}.

Social network analysis is one main focus of video network analysis, so naturally most of the related works put into relation characters at play in a story. But this only reveals one part of the story. In order to investigate an event, journalists use the 5 W-questions~\cite{chen2009novel,kipling1998just,kurzhals2016visual} (which are \emph{Who?}, \emph{What?}, \emph{When?}, \emph{Where?} and \emph{How/Why?}). Answering the most complex question \emph{How/Why?} is the whole focus of analytics at large, often done through the articulation of the other four questions. Social network analysis then mostly focuses on \textit{Who?} and puts it in perspective with other questions such as time (\textit{When?}) for dynamic social networks~\cite{sekara2016fundamental}, or with semantics (\textit{What?}) in content analysis~\cite{park2012social,renoust2014entanglement}, location (\textit{Where?}) with additional sensor networks~\cite{bao2015recommendations}, and even the multiple combinations of those (\textit{i.e.} streamgraphs)~\cite{latapy2018stream,viard2018movie}. Our goal is to provide a more holistic analysis over the different story elements by using a multilayer network modeling.

The recommended process of movie creation starts with the writing of the script, which is a text that is usually structured. A movie script assembles all movie elements in a temporal fashion (scenes, dialogues) and highlights specific information such as characters and setting details, so that it supports automatic movie analysis~\cite{jhala2008exploiting,mourchid_multilayer18}.
In recent years, image analysis tools have tremendously enhanced our automatic understanding of image content~\cite{guo2016deep}, and although tasks such as picture localization remain challenging ~\cite{demirkesen2008comparison,pastrana2006predicting}, we may enrich textual approaches with face detection and recognition~\cite{jiang2017face,cao2018vggface2} or with scene description~\cite{johnson2016densecap,yang2017dense}.

In our previous work~\cite{mourchid_multilayer18}, we introduced a network analysis that deploys across \textit{Who?}, \textit{What?} and \textit{Where?} extracted from the textual cues contained in the script, articulated around \textit{When?} as the script unfolds. We capture these by proposing a multilayer network model that describes the structure of a movie in a richer way as compared  to regular networks. It enriches the single character network analysis, and allows to use new topological analysis tools~\cite{domenico2014multilayer}.

In this paper, we extend this approach into multiple direction.
\begin{itemize}
    \item We extend the original model based only on  the script information in order to exploit the multimedia nature of information. It integrates, now, information contained in the movie (through shot segmentation, dense captioning, and face analysis) and in the subtitles. 
    \item We additionally root the model on the multilayer network formalism proposed by Kivel\"{a}~\cite{kivela2014multilayer}, to articulate characters, places, and themes across modalities (text and image).
    \item From single movies, we extend our model analysis to the first six movies of the Star Wars saga.   
\end{itemize}

After discussing the related work in the next section, we introduce the proposed model called Movienet in Section~\ref{sec:model}. We describe how we extract the multilayer network in Section~\ref{sec:processing}, before deploying the analysis in Section~\ref{sec:usecases} on the Star Wars saga~\cite{starwars1977episode,starwars1980episode,starwars1983episode,starwars1999episode,starwars2002episode,starwars2005episode}.
We finally conclude in Section~\ref{sec:conclusion}.

\section{Related work}
\label{sec:relatedwork}

 Network-based analysis of stories is widely spread, first for topical analysis \cite{kadushin2012understanding,renoust2014entanglement}. But when applied to multimedia data and movies, the analysis first focused on scene graphs~\cite{yeung1996extracting,jung2004narrative,correa2019semantic} for their potential for summarization. Character networks then became a natural focus for story analysis which from literature~\cite{knuth1993stanford,waumans2015topology,chen2019unsupervised} expanded to multimedia content~\cite{weng2009rolenet,tan2014character,tran2015cocharnet,renoust2015social,mish2016game,he2018srn}. Particular attention has been paid to dialogue structure~\cite{park2012social,gorinski2018s}, which leads to an extension of network modeling to multilayer models~\cite{lv2018storyrolenet,ren2018generating,mourchid_multilayer18}.

\textbf{Scene graphs: } Some studies have proposed graphs based on scenes segmentation and scenes detection methods to analyze movie stories. 
Yeung \emph{et al.}~\cite{yeung1996extracting} proposed an analysis method using a graph of shot transitions for movie browsing and navigation, to extract the story units of scenes. 
Edilson \emph{et al.}~\cite{correa2019semantic} extends this approach by constructing a narrative structure to documents. They connect a network of sentences based on their semantic similarity, which can be employed to characterize and classify texts.
Jung \emph{et al.}~\cite{jung2004narrative} use a narrative structure graph of scenes for movie summarization, where scenes are connected by editorial relations.
Story elements such as major characters and their interactions cannot be retrieved from these networks. 
Our work contrasts in using additional sources (scripts, subtitles, etc).

\textbf{Character networks in stories:} Character network analysis is a traditional exercise of social network analysis, with the network from \textit{Les Mis\'{e}rables} now being a classic of the discipline~\cite{knuth1993stanford}, and still inspires current research. Waumans \emph{et al.}~\cite{waumans2015topology} create social networks from the dialogues of the \emph{Harry Potter} series, including sentiment analysis and generating multiple kind of networks, with the goal of defining a story signature based on the topological analysis of its networks. Chen \emph{et al.}~\cite{chen2019unsupervised} propose an integrated approach to investigating the social network of literary characters based on their activity patterns in the novel. They use the minimum span clustering (MSC) algorithm for the identification of the character network’s community structure, visualizing the community structure of the character networks, as well as to calculate centrality measures for individual characters.

\textbf{Co-appearance social networks:} Co-appearance networks, connecting when co-appearing characters on screen, have been an important subject of research, even reaching the characters of the popular series Game of Thrones~\cite{mish2016game}.
\textit{RoleNet}~\cite{weng2009rolenet} identifies automatic leading roles and corresponding communities in movies through a social network analysis approach to analyze movie stories. 
He \emph{et al.}~\cite{he2018srn} extend co-appearance network construction with a spatio-temporal notion. They analyze social centrality and community structure of the network based on human-based ground truth. 
Tan \emph{et al.}~\cite{tan2014character} analyze the topology of character networks in TV series based on their scene co-occurrence in scripts. 
\textit{CoCharNet}~\cite{tran2015cocharnet} uses manually annotated co-appearance social network on the six Star Wars movies, and propose a centrality analysis.  
Renoust \emph{et al.}~\cite{renoust2015social} propose an automatic political social network construction from face detection and tracking data in news broadcast. The network topology and importance of nodes (politicians) is then compared across different time windows to provide political insights. 
Our work is very inspired by these co-appearance social networks, which give an interesting insight for the roles of characters, but they are still insufficient to fully place the characters in a story, which is why we rely on additional semantic cues. 

\textbf{Dialogue-based social networks:} Social networks derived from dialogue interaction in movie scripts have been used for different purposes. \textit{Character-net}~\cite{park2012social} proposes a story-based movie analysis method via social network analysis using movie script. They construct a weighted network of characters from dialogue exchanges in order to rank their role importance. 
Based on a corpus of movie scripts, Gorinski \emph{et al.}~\cite{gorinski2018s} proposed an end-to-end machine learning model for movie overview generation, that uses graph-based features extracted from character-dialogue networks built from movie scripts.

Similar to co-appearance networks, these approaches only use a social network for video analysis based on dialogue interaction, which cannot provide a socio-semantic construct of the video narration content. Having a different purpose, the proposed model gives a \textit{W}-question based semantic overview of the movie story, tapping into the very multimedia nature of movies.

\textbf{Multilayer network approaches: }Recent approaches use multiplex networks to combine both  visual and textual semantic cues.
StoryRoleNet~\cite{lv2018storyrolenet} is not properly a multilayer approach, but it well displays the interest of multimodal combination. It provides an automatic character interaction network construction and story segmentation by combining both visual and subtitle features. 
In the \textit{Visual Clouds}~\cite{ren2018generating} networks extracted from TV news videos are used as a backbone support for interactive search refinement on heterogeneous data. However, layers  cannot be investigated individually. In a previous work~\cite{mourchid_multilayer18}, we introduced a multilayer model to describe the content of a movie based on the movie script content. Keywords, locations, and characters are extracted from the textual information to form the multilayer network. This paper builds on this work by further exploiting additional medium sources, such as subtitles and the image content of the video to enrich the model and to refine the multilayer extraction process. The proposed model is fully multimedia, as it takes into account text-based semantic extraction, and image-based semantic cues from face recognition and scenes captioning, in order to capture a richer structure for the movies.

\section{Modeling stories with Movienet}
\label{sec:model}

To describe a complete story, four fundamental questions are investigated (\emph{Who?}, \emph{Where?}, \emph{What?}, \emph{When?} often refered as the four \emph{Ws})~\cite{flint1917newspaper,kipling1998just}. Inferring \emph{How/Why?} can be done while articulating the other \emph{Ws} making them essential bricks of analysis:

Given our context of movie understanding, we may reformulate the four \emph{Ws} as follows: 

\begin{itemize}
    \item \emph{Who?} denotes \textbf{characters} and \textit{people appearing} in a movie; 
    \item \emph{Where?} denotes  \textbf{locations} where actions of a movie take a place; 
    \item \emph{What?} denotes \textbf{subjects} which the movie talks about and \textit{other elements that describes} a movie scene. 
    \item \emph{When?} denotes the \textbf{time} that guide the succession of events in the movie. 
\end{itemize}

Answering these questions form the entities \emph{characters} (mentioned in the script), \emph{locations} (as depicted by the script), \emph{keywords} (conversation subjects understood from dialogues), \emph{faces} (as people \textit{appear} on screen), and \emph{captions} (that describe a scene) -- which ground our study. \emph{Time} is a special case to infer connections, but we do not treat it as an entity in our model.

Our goal is to help formulate movie understanding by articulating these four \emph{Ws}. In a preliminary work, we exploited the information contained in the movie script in order to construct a multilayer network. However, we neglected the complementary information contained in the movie and the subtitle. Using both visual and textual information allows a better understanding of the content and therefore a richer representation.
 
We propose a multilayer graph model that complete the previous model formulation~\cite{mourchid_multilayer18} by exploiting two additional layers, \emph{faces} and \emph{captions}. The multilayer graph puts these elements together as they form a story by exploiting two new sources that are subtitles and the video content. This model is made of five layers in order to represent each type of entity \textit{characters}, \textit{keywords}, \textit{locations}, \textit{faces}, and \textit{captions}, with multiple relationships between them.

\begin{figure}[htbp]
\centering
\includegraphics[width=0.9\linewidth]{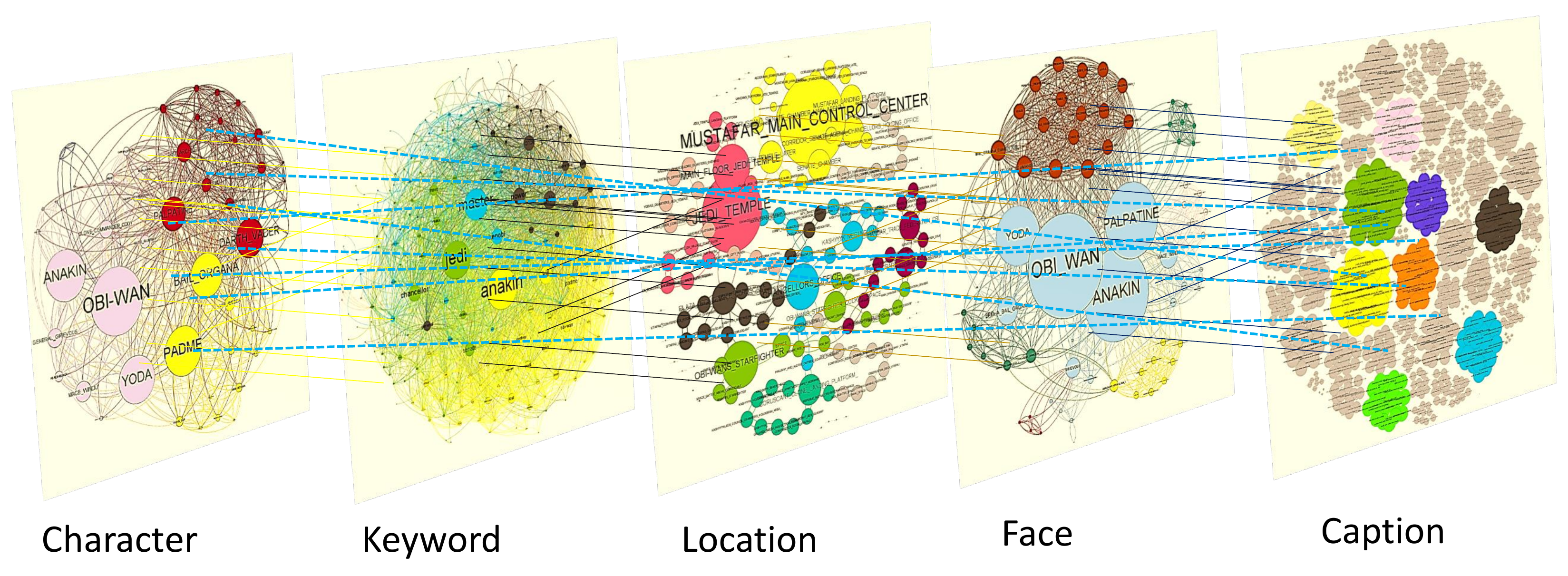}
\caption{A conceptual presentation of our multilayer network model: Five main  layers \newline of nodes, \emph{Character} $G_{CC}$, \emph{Keyword} $G_{KK}$, \emph{Location} $G_{LL}$, \emph{Face} $G_{FF}$ and \emph{Caption} $G_{CaCa}$ \newline are interacting  within and across each layer.}
\label{fig:model}
\end{figure}

Following Kivel\"{a}'s definition~\cite{kivela2014multilayer} of multilayer networks, we model two main classes of relationships: intra-layer relationships, between nodes of a same category, such as two faces appearing in the same scene; and inter-layer relationships which capture the interactions between nodes of different categories, such as when a caption describes a scene where a character is present. 
Altogether, the multiple families of nodes and edges form a multilayer graph as illustrated in Figure~\ref{fig:model}.

We now define our multilayer graph $\mathbb G = (\mathbb V, \mathbb E)$ such that:

\begin{itemize}
\item $V_C \subseteq \mathbb V$ represents the set of characters $c \in V_C$,
\item $V_L \subseteq \mathbb V$ represents the set of locations $l \in V_L$,
\item $V_K \subseteq \mathbb V$ represents the set of keywords $k \in V_K$.
\item $V_F \subseteq \mathbb V$ represents the set of faces $f \in V_F$.

\item $V_{Ca} \subseteq \mathbb V$ represents the set of captions $ca \in V_{Ca}$.
\end{itemize}

The different families of relationships can then be defined as:

\emph{Intra-layer:}

\begin{itemize}
\item $e \in E_{CC} \subseteq \mathbb E$ between two characters such that $e=(c_i, c_j) \in V_C^2$, when a character $c_i \in V_C$ is conversing with another character $c_j \in V_C$.

\item $e \in E_{LL} \subseteq \mathbb E$ between two locations such that $e=(l_i, l_j) \in V_L^2$, when there is a temporal transition from one location $l_i \in V_L$ to the other $l_j \in V_L$.

\item $e \in E_{KK} \subseteq \mathbb E$ between two keywords such that $e=(k_i, k_j) \in V_K^2$, when $k_i \in V_K$ and $k_j \in V_K$ belong to the same subject.
\item $e \in E_{FF} \subseteq \mathbb E$ between two faces such that $e=(f_i, f_j) \in V_F^2$, when $f_i \in V_F$ and $f_j \in V_F$ appear in the same scene.

\item $e \in E_{CaCa} \subseteq \mathbb E$ between two captions such that $e=(ca_i, ca_j) \in V_{Ca}^2$, when $ca_i \in V_{Ca}$ and $ca_j \in V_{Ca}$ describe the same scene.

\end{itemize}
\emph{Inter-layer:}

\begin{itemize}

\item $e \in E_{CK} \subseteq \mathbb E$ between a character and a keyword such that $e=(c_i, k_j) \in V_C \times V_K$, when the keyword $k_j \in V_K$ is pronounced by the character $c_i \in V_C$. 

\item $e \in E_{CL} \subseteq \mathbb E$ between a character and a location such that $e=(c_i, l_j)  \in V_C \times V_L$, when a character $c_i \in V_C$ is present in location $l_j \in V_L$.

\item $e \in E_{CF} \subseteq \mathbb E$ between a character and a face such that $e=(c_i, f_j)  \in V_C \times V_F$, when a character $c_i \in V_C$ appears in the same scene of $f_j \in V_F$.

\item $e \in E_{CCa} \subseteq \mathbb E$ between a character and a caption such that $e=(c_i, ca_j)  \in V_C \times V_{Ca}$, when a character $c_i \in V_C$ appears in the same scene which $ca_j \in V_{Ca}$ describes.

\item $e \in E_{KL} \subseteq \mathbb E$ between a keyword and a location such that $e=(k_i, l_j)  \in V_K \times V_L$, when a keyword $k_i \in V_K$ is mentioned in a conversation taking place in the location $l_j \in V_L$.

\item $e \in E_{KF} \subseteq \mathbb E$ between a keyword and a face such that $e=(k_i, f_j)  \in V_K \times V_F$, when a keyword $k_i \in V_K$ is mentioned in a scene where $f_j \in V_F$ appears.

\item $e \in E_{KCa} \subseteq \mathbb E$ between a keyword and a caption such that $e=(k_i, ca_j)  \in V_K \times V_{Ca}$, when a keyword $k_i \in V_K$ is mentioned in a scene which $ca_j \in V_{Ca}$ describes.

\item $e \in E_{LF} \subseteq \mathbb E$ between a location and a face such that $e=(l_i, f_j)  \in V_L \times V_F$, when a face $f_j \in V_F$ appears in the same scene which contains the location $l_i \in V_L$.

\item $e \in E_{LCa} \subseteq \mathbb E$ between a location and a caption such that $e=(l_i, ca_j)  \in V_L \times V_{Ca}$, when a caption $ca_j \in V_{Ca}$ describe a scene that contains the location $l_i \in V_L$.

\item $e \in E_{FCa} \subseteq \mathbb E$ between a face and a caption such that $e=(f_i, ca_j)  \in V_F \times V_{Ca}$, when a face $f_i \in V_F$ appears in the same scene that $ca_j \in V_{Ca}$ describes.
\end{itemize}

Edge direction and weight are not considered for the sake of simplicity. Moreover, as we do not intend to study the network dynamics, time is not directly taken into account. However, time supports everything: the existence of a node or an edge is defined upon time, unrolled by the order of movie scenes.



As a shortcut, we can now refer to subgraphs by only considering one layer of links and its induced subgraph: 
\begin{itemize}
\item $G_{CC} = (V_C, E_{CC}) \subseteq \mathbb G$ refers to the subgraph of character interaction;

\item $G_{KK} = (V_K, E_{KK}) \subseteq \mathbb G$ refers to the subgraph of keyword co-occurrence;

\item $G_{LL} = (V_L, E_{LL}) \subseteq \mathbb G$ refers to the subgraph of location transitions;

\item $G_{FF} = (V_F, E_{FF}) \subseteq \mathbb G$ refers to the subgraph of face interaction;

\item $G_{CaCa} = (V_{CaCa}, E_{CaCa}) \subseteq \mathbb G$ refers to the subgraph of caption co-occurrence;

\item $G_{CK} = (V_C \cup V_K, E_{CK}) \subseteq \mathbb G$ refers to the subgraph of characters speaking keywords;
\item $G_{CL} = (V_C \cup V_L, E_{CL}) \subseteq \mathbb G$ refers to the subgraph of characters standing at locations;

\item $G_{CF} = (V_C \cup V_F, E_{CF}) \subseteq \mathbb G$ refers to the subgraph of characters appearing with faces;

\item $G_{CCa} = (V_C \cup V_{Ca}, E_{CCa}) \subseteq \mathbb G$ refers to the subgraph of characters described by captions;

\item $G_{KL} = (V_K \cup V_L, E_{KL}) \subseteq \mathbb G$ refers to the subgraph of keywords mentioned at locations.

\item $G_{KF} = (V_K \cup V_F, E_{KF}) \subseteq \mathbb G$ refers to the subgraph of keywords said by faces.

\item $G_{KCa} = (V_K \cup V_{Ca}, E_{KCa}) \subseteq \mathbb G$ refers to the subgraph of keyword said at the same scene which caption describe.

\item $G_{LF} = (V_L \cup V_F, E_{LF}) \subseteq \mathbb G$ refers to the subgraph of faces appearing at locations.

\item $G_{LCa} = (V_L \cup V_{Ca}, E_{LCa}) \subseteq \mathbb G$ refers to the subgraph of captions describing locations.

\item $G_{FCa} = (V_F \cup V_{Ca}, E_{FCa}) \subseteq \mathbb G$ refers to the subgraph of captions describing faces.
\end{itemize}


Now that we have set the model, we need to extract elements from scripts, subtitles, and movie clips. This allows for the analysis of various topological properties of the network in order to gain a better understanding of the story.

\section{Extracting the multilayer network}
\label{sec:processing}

\begin{figure}[htbp]
\centering
\includegraphics[width=0.95\linewidth, height=6.5cm]{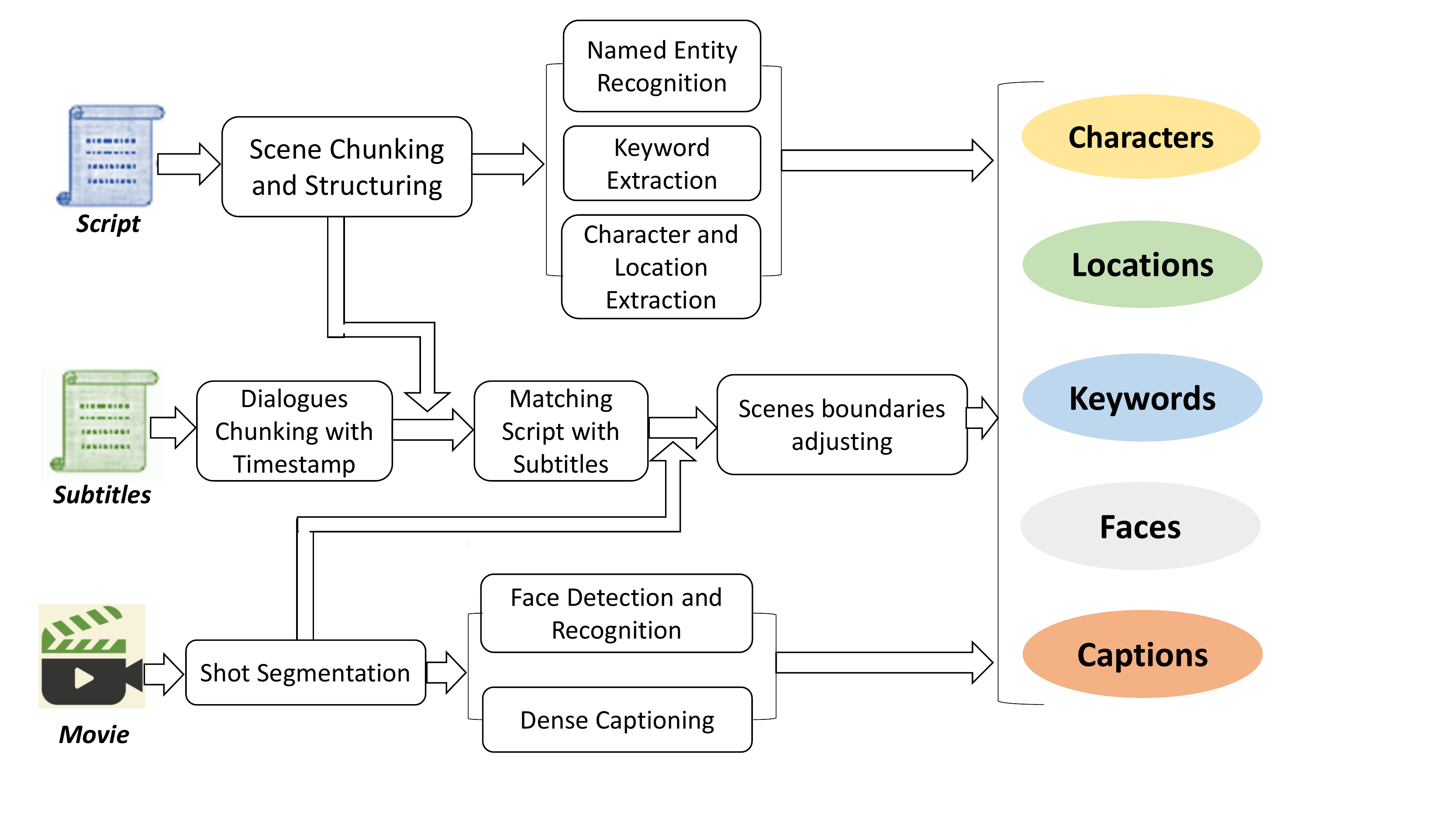}
\caption{Schematic view of the model construction process.}
\label{fig:methodology}
\end{figure}

We now describe the data and methodology used to build the multilayer network of a movie. Figure \ref{fig:methodology} illustrates the methodology processing pipeline. Very much inspired by the work from Kurzahls \textit{et al.} \cite{kurzhals2016visual}, we align scripts, subtitles and video, from which we extract different entities.
After introducing the extraction of the various entities and interactions from each data source, we explain how to build the network based on this information. 

\subsection{Data description}
Three data sources are used for this task: script, subtitles and video. 

\subsubsection{Definitions}

In order to remove any ambiguity, we first define the following dedicated glossary.

\begin{itemize}
\item Script: A text source of the movie which has descriptions about scenes, with setting and dialogues.
\item Scene: Chunk of a script, temporal unit of the movie. The collection of all scenes form the movie script.
\item Shots: Continuous (uncut) piece of video, a scene is composed of a series of shots.
\item Setting: The location a scene takes place in, and its description.
\item Character: Denotes a person/animal/creature who is present in a scene, often impersonated by an actor.
\item Dialogues: A collection of utterances, what all characters say during a scene.
\item Utterance: An uninterrupted block of a dialogue pronounced by one character.
\item Conversation: A continuous series of utterances between two characters.
\item Speaker: A character who pronounced an utterance.
\item Description: A script block which describes the setting.
\item Location: Where a scene takes place, or mentioned by a character.
\item Keyword: Most relevant information from an utterance, often representative of its topic.
\item Time: the time information extracted by aligning the script and subtitles.
\item Subtitles: a collection of blocks which have a time information.
\item Subtitles block: a block of the collection of utterance that has a start and end time.
\item Keyframe: a keyframe is a picture extracted from the movie. Keyframes are extracted at regular intervals (every second) to ease image processing.
\item Face: a character's face detected in a keyframe, associated to an image bounding box.
\item Caption: a descriptive sentence detected in a keyframe, associated to an image bounding box.

\end{itemize}

\subsubsection{Script}
\begin{figure}[htbp]
\centering
\includegraphics[width=0.7\linewidth]{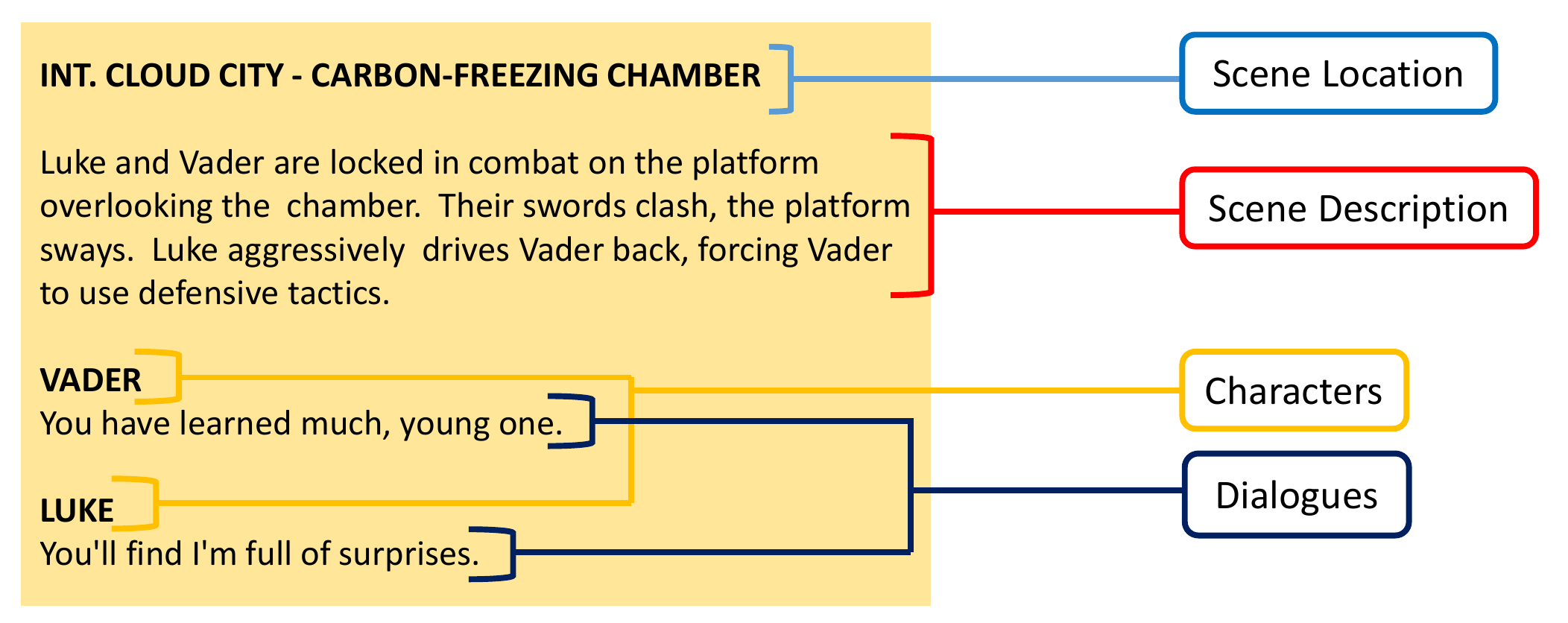}
\caption{Snippets of a resource script describing the movie \emph{The Empire Strikes  \newline Back}, displaying different elements manipulated (characters, dialogues and locations).}
\label{fig:script}
\end{figure}

Scripts happen to be very well-structured textual documents~\cite{jhala2008exploiting}. A script is composed of many scenes, each scene contains a location, scene description, characters and their dialogues. The actual content of a script often follows a semi-regular format \cite{jhala2008exploiting} such as depicted in Figure \ref{fig:script}. It usually starts with a heading describing the location and time of the scene. Specific keywords give important setting information (such as inside or outside scene) and character and key objects are often emphasized. The script then follows in a series of dialogues and setting descriptions.

\subsubsection{Subtitles}

\begin{figure}[t!]
\begin{center}
\includegraphics[width=9cm]{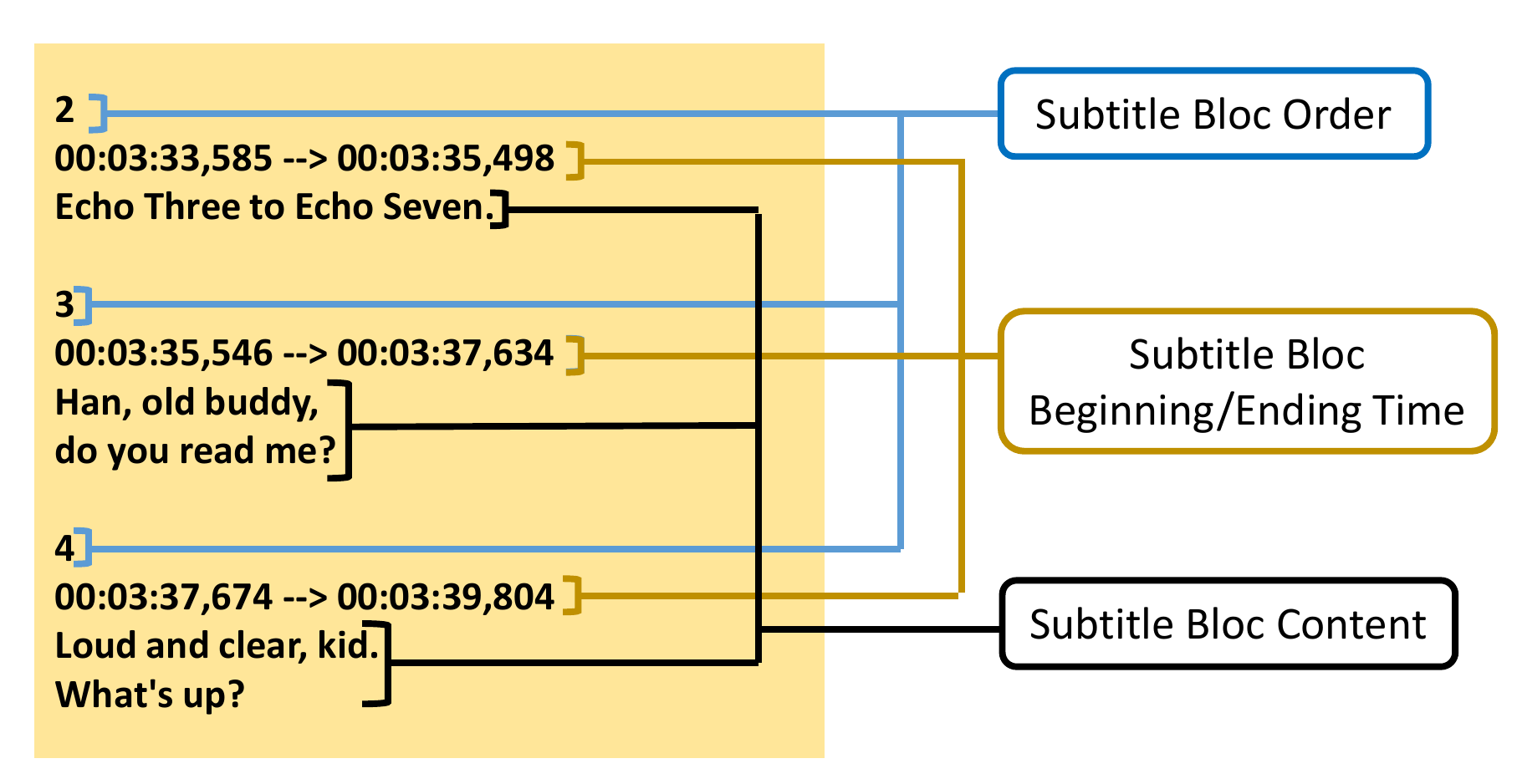}
\end{center}
\caption{Snippets of a resource file from \emph{The Empire Strikes Back} subtitles.}
\label{fig:sub}
\end{figure}

Subtitles are available in a SubRip Text (SRT) format and consist of four basic information (Figure \ref{fig:sub}): (1) a number to identify the order of the subtitles; (2) the beginning and ending time (hours, minutes, seconds, milliseconds) in which the subtitle should appear in the movie; (3) the subtitle text itself on one or more lines and (4) typically an empty line to indicate the end of the subtitle block. However, subtitles do not include information about characters, scenes, shots, and actions whereas dialogues in a script do not include time information.

\subsubsection{Video}

A movie's video can be divided into two components: a soundtrack (that we do not approach in this work) and a collection of images (the motion is then implied from the succession of these images). A movie is composed of scenes which are decomposed in shots. Scenes make up the actual unit of action which composes the movie. Each scene provides visual information about characters, locations, events, \emph{etc}.

\subsection{Script processing}

We now describe each step of the script processing pipeline. This process is language dependent, so we restrict our study to English scripts only. However, note that the framework can be easily adapted to other languages.

\subsubsection{Scene chunking and structuring}

As we mentioned above, scenes are the main subdivisions of a movie, and consequently our main unit of analysis. During a scene, all the critical elements of a movie  (all previously defined entities) interact. Each scene contains information about characters who talk, location where the scene takes place, and actions that occur. 
Our first goal is then to identify those scenes. 

Fortunately scripts are structured and give away this information. We then need to \textit{chunk} the script into scenes. In a script, a scene is composed as follows. First, there is a technical description line written in capital letters for each scene. It establishes the physical context of the action that follows. The rest of a scene is made of dialogue and description. 
 Each scene starts by a set information, \emph{INT} or \emph{EXT}, which indicates whether a scene takes place inside or outside, the name of the location, and also the time of day (\textit{e.g.} \emph{DAY} or \emph{NIGHT}). 

Within a scene heading description, important people and key objects are usually highlighted in capital letters that we may harvest while analyzing the text. 
Character names and their actions are always depicted before the actual dialogue lines. A line indent also helps to identify characters and dialogue parts in contrast to scene description. We can harvest scene locations and utterance speakers, by structuring each scene into its set of descriptions and dialogues. Finally, we identify conversations and characters present at a scene. Specific descriptions can then be associated to locations, and dialogues to characters. After chunking, we then obtain a scene structured into the following elements (as illustrated in Figure~\ref{fig:script}): a scene location, a description block, and a series of dialogues blocks assigned to characters.

\subsubsection{Semantic extraction}

The next step is to identify the actual text content that is attributed to locations or to speakers. Fortunately, Named Entity Recognition (NER)~\cite{nadeau2007survey} is a tool of natural language processing that labels significant words extracted from a text content with categories such as \textit{organizations}, \textit{people}, \textit{locations}, \textit{cities}, \textit{quantities}, \textit{ordinals}, \emph{etc.}
We apply NER to each scene description block and discard the irrelevant categories. However, this process is not perfect and many words can end up mislabelled due to the ambiguous context of the movie, especially within the science-fiction genre. In a second pass, we manually curate the resulting list of words and assign them to our fundamental categories: \textit{characters}, \textit{locations}, and \textit{keywords}.

Because ambiguity also includes polymorphism of semantic concepts, we next assign a unique class for synonyms referring to the same concept (\emph{i.e.} $\{LUKE, SKYWALKER\}\xrightarrow{}LUKE$). NER also helps us identifying characters present at a scene who are mentioned in utterances. Many public libraries are available for NER, and we used the spaCy library~\cite{al2017choosing} because of its efficiency in our context.

We may now identify \textit{keywords} within dialogues. We investigated three methods to measure the relevance of keywords: \textit{TF-IDF}~\cite{salton1975vector,li2007keyword}, LDA~\cite{blei2003latent} and Word2Vec~\cite{yuepeng2015keyword}. Because dialogue texts are made of short sentences (even shorter after stop-words removal), empirical results of Word2Vec and \textit{TF-IDF} rendered either too few words with a high semantic content, or too much words without semantic content. Only LDA, brought the best trade-off, but still included some level of noisy semantic-less words. We manually curated the resulting words by removing the remaining noise (such as \emph{can}, \emph{have}, and so on).

\subsection{Video processing}

Since video information also allows for answering a few of the \textit{W} questions, we introduce two techniques in this paper borrowed from computer vision: face detection and recognition to address \textit{Who}, and dense captioning to address \textit{What}. These are computationally intensive processes, so we first apply a rough shot detection using the PySceneDetect tool \cite{castellano2012pyscenedetect}, then
extract for each shot only one keyframe every second, which should maintain a good granularity to match with scenes. This renders an average of ${\sim}$8k key-frames per movie. Key-frames can then be analyzed in parallel.

\subsubsection{Face detection and recognition} 

Before knowing who appears in a scene, we need to detect if there is a face or not. This is the task of face detection applied in each frame. To extract those faces, we deployed a state-of-the-art face detector based on the faster R-CNN architecture~\cite{jiang2017face} that is trained with WIDER~\cite{yang2016wider}. 
This algorithm proposes bounding boxes for each detected face (in average obtaining ${\sim}$5k detected faces per movie). We then manually remove all false positive detections (around 6.5\% in average). 

We now need to identify who the faces belong to. We also wish to match the faces that belong to the same people. For each of the valid faces we use another state-of-the-art embedding technique, the ResNet50 architecture~\cite{he2016deep} trained on the VGGFace2 dataset~\cite{cao2018vggface2}. 
This allows us to obtain a 2048 dimensional vector that corresponds to each detected face. Traditional retrieval approaches are challenged because of the specific characteristics of our dataset (pairwise distances are very close within a shot \textit{and} very far between shots, in addition to other motion blur and lighting effects). Since the number of detected faces is limited for each movie, we only use automated approaches to assist manual annotation. We project the vector space in 2D using $t$-SNE~\cite{gisbrecht2015parametric} and manually extract obvious clusters within the visualization framework Tulip~\cite{auber2017tulip}. In order to quick-start the cluster creation, we applied a DBScan clustering~\cite{ester1996density}, for which we fine tuned parameters on our first manually annotated dataset, reaching a rough 17\% accuracy. Based on the detected clusters, and on the movie distribution, we then create face models as collections of pictures to incrementally help retrieving new pictures of the same characters. With the results still containing many errors, we finally manually curated them all to obtain a clean recognition for each character.

\subsubsection{Dense captioning} 

One could wish also to explore what objects and relations could be inferred from the scenes themselves. The dense captioning task~\cite{johnson2016densecap} attempts to use tools of computer vision and machine learning to describe textually the content of an image. We used an approach with inner joints~\cite{yang2017dense} trained with the Visual Genome \cite{Krishna2017}. This computes bounding boxes and sentences for each frame, accompanied with a confidence index $w \in [0, 1]$. 

Depending on the rhythm of the movie, frame extraction may still result in very similar consecutive frames.
As a consequence, dense captioning of these consecutive frames may be very similar. 
However, the similar captions may be assigned very different confidence index.
In order to extract the most relevant captions in this context, we propose to use this confidence index to rank then filter captions. 

We extend the \textit{TF-IDF} definition~\cite{salton1975vector} $tfidf=tf*idf$ to one incorporating caption confidence index. The notion of document here corresponds to a scene, and instead of a term, we have a caption. We define $tf(ca_i,s)$ the weighted frequency of caption $ca_i$ in a scene $s$ as follows:
$$
tf(ca_i,s) = \frac{\sum_{fr \in s}{w_{ca_i,fr}}}{\sum_{fr \in s}\sum_{ca \in f}{w_{ca,fr}}}
$$
where $ca$ denotes a caption having a confidence index $w_{ca,fr}$ in a frame $fr$ of a scene $s$. We then define $idf(ca_i,S)$ the inverse scene frequency such as:
$$
idf(ca_i,S) = log\left(\frac{|S|}{|\{s \in S:ca_i \in s\}|}\right)
$$
with $\{s \in S:ca_i \in s\}$ denoting the scenes $s$ which contain the caption $ca_i$ in the corpus made of all the scenes in the movie $S$.

We keep the top 40 captions per scene.
Captions are simple sentences, such as \textit{"a white truck parked on the street"}, and their generation process make them resemble a lot one another (due to the limitations of the training vocabulary and relationships). To further extract their semantic content, we compute their $n$-grams~\cite{cavnar1994n} ($n=4$, keeping a maximum of one stop word in the $n$-gram). 

Each resulting $n$-gram is then represented by a bag of unique words that we sort in order to cover permutations and help matching between scenes. The piece of sentences formed may then be used as an additional keyword layer obtained from the visual description of the scene,

\subsection{Time alignment between script and subtitles}

We now need to match the semantic information extracted from the script to the one extracted from the video. This can naturally be done by aligning the script with the time of the movie. The movie is played along time, but the script has no time information. Fortunately dialogues are reported in the script, and they correspond to people speaking in the movie. Subtitles are the written form of these dialogues, and they are time-coded in synchronization with the movie. The idea is to use them as a proxy to assign time-codes of matching dialogues in the script. Hence, we should have rough approximations of when scenes occur through dialogues start/end boundaries.

Unfortunately, the exact matching of scripts and dialogues greatly varies between versions of the script and movie. Sometimes a scene may appear in the script but not in the movie, and vice versa. Additionally, the order and wording may greatly differ between the two. 

To deal with these issues, we proceed in multiple steps as introduced by Kurzhals \textit{et al.}~\cite{kurzhals2016visual}. Scenes are decomposed in blocks, for which each is a character utterance. We then normalize the text on both sides through stemming. The idea is then to assign each of the utterance block to its corresponding counterpart in the subtitles. A first step checks for an absolute equality of subtitles and script dialogue. A second step is for textual inclusion between script and subtitles. This does not work for all utterances but the matching part gives search window constraints for our next step. For the remaining blocks, we compute their \textit{TF-IDF} weighted vectors~\cite{salton1975vector} and match with minimal cosine similarity. 

\begin{figure}[t!]
\begin{center}
\includegraphics[width=12.5cm,height=8.5cm]{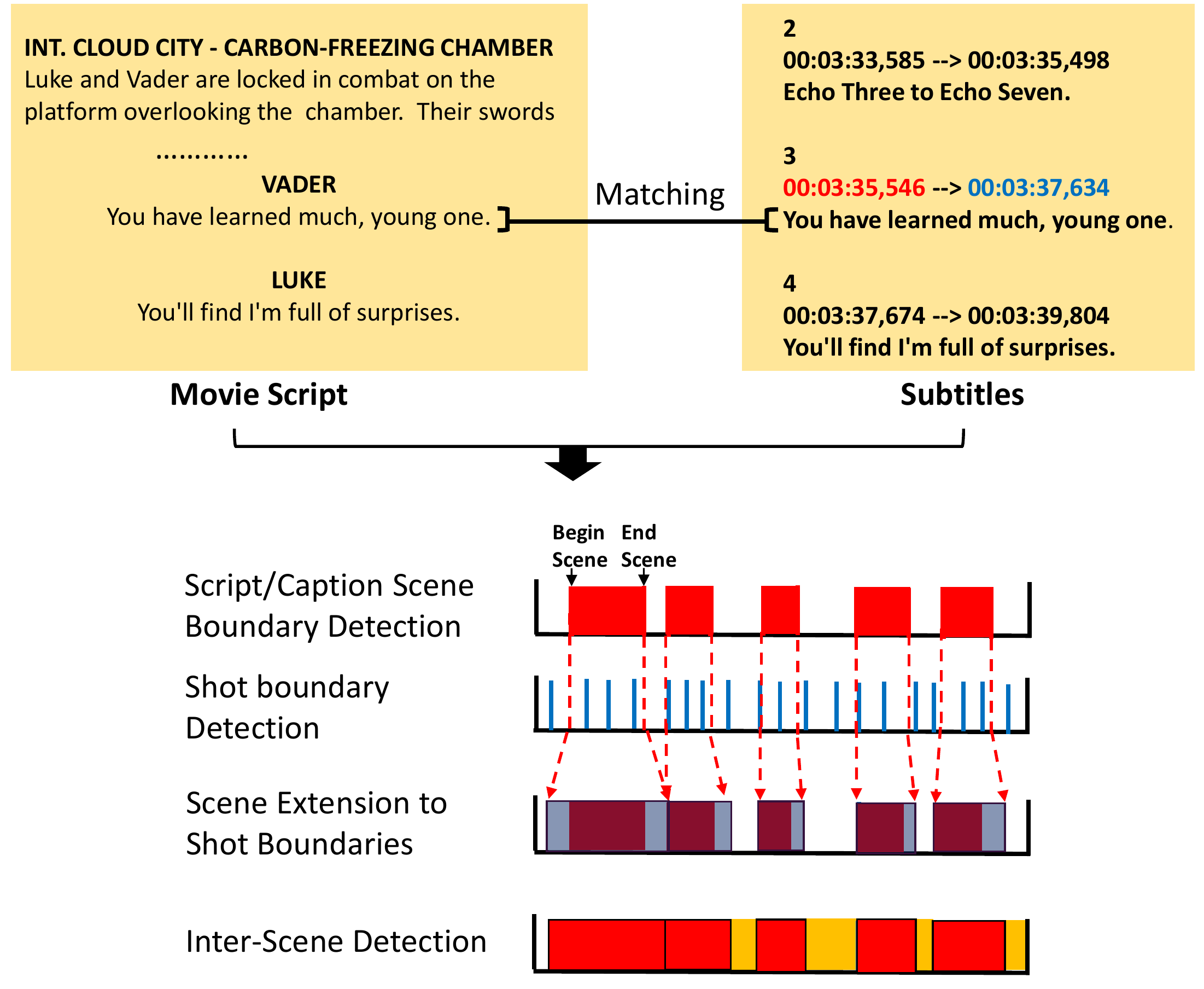}
\end{center}
\caption{Data fusion for script/subtitles alignment. First, the script is matched with \newline the subtitles. Then, we refine the scene boundary with the beginning and ending shot \newline boundaries.}
\label{fig:matching}
\end{figure}

Keywords and characters can then precisely be identified. But since a scene compiles a series of utterance, we get as a result a rough approximation of each scene's time boundaries, and each location too. To better align scenes and the video, we further refine the scene boundaries to those of the beginning and ending shot boundaries each scene is falling into, as shown in Figure \ref{fig:matching}.

Many scenes however do not contain any dialogue (a battle scene which contains only a description of what’s happening in it) and therefore cannot be matched to any subtitle block (these scenes are often used to better pace the narration, and may typically display an action from the outside, for example a moving vehicle). In other cases, scenes cannot be matched with subtitles when the dialogues are too small or have changed too much, and many scenes have actually been erased from script to the final movie cut. Table~\ref{tab:unmatched} summarizes these statistics.

\begin{table}[]

\begin{adjustbox}{max width=1.2\textwidth}
\begin{tabular}{c c c c c }
\toprule
Episode & \begin{tabular}{@{}c@{}}\# Script-caption \\ matching scenes\end{tabular}  & \begin{tabular}{@{}c@{}}\# Boundary based \\ retrieved scenes (\#empty)\end{tabular}   & \begin{tabular}{@{}c@{}}\# Meta scenes \\ (\#empty)\end{tabular}   & \begin{tabular}{@{}c@{}}\# Total scenes \\ (\#empty)\end{tabular} \\ \midrule
SW1     & 109                               & 23 (14)                                         & 51 (36)               & 183 (50)                \\
SW2     & 58                                & 22 (17)                                       & 70 (46)               & 150 (63)                \\
SW3     & 75                               & 16 (14)                                       & 97 (66)               & 188 (80)                \\
SW4     & 223                               & 66 (52)                                       & 193 (172)             & 479 (224)               \\
SW5     & 146                               & 52 (51)                                       & 77 (70)               & 275 (121)               \\
SW6     & 89                                & 27 (19)                                       & 22 (23)               & 138 (42)                \\ \bottomrule
\end{tabular}
\end{adjustbox}

\captionof{table}{Number of scenes, matched, retrieved, and missed from the script to caption, for each episode of the Star Wars saga as a pre-processing for use cases in Section~\ref{sec:usecases}. Note that, in the creation process, many scenes were actually removed and changed from their original version to the final cut, explaining the amount of mismatches (empty scenes are usually scene cuts giving a rhythm to the movie.}

\label{tab:unmatched}
\end{table}

The placement of some of these scenes may still be inferred from the matching of other scenes. Indeed, a scene that has not been matched can be fitted between its two neighboring scenes if they have been matched previously.
When more than one consecutive scenes cannot be matched, we create a meta scene to regroup them. For instance, if we have a gap of consecutive scenes between \emph{Scene 1 (00:02:00--00:02:20)} and \emph{Scene 5  (00:02:46--00:03:52)}, we create the \emph{Meta Scene 2--4 (00:02:20--00:02:46)} which starts from the end of \textit{Scene 1} and ends at the beginning of \textit{Scene 5}.

\subsection{Network construction}

As a result of the previous steps, we now have alignment between scenes, with location, characters, and keywords, and video frames, with faces, and descriptive captions. These form the entities to build the multilayer network made of the individual layers $V_L$, $V_C$, $V_K$, $V_F$, and $V_{Ca}$.

Let us revisit our investigative questions in the context of a scene: \emph{Where does a scene take place?} is identified by the \textit{locations}. \emph{Who is involved in a scene?} may be tackled by \textit{characters}, but also through the other question \emph{Who appears in a scene?} which is identified through \textit{faces}. \emph{What is a scene about?} is identified through keywords, but also partly by answering \emph{What is represented in a scene?}, tackled by captions.

We now wish to infer the relationships we described in Section \ref{sec:model}. 
Two characters $c_i$, $c_j$ can be connected when they participate in a same conversation, hence forming an edge $e_{c_i, c_j} \in E_{CC}$. We connect two locations $e_{l_i, l_j} \in E_{LL}$ when there is a temporal transition between the locations $l_i$ and $l_j$ (analogous to geographical proximity), \emph{i.e.} following the succession of two scenes. Keywords $k_i$, $k_j$ co-occurring in a same conversation create an edge $e_{k_i, k_j} \in E_{KK}$. If two faces $f_i$ and $f_j$ appear in the same scene, an edge  $e_{f_i, f_j} \in E_{FF}$. Two captions $ca_i$ and $ca_j$ describing the same scene can also be associated by an edge $e_{ca_i, ca_j} \in E_{CaCa}$.

Using the structure extracted from the script, subtitles, and movie content, we can add additional links between categories. An edge $e_{c_i,l_j} \in E_{CL}$ associates a character $c_i$ with a location $l_j$ when the character $c_i$ appears in a scene taking place at location $l_j$. 
When a character $c_i$ speaks an utterance in a conversation, for each keyword $k_j$ that is detected in this utterance, we create an edge $e_{c_i,k_j} \in E_{CK}$. If a character $c_i$ is present in the same scene as the face $f_j$ an edge $e_{c_i,f_j} \in E_{CF}$ is created between them. An edge $e_{c_i,ca_j}  \in E_{CCa}$ links a character $c_i$ with a caption $ca_j$ if the caption describes a scene in which the character appears. We can associate the keywords $k_i$ extracted in conversation placed in a location $l_j$ to form the edge $e_{k_i,l_j} \in E_{KL}$. We create an edge $e_{k_i,f_j} \in E_{KF}$ between a keyword $k_i$ and a face $f_j$ if the keyword is mentioned in a scene where the face is present. When a keyword $k_i$ is mentioned in a scene which the caption $ca_i$ describes, we create an edge $e_{k_i,ca_j} \in E_{KCa}$. A link $e_{l_i,f_j} \in E_{LF}$ is created between a location $l_i$ and a face $f_j$ when a location is in the scene where the face appears. We associate an edge $e_{l_i,ca_j} \in E_{LCa}$ between a location $l_i$ and a caption $ca_j$, if the location is in the scene that the caption describes. Finally, when a face $f_i$ appears in a scene that the caption $ca_j$ describes, an edge $e_{f_i,ca_j} \in E_{FCa}$ is created. A resulting graph combining all layers is visualized in Figure \ref{fig:model}.

\section{Network analysis}
\label{sec:usecases}
We now wish to perform a network analysis of the whole 6-movie Star Wars saga (hereafter SW). With many people to keep track of during the six movies, it can be a challenge to fully understand their dynamics. To demystify the saga, we turn to network science. After turning every episode of the saga into a multilayer network following the proposed model, our first task is to investigate their basic topological properties. We then further investigate node \emph{influence} as proposed by Boglio \textit{et al.}~\cite{bioglio2017movie}, on centralities that are defined for single-layer and multilayer cases: the \textit{Influence Score} is computed by the average ranking of three centralities. 

The three centrality measures we consider are defined for both single and multilayer cases~\cite{domenico2013centrality,notre2}. Additionally \emph{Degree}, \emph{Betweenness} and \emph{Eigenvector} centrality are among the most influential measures. \emph{Degree} centrality measures the direct interactions of a story element. The \emph{Betweenness} centrality measures how core to the plot a story element might be. The \emph{Eigenvector} centrality then measures the relative influence of a story element in relation to other influential elements. As a result, after studying influence score on separated layers, we then study it on our multilayer graphs.

\subsection{Description of the data}
First, a quick introduction to the SW saga: The saga began with \textit{Episode IV – A New Hope (1977)}~\cite{starwars1977episode}, which was followed by two sequels, \textit{Episode V – The Empire Strikes Back (1980)}~\cite{starwars1980episode} and \textit{Episode VI – Return of the Jedi (1983)}~\cite{starwars1983episode}, often referred to as \textit{the original trilogy}. Then, the prequel trilogy came, composed of  \textit{Episode I – The Phantom Menace (1999)}~\cite{starwars1999episode}, \textit{Episode II – Attack of the Clones (2002)}~\cite{starwars2002episode}, and \textit{Episode III – Revenge of the Sith (2005)}~\cite{starwars2005episode}. Movies and subtitles are extracted from DVD copies, and scripts can be acquired from the Internet Movie Script Database~\cite{imsdb2019} and Simply Scripts~\cite{simply2019} depending on the format.

The SW saga tells the story of a young boy (Anakin), destined to change the fate of the galaxy, who is rescued from slavery and trained by the Jedi (the light side), and groomed by the Sith (the dark side). He falls in love and marries a royalty, who fell pregnant. The death of his mother pushes him to seek revenge, so he gets coerced by the Sith. He is nearly killed by his former friend, but is saved by the Sith Emperor to ultimately stay by his side. His twin children are taken and hidden away, they grow up independently, one becomes a princess (Leia) and the other one becomes a farm hand (Luke). Luke stumbles upon a message from a princess in distress and seeks out an old Jedi who, knowing Luke's heritage, begins training him. To rescue the princess, they hire a mercenary (Han Solo) and save her. She turns out to be Luke's long lost twin sister. Discovering the identity of Luke, the emperor tries, with the help of Anakin, to turn him to the dark side. When that fails, he attempts to execute him, but Anakin, at the sight of his son's suffering, turns against the emperor saving the galaxy.

\subsection{Topological properties of individual layers}
\label{sec:topology}

\begin{figure}[t!]
        \centering
        \begin{subfigure}[b]{0.45\textwidth}
            \centering
            \includegraphics[width=\textwidth]{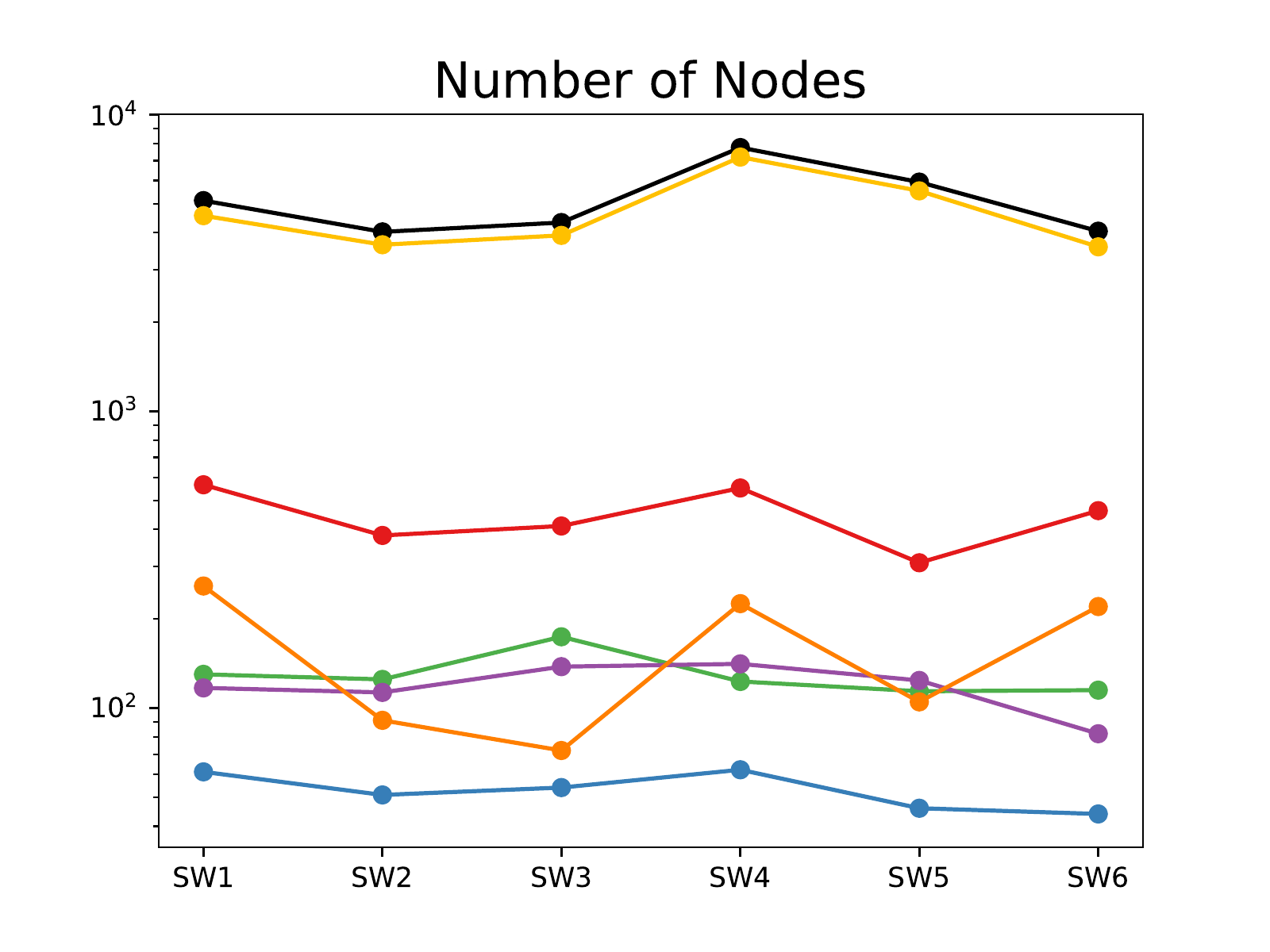}
             
             \label{fig:nodes}
        \end{subfigure}
        \begin{subfigure}[b]{0.45\textwidth}  
            \centering 
            \includegraphics[width=\textwidth]{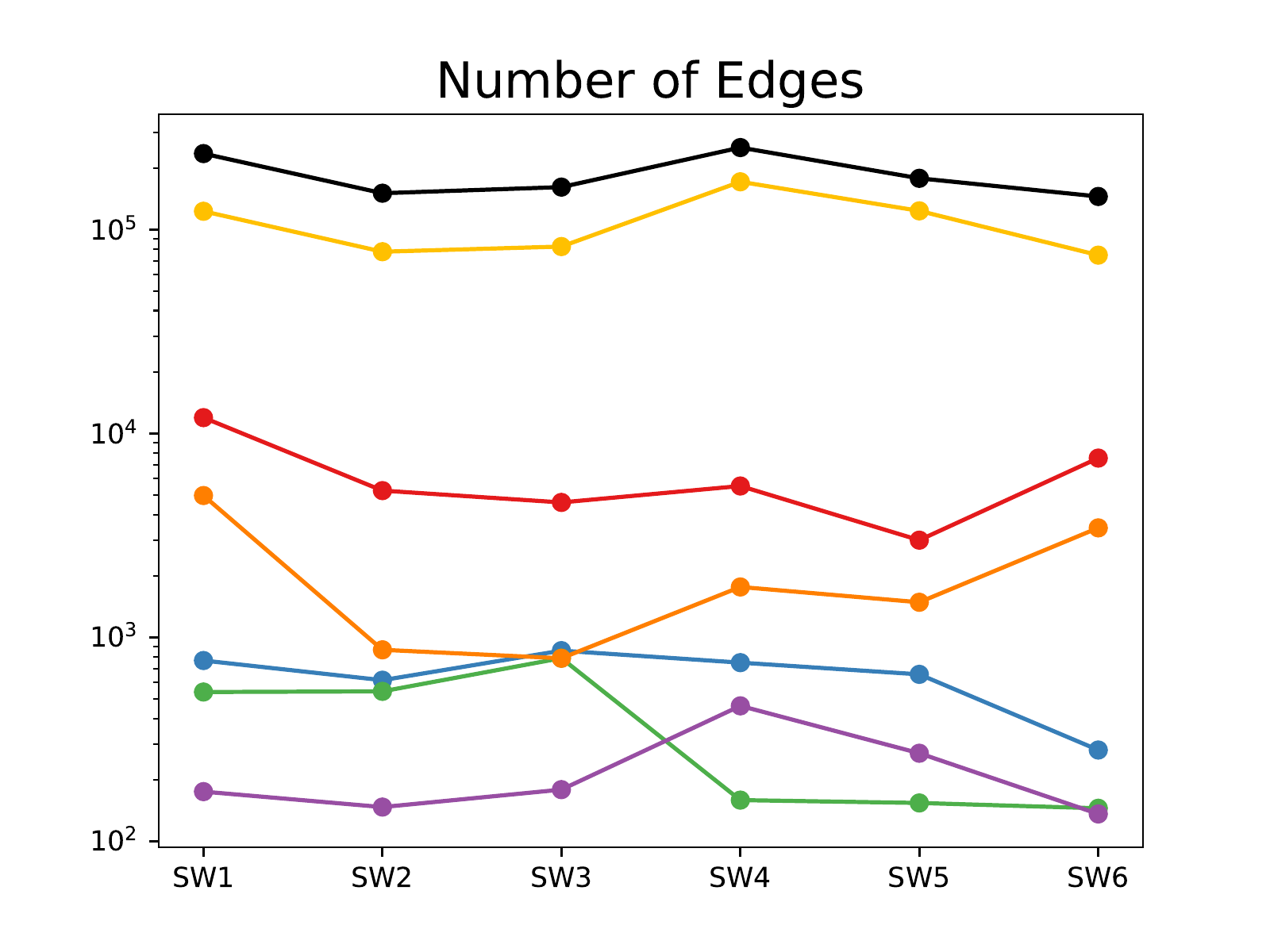}
       
        \label{fig:edges}
        \end{subfigure}
        \begin{subfigure}[b]{0.45\textwidth}  
            \centering 
            \includegraphics[width=\textwidth]{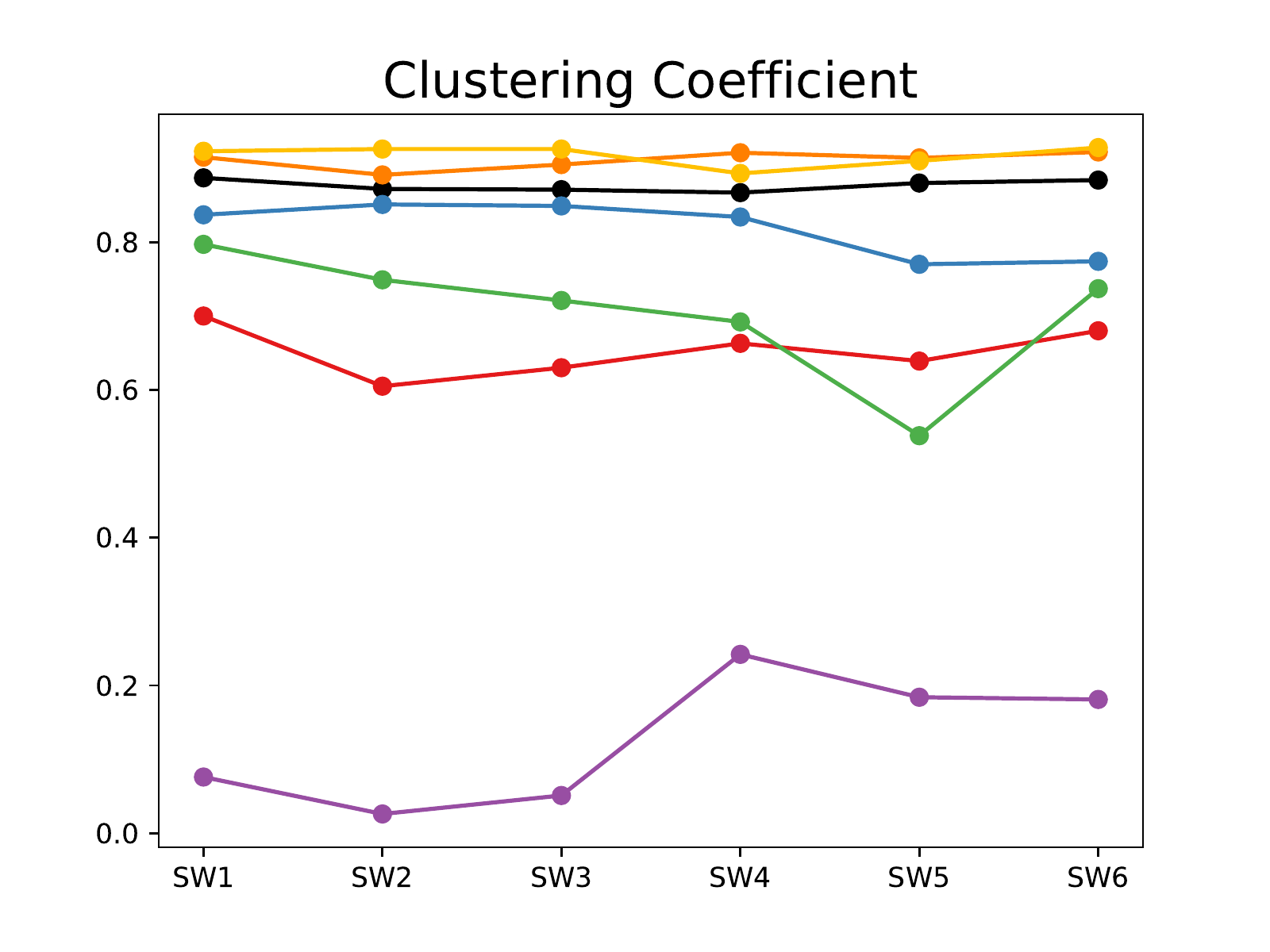}

      \label{fig:coefficient}
        \end{subfigure}
        \begin{subfigure}[b]{0.45\textwidth}   
            \centering 
            \includegraphics[width=\textwidth]{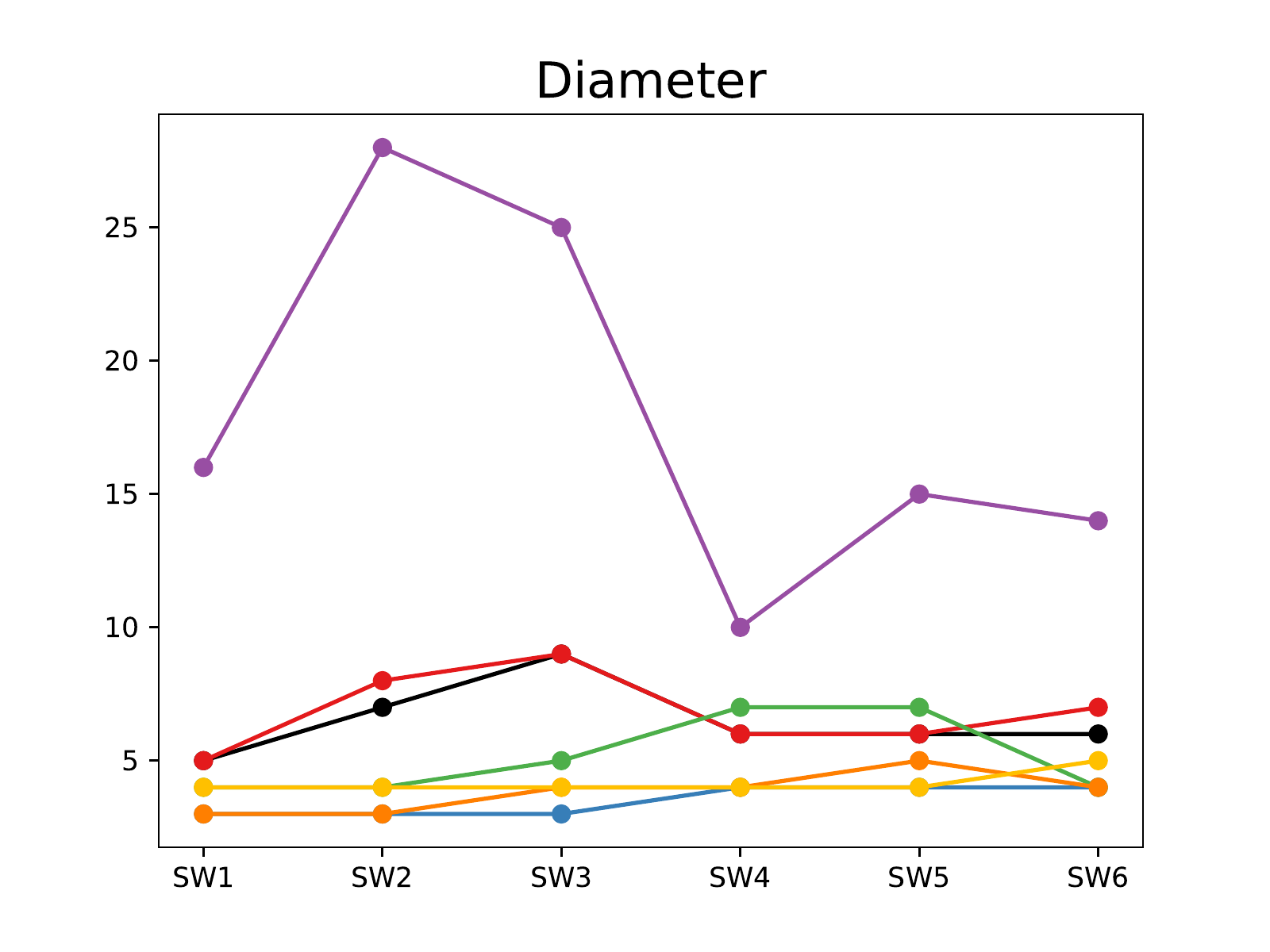}

        \label{fig:diameter}
        \end{subfigure}
        \begin{subfigure}[b]{0.45\textwidth}   
            \centering 
            \includegraphics[width=\textwidth]{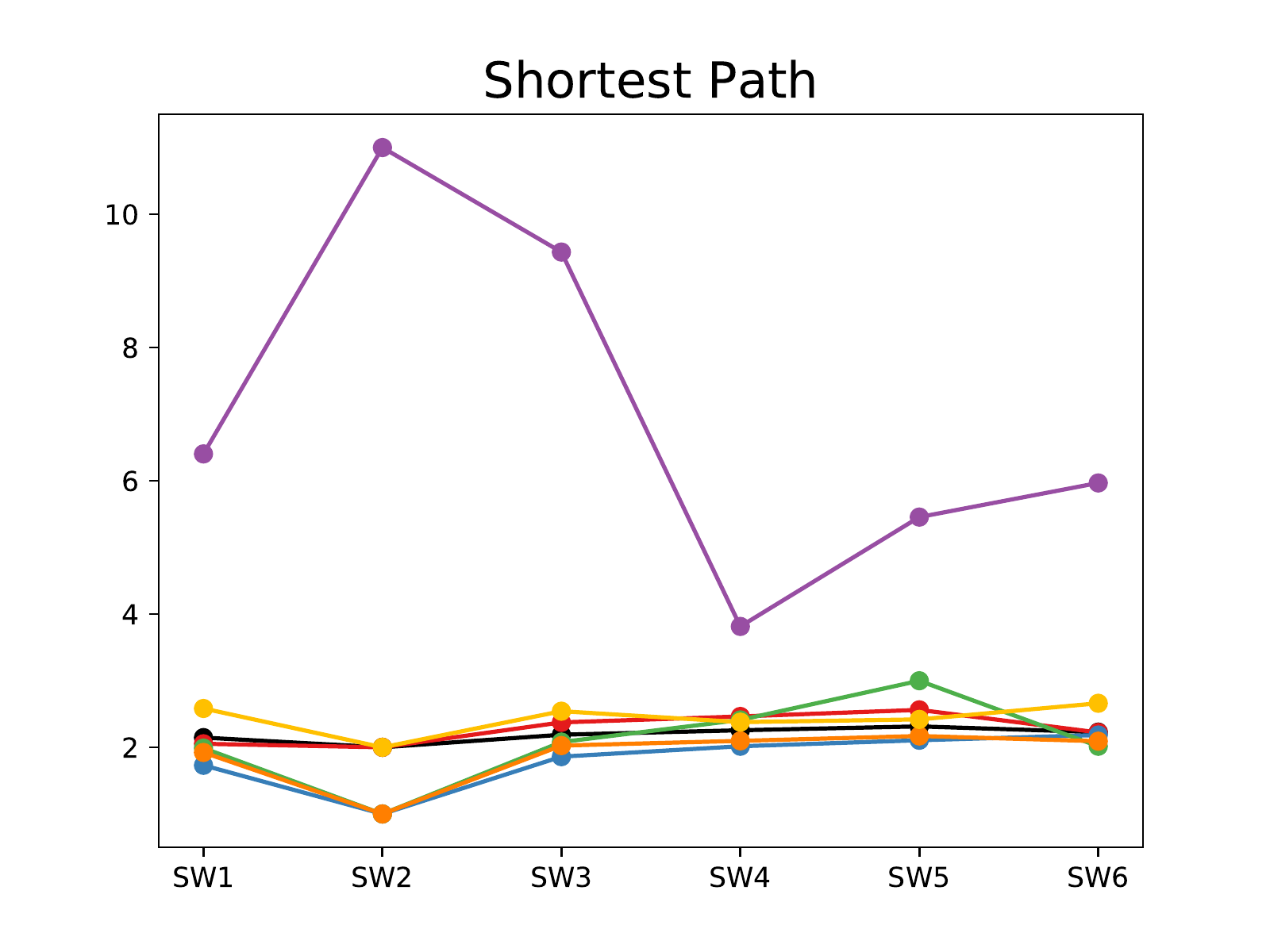}

        \label{fig:shortest}
        \end{subfigure}
        \begin{subfigure}[b]{0.45\textwidth}   
            \centering 
            \includegraphics[width=\textwidth]{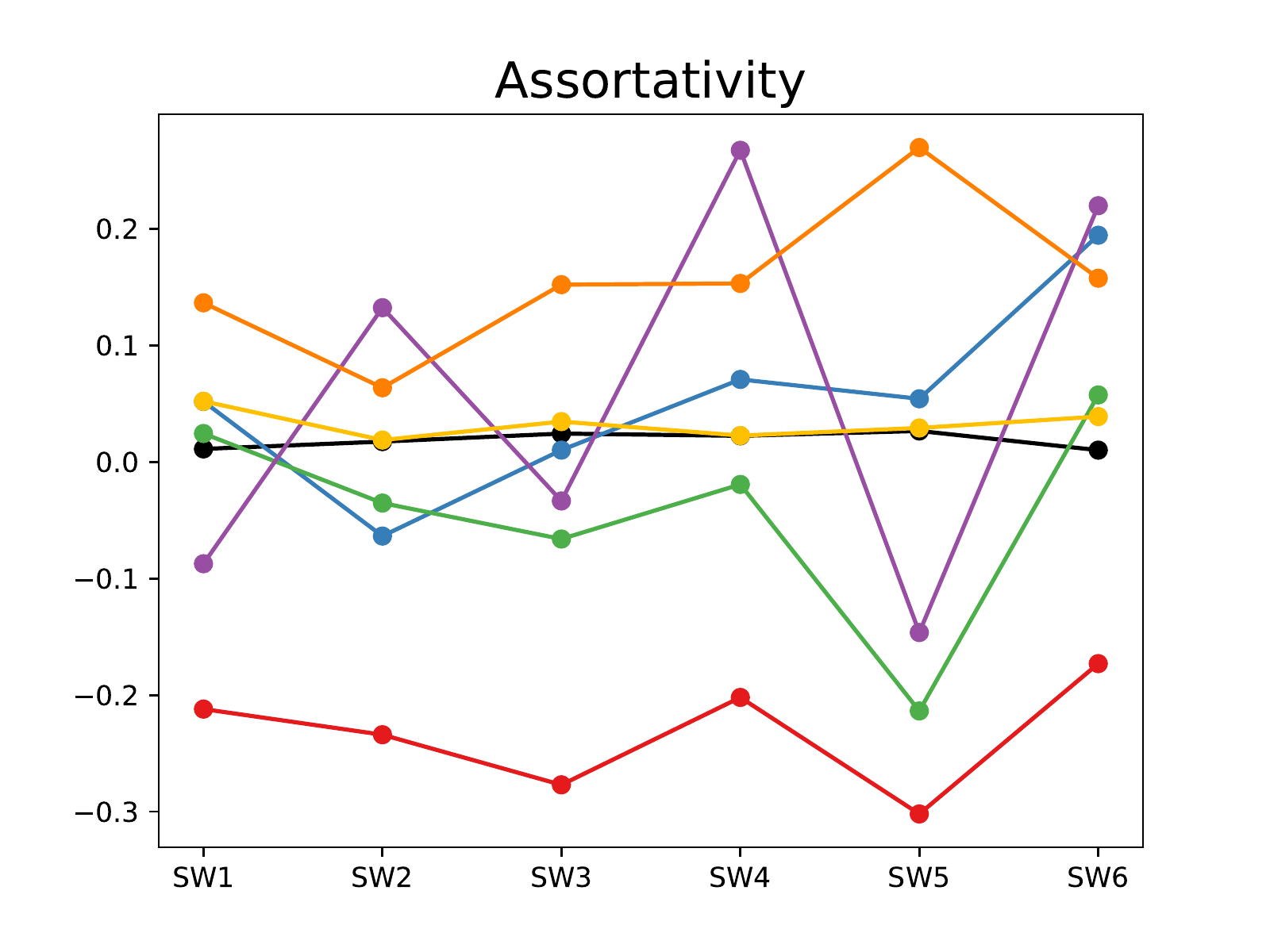}
            
            \label{fig:assortativity}
        \end{subfigure}
        \begin{subfigure}[b]{0.45\textwidth}   
            \centering 
            \includegraphics[width=\textwidth]{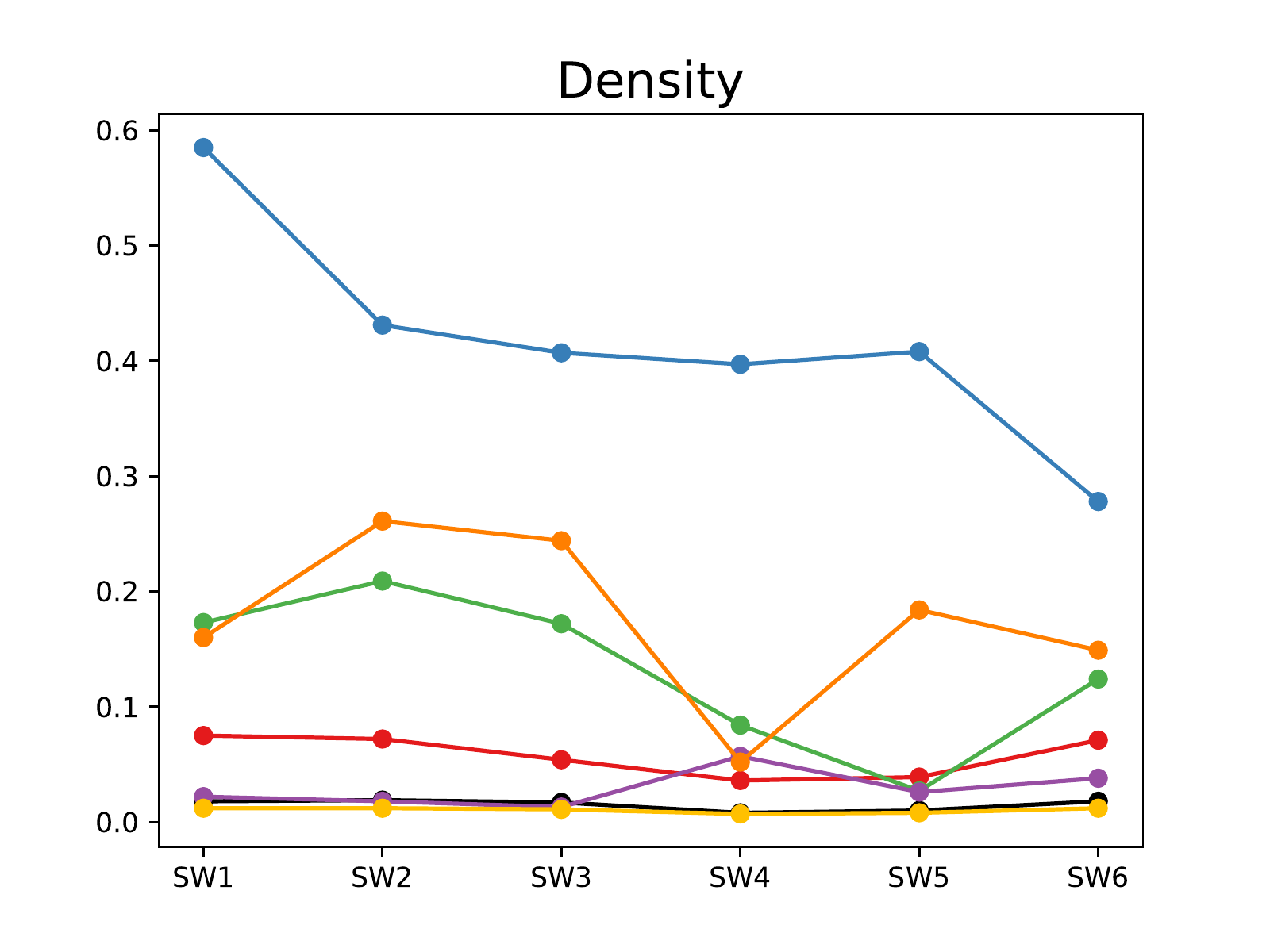}
       
        \label{fig:density}
        \end{subfigure}
        \begin{subfigure}[b]{0.45\textwidth}   
            \centering 
            \includegraphics[width=\textwidth]{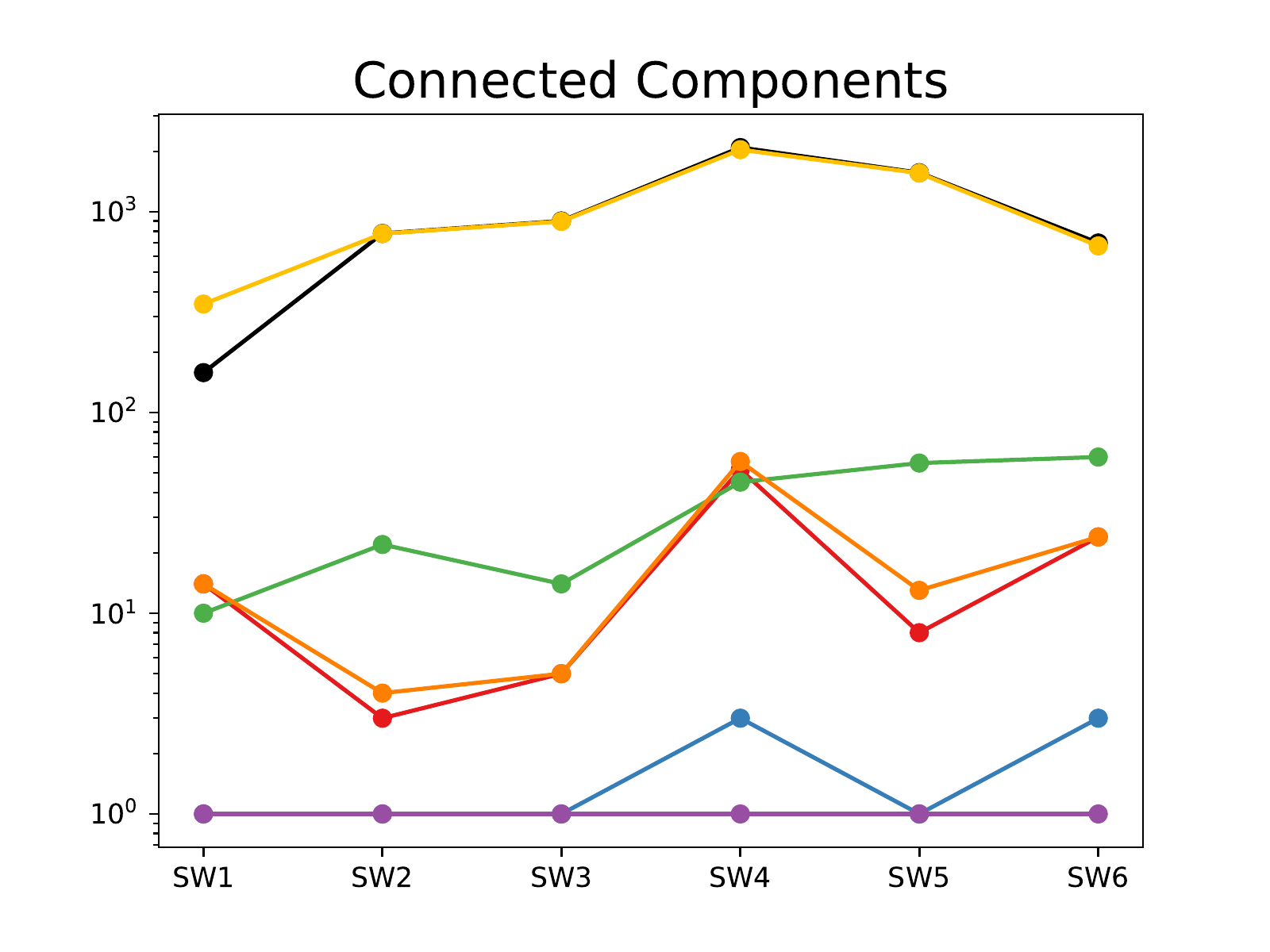}
           
         \label{fig:component}
        \end{subfigure}
        
        \begin{subfigure}[b]{0.8\textwidth}   
            \centering 
            \includegraphics[width=\textwidth]{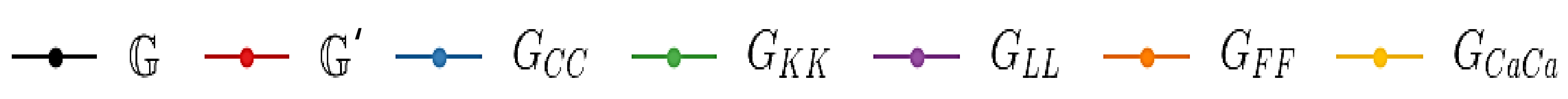}

        \end{subfigure}
        
        \caption{Basic topological properties per layer for each movie of the SW saga.}

         \label{fig:statistics2}
    \end{figure}

Now that we have set the model, we are able to compute measures characterizing it at a macro level. To do so, we  measure the basic topological properties of each layer.  The number of nodes,  number of edges, the network density, the diameter, the average shortest path length, the clustering coefficient and assortativity measure (degree correlation coefficient) are measured for each layer and reported in Figure~\ref{fig:statistics2}.

A first observation is that the character layer $G_{CC}$ contains less nodes than the face layer $G_{FF}$. The number of nodes of location $G_{LL}$ and keyword $G_{KK}$ layers are rather stable across the movies, but the number of nodes in the caption layer $G_{CaCa}$ is varying a lot, and looks quite different between the original and prequel series. 

For all movies, the location $G_{LL}$ layer are made of one single connected component and also for the character $G_{CC}$ layer except for episodes IV and VI. The face layer $G_{FF}$ has a few isolated components, related to extra characters that play no significant role in the story. From the semantic point of view, the keyword layer $G_{KK}$ has a few isolated nodes, and the caption layer $G_{CaCa}$ has a large number of isolated components. 

Results show that the character layer $G_{CC}$ is denser in comparison to all other layers. Indeed, we can expect much more connections among characters, since they exchange dialogues. By comparison, the face layer $G_{FF}$ shows a much higher number of edges than the character layer, both having a very high clustering coefficient, suggesting the existence of social communities. The keyword layer $G_{KK}$ also shows a large clustering coefficient, despite a more limited number of edges.

Location layers $G_{LL}$ display quite a high diameter and the longest average shortest path. This is due to the limited amount of locations and very few temporal transitions between locations that introduce long paths. Only a few sets can be considered hubs. On the opposite, the caption layer $G_{CaCa}$ shows a diameter of 4 and clustering coefficient much closer to 1, because each scene creates a clique of unique captions. The face layer $G_{FF}$ shows the highest assortativity, as we may suspect for main and secondary characters to appear together most often, while tertiary characters (\textit{i.e.} extras) often appear in group.

Caption layers $G_{CaCa}$ show the largest number of nodes and edges with the lowest density. This is due to their generation and construction which creates cliques of many captions for each scenes, which are connected only later on through a few number of captions. As a consequence, captions have many connected components, and display a very short diameter and average shortest past with a high clustering coefficient and an almost null assortativity.

Another consequence is that global characteristics of the multilayer graphs follow mostly those of the caption layers because of their overwhelming number of nodes and clique edges in comparison to all other layers.

We now compare the prequel series (SW1--3) with the original movies (SW4--6). While the average number of nodes in the character layer is comparable, the number of nodes in faces are very different, with much more faces in the original series and the first episode of the prequel. This may be due to the increase use of storm trooper faces during the prequel trilogy, which are not properly detected with our face detector due to their mask. SW1 displays an extremely large amount of face co-occurrence. This is probably due to the scenes putting in action large crowds like during the pod race and other ceremonies. The original trilogy shows on average a high number of face links, with a peak at the last episode, due to the presence of the many Ewoks.

With the exception of SW6, which displays the lowest number of location nodes, the average number of locations are rather similar between the movies, but SW4, the original movie, contains the highest number of transitions between locations. However, this episode does not exhibit a high diameter in comparison to the prequel series, and it displays, together with the original trilogy, the highest clustering coefficient and lowest average shortest path length, suggesting that clusters of locations may occur. This may be the mark of a different style of cuts that depends on the generation of the movie.

The number of keyword nodes is quite comparable between the movies, but the connectivity of those keywords greatly varies between the two trilogies, the prequel trilogy shows a lot more edges in keywords. The number of captions seems, on average, slightly higher in the original series than in the prequel. 

As illustrated in Figure~\ref{fig:statistics2}, there seems to be a significant difference rather consistent across both trilogies in terms of global metrics, all layers considered. Nonetheless, the clustering coefficients remain stable across movies for their individual layers.

\subsection{Node influence within individual layers}

We first investigate the movies for each individual layer. Due to the large number of movie$\times$layer combinations, we only present the result of the influence score (IS)~\cite{bioglio2017movie}. A full detailed account for each episode may be found in the supplementary materials of this paper. For each layer, we report the top 10 nodes sorted by their influence score for each SW episode. 

\subsubsection{Ranking characters}

We first report on the ranking of IS as collected in Table~\ref{tab:character}.
In the prequel trilogy, \textit{Anakin} is always among the top 3 characters. In the original trilogy, his second identity \textit{Vader}, who is first seen in SW3, only appears in the second top tier. \textit{Obi-Wan} gradually gains importance in the prequel trilogy being the top character of the third movie, while his second identity as \textit{Ben} only gets in the last tier of the first movie of the original trilogy. 

Focusing on the first trilogy, \textit{Padme/Amidala} is in the second tier in the first movie, then becomes the main character of the second movie, before being overtaken by \textit{Palpatine} in the third movie, who has a steady growth from the first to the third movie (note that his second identity as \textit{the Emperor} does not appear in the top of the original trilogy). We can add that \textit{Qui-Gon} is the main character of the first movie. The main antagonist characters are also well presented in this top 10 ranking. We have \textit{Nute Gunray} in the first episode, \textit{Count Dooku} in the second episode, and \textit{Grievous} in the third episode.

In the original series, \textit{Luke Skywalker} and \textit{Han Solo} are always in the top 3 characters, with the intrusion of \textit{C-3Po} and \textit{princess Leia}. Beyond \textit{Vader}, antagonists like \textit{Tarkin}, \textit{Piett}, \textit{Veers}, and \textit{Jabba} make their appearance in the top 10 characters too. We can notice that \textit{Lando} only appears in the top of the 6th movie. In addition, \textit{Artoo} and \textit{Chewbacca} are also important protagonists who did not appear in this ranking because they were not properly identified as speakers.

\begin{table}[]
\begin{adjustbox}{max width=1.2\textwidth}
\begin{tabular}{cc cc cc cc cc cc}
\toprule
\multicolumn{12}{c}{\textbf{CHARACTERS} $G_{CC}$}                                                                                            \\ 
\multicolumn{2}{c}{SW1} & \multicolumn{2}{c}{SW2} & \multicolumn{2}{c}{SW3} & \multicolumn{2}{c}{SW4} & \multicolumn{2}{c}{SW5} & \multicolumn{2}{c}{SW6} \\
\midrule
QUI-GON     & 1,00  & PADME     & 1,00  & OBI-WAN     & 1,00  & LUKE      & 1,00  & HAN SOLO    & 1,00  & HAN SOLO    & 1,33  \\ 
ANAKIN    & 2,33  & ANAKIN    & 2,00  & PALPATINE & 2,67  & C-3PO     & 2,00  & LUKE      & 2,33  & C-3PO     & 2,67  \\
JAR JAR     & 2,67  & OBI-WAN     & 3,00  & ANAKIN    & 3,00  & HAN SOLO    & 3,00  & LEIA      & 3,00  & LUKE      & 3,33  \\
OBI-WAN     & 4,67  & M.WINDU   & 4,33  & PADME     & 3,33  & LEIA      & 4,00  & C-3PO     & 3,67  & LANDO     & 3,67  \\ 
PADME     & 5,00  & YODA      & 4,67  & YODA      & 5,00  & VADER     & 5,33  & PIETT     & 5,00  & LEIA      & 4,00  \\ 
AMIDALA   & 5,67  & PALPATINE & 6,33  & B.ORGANA  & 6,33  & BIGGS     & 6,67  & VADER     & 6,00  & VADER     & 6,00  \\ 
PANAKA    & 7,33  & C-3PO     & 6,67  & D.VADER   & 8,00  & I.OFFICER & 8,67  & RIEEKAN   & 7,33  & ACKBAR    & 8,00  \\ 
NUTE      & 8,00  & JAR JAR     & 8,33  & N.GUNRAY  & 8,33  & BEN       & 8,67  & ANNOUNCER & 8,33  & WEDGE     & 8,67  \\ 
PALPATINE & 9,33  & C.DOOKU   & 9,67  & GRIEVOUS  & 10,00 & TARKIN    & 9,33  & WEDGE     & 9,33  & COMMANDER & 9,00  \\ 
R.OLIE    & 10,67 & M.AMEDDA  & 11,00 & M.AMEDDA  & 10,67 & R.LEADER  & 11,33 & VEERS     & 10,33 & JABBA     & 10,33 \\ \bottomrule
\end{tabular}
\end{adjustbox}
\captionof{table}{Top 10 nodes sorted and their influence score of the character layer $G_{CC}$ for each of the 6 SW movies.}

\label{tab:character}

\end{table}

\begin{table}[]
\begin{adjustbox}{max width=1.2\textwidth}

\begin{tabular}{ cc cc cc cc cc cc }
\toprule
\multicolumn{12}{c}{\textbf{FACES} $G_{FF}$}\\ 
\multicolumn{2}{c}{SW1} & \multicolumn{2}{c}{SW2} & \multicolumn{2}{c}{SW3} & \multicolumn{2}{c}{SW4} & \multicolumn{2}{c}{SW5} & \multicolumn{2}{c}{SW6}    \\ \midrule
QUI-GON            & 1,00   & ANAKIN        & 1,33     & ANAKIN         & 1,00    & LUKE           & 1,00    & LEIA           & 1,00    & HAN SOLO              & 1,67  \\ 
A.DOPPELGANGER   & 2,00   & OBI-WAN         & 2,00     & OBI-WAN          & 2,00    & HAN SOLO         & 2,33    & HAN            & 2,67    & LUKE                & 2,00  \\ 
ANAKIN           & 3,00   & AMIDALA       & 2,67     & PALPATINE      & 3,00    & LEIA           & 3,33    & LUKE           & 4,00    & LEIA                & 3,00  \\ 
JAR JAR            & 4,00   & M.WINDU       & 4,00     & YODA           & 4,00    & OBI-WAN          & 4,33    & CHEWBACCA      & 4,00    & C-3PO               & 4,67  \\ 
OBI-WAN            & 5,00   & YODA          & 5,67     & B.B.ORGANA     & 5,33    & CHEWBACCA      & 4,67    & L.TECHNICIAN   & 5,67    & CHEWBACCA           & 5,33  \\ 
PANAKA           & 6,33   & DOOKU         & 6,67     & GRIEVOUS       & 7,00    & C-3PO          & 5,33    & C-3PO          & 6,00    & LANDO               & 6,67  \\ 
SHMI             & 6,67   & J.FETT        & 8,67     & M.WINDU        & 7,33    & DODGE          & 8,67    & PIETT          & 8,00    & P.FOLLOWER          & 7,67  \\
PADME            & 8,33   & B.FETT        & 9,67     & NUTE\_GUARD    & 8,00    & TARKIN         & 9,00    & RIEEKAN        & 8,33    & BIGGS               & 9,00  \\ 
SEBULBA          & 9,67   & KI-ADI        & 10,33    & M.AMEDDA       & 9,00    & R.LEADER       & 9,00    & STORMTROOPER   & 10,33   & J.MUSICIAN\_1006\_0 & 10,67 \\ 
PALPATINE        & 11,67  & P.FOLLOWER    & 11,67    & CHEWBACCA      & 9,33    & C.BARTENDER    & 10,33   & R.OFFICER      & 11,33   & EWOK\_1             & 14,33 \\ \bottomrule
\end{tabular}
\end{adjustbox}
\captionof{table}{Top 10 nodes sorted and their influence score of the face layer $G_{FF}$ for each of the 6 SW movies.}
\label{tab:face}
\end{table}

\subsubsection{Ranking faces}

Observing the ranking of faces in Table~\ref{tab:face} gives a different side of the story, and some new characters make it to the top, due to the length of some scenes. The main changes we observe happen in the second and last tiers of the rankings.

For example, Padme is a role, that was played by different characters, and since Amidala is also Padme, \textit{Amidala's doppelganger} makes it to top ranking. It seems that she is not playing an important role in the movie, but its presence in almost all scenes makes her in the top of the list. \textit{Shmi} (the mother of \textit{Anakin}) and \textit{Sebulba} (Anakin's main opponent during the pod race) are two important characters for the narration of Anakin's side of the story. \textit{Jango Fett} and \textit{Boba Fett} are two key characters in the construction of the drone army, who appear only from their face occurrence in SW2. In SW3, we may notice the addition of \textit{Chewbacca} first, who happen to be a key character in the following trilogy. We may also underline the appearance of \textit{Mace Windu} who does not play a major role for this episode, but who is played by the very popular actor Samuel Lee Jackson.

In the original trilogy of SW, we may also confirm the characters ranking with \emph{Luke Skywalker}, \emph{Leia}, and \emph{Han Solo} on the top rankings. However in the whole trilogy, we see \textit{Chewbacca} reach the first half of the rankings, and interesting newcomers such as the \textit{Cantina's bartender}, central to the iconic Cantina scene in SW4. Secondary characters as technicians and stormtroopers reach in SW5, which exposes more the military organization of the rebellion. SW5 introduces a lot of new characters such as one of \textit{Jabba's musicians}, \textit{Biggs} (a member of the rebel) and an \textit{Ewok}. 

All in all, faces and characters are mostly common when we compare the top protagonists, but interesting changes occur on the secondary characters, and introduces key characters either from the length of scenes (like \textit{Sebulba}), because they would not speak (like \textit{Chewbacca}), or for more commercial reasons (like \textit{Mace Windu}).

\subsubsection{Ranking locations}

We report the ranking results of locations in Table~\ref{tab:location}. Note that we made abbreviations to improve the table readability. The table of abbreviations may be found in the supplementary materials. We may first notice that in the prequel series, there is no actual redundancy of locations, whereas the original series has the \textit{Millenium Falcon} as a key location to access most of the others. However the locations are described in a tree manner (\textit{e.g.} \textit{Hoth - Ice plain - Snow Trench)}, but since it is not consistent across all movies, we only consider them as leaves in this study and keep the hierarchical analysis for a future work.

The top location of the first movie is the \textit{Federation Battleship Bridge} (\textit{FBB}), where the movie starts, and where the two first antagonists are introduced. The ship is wide and contains many different areas hence making a central area in the location layer. In the second movie, there is no one top location but a more evenly distributed top locations, among them \textit{Cockpit Naboo Starship - Sunset} (\textit{CNSS}) in which Anakin and Padme travel to Tatooin, the \textit{Senate Building - Padme's Appartement Bedroom} (\textit{SBPAB}) in which Anakin and Padme start developing their relationship, and \textit{Space} (SP) which is central to battles. In SW3, the \textit{Plaza Jedi Temple-Coruscant} (\textit{PJTC}) is the heart location where all dramatic development happened.

In the original series, from SW4, the main locations are the \textit{Space Craft in Space} (\textit{SIS}) because space battles are central to movie, and even from the first scene, and the final battle from \textit{Luke's XWing Fighter - Cockpit} (\textit{LXFC}), where he destroys the Death Star. These locations are central because these scenes display a lot of cuts between different vessels. The last two movies are really centred on the \textit{Millenium Falcon}, from the \textit{Main Hangar} (\textit{MHMFC}) in SW5 and the \textit{cockpit} (\textit{MFC}) in SW6. The Millenium is iconic of the original series, and the main protagonists travel in this space ship.

\begin{table}[]
\begin{adjustbox}{max width=1.2\textwidth}

\begin{tabular}{cc cc cc cc cc cc}
\toprule
\multicolumn{12}{c}{\textbf{LOCATIONS} $G_{LL}$}                                                                                                                       \\ 
\multicolumn{2}{c}{SW1} & \multicolumn{2}{c}{SW2} & \multicolumn{2}{c}{SW3} & \multicolumn{2}{c}{SW4} & \multicolumn{2}{c}{SW5} & SW6   &       
\\ 
\midrule
FBB         & 1,00        & CNSS        & 3,00       & PJTC        & 1,33       & SIS         & 3,00       & MHMFC        & 1,00      & MFC   & 1,00  \\ 
NGP         & 4,00        & SBPAB       & 3,67       & MMCC        & 2,67       & LXFC        & 3,67       & HB           & 4,67      & DSCR  & 2,67  \\ 
TCH         & 4,67        & SP          & 4,00       & MCP         & 11,00      & DSCR        & 6,33       & HRBMHD       & 5,00      & RSCB  & 3,67  \\ 
TDNS        & 6,67        & TCKLP       & 5,33       & CSCMA       & 11,33      & SATDS       & 7,00       & HRBCC        & 8,00      & ETTR  & 4,67  \\ 
FBCR        & 9,33        & CCD         & 6,00       & ASH         & 12,00      & MFC         & 7,67       & BOCCWVD      & 8,00      & SKI   & 7,00  \\
AHMR        & 11,33       & TCCE        & 6,33       & OBS         & 13,00      & MOWR        & 9,00       & DVSDBMCD     & 8,33      & SRF   & 7,67  \\ 
NSMA        & 11,67       & GLA         & 8,00       & PJTCR       & 13,00      & SOTDS       & 9,00       & SIF          & 9,33      & DSMDB & 8,33  \\ 
SCU         & 12,33       & THMF        & 10,00      & ULP         & 13,33      & MFGC        & 10,33      & HIPST        & 12,00     & FGB   & 9,00  \\ 
NSC         & 13,00       & GEA         & 10,67      & IDC         & 15,33      & RLC         & 12,67      & MFGAC        & 13,67     & RTJPT & 9,33  \\ 
NPTR        & 14,67       & TC          & 15,33      & LPN         & 17,00      & DSCOR       & 13,00      & LSRLC        & 16,33     & JTR   & 9,67  \\ \bottomrule
\end{tabular}
\end{adjustbox}
\captionof{table}{Top 10 nodes sorted and their influence score of the location layer $G_{LL}$ for each of the 6 SW movies.}
\label{tab:location}
\end{table}

\subsubsection{Ranking keywords}

We now report the ranking of keywords in Table~\ref{tab:keyword}, of which we find mentions to some key characters. 

In the prequel series, there is mention of the \textit{chancellor} as a key word in all three episodes, and growing to the third episode since the chancellor is the Emperor corrupting Anakin. \textit{Queen} is specific to SW1 which the movie revolves around. \textit{Annie} (Anakin) is mentioned in the second movie, which is interesting since it is his tender name, and the movie develops their relationship with Padme. \textit{Windu} and \textit{Yoda} are mentioned in the third movie, which revolves around the conflict between the Jedi council they represent and Anakin.

Beyond character keywords, the \textit{federation}, \textit{senate}, \textit{republic} are recurring keywords highlighting the political tone of the first series. \textit{Master}, \textit{Jedi} and the \textit{Force} make the relationship with the ``religious/magic'' part of the series. 

In the original series, a lot of main characters enter the top ranking. We can mention that \textit{Han} is on the top of SW4, beyond the main character who is \textit{Luke}. \textit{Artoo} (R2-D2) and \textit{Chewie} (Chewbacca) are also introduced SW4, which is interesting because neither the script characters or the face detection helped reveal Artoo in the main protagonists. From SW5, \textit{father} is by far the top keyword, which is the key revelation of this episode. \textit{Han} loses some ranks, and the reference to the \textit{Princess} (Leia) enter the top. In the last episode, references to one main antagonist, \textit{Jabba} enters the top keywords, and \textit{Threepio} (C-3PO) and \textit{Yoda} enter the top. 

Beyond characters, vocabulary related to space vessels appear (\textit{ship}, \textit{main}, \textit{energy}, \textit{field}). The philosophical question of \textit{good} (in opposition to the dark side) is also as an important keyword, in combination with \textit{master} which is core to the structure of Jedi (protagonist) and Sith (antagonist) organizations.

\begin{table}[]

\begin{adjustbox}{max width=1.2\textwidth}

\begin{tabular}{ cc cc cc cc cc cc }
\toprule
\multicolumn{12}{c}{\textbf{KEYWORDS} $G_{KK}$}                                                                                                                                  \\ 
\multicolumn{2}{c}{SW1} & \multicolumn{2}{c}{SW2} & \multicolumn{2}{c}{SW3} & \multicolumn{2}{c}{SW4} & \multicolumn{2}{c}{SW5} & \multicolumn{2}{c}{SW6} \\ \midrule
federation     & 2,00     & jedi          & 1,00     & jedi          & 1,33     & ship         & 1,00      & father       & 1,00      & han          & 1,00      \\
jedi           & 3,00     & senator       & 2,00     & anakin        & 1,67     & han          & 2,33      & master       & 4,33      & luke         & 2,00      \\
queen          & 3,67     & master        & 3,00     & chancellor    & 4,33     & imperial     & 4,00      & ship         & 4,67      & artoo        & 4,00      \\ 
senate         & 5,33     & great         & 4,33     & master        & 4,33     & hear         & 4,00      & luke         & 5,67      & jabba        & 5,33      \\ 
time           & 6,33     & republic      & 4,67     & force         & 5,33     & main         & 5,00      & artoo        & 6,00      & father       & 5,67      \\
people         & 6,33     & continuing    & 7,00     & council       & 8,00     & luke         & 5,33      & energy       & 7,67      & master       & 6,33      \\
naboo          & 6,67     & annie         & 7,67     & time          & 8,33     & artoo        & 9,67      & han          & 8,67      & vader        & 9,00      \\ 
master         & 8,67     & time          & 8,00     & windu         & 8,33     & good         & 10,67     & chewie       & 10,00     & threepio     & 9,33      \\ 
back           & 9,33     & chancellor    & 8,33     & republic      & 9,33     & chewie       & 11,00     & field        & 10,33     & good         & 12,33     \\ 
chancellor     & 9,33     & naboo         & 11,67    & yoda          & 12,33    & shut         & 11,67     & princess     & 12,00     & yoda         & 14,00     \\ \bottomrule
\end{tabular}

\end{adjustbox}
\captionof{table}{Top 10 nodes sorted and their influence score of the keyword layer $G_{KK}$ for each of the 6 SW movies.
}
\label{tab:keyword}

\end{table}

\subsubsection{Ranking captions}

The ranking of captions, reported in Table~\ref{tab:caption}, suggests that most of the visual similarity between scenes is focused on people's outfit rather than anything else, thanks to the term \textit{wearing} which is almost all of the top captions. Nonetheless, this capture well the visual identity of the movies. 

In the prequel series, the appearances of Queen Amidala is remarked from her multiple outfits, and those of her followers. The term \textit{woman} appears a lot in the top captions of SW1 and gradually decreases in the following episodes. SW1 shows a wide range of colors associated with \textit{wearing}: \textit{black}, \textit{red}, \textit{white}, \textit{blue}, and \textit{gray}. The following two episodes mostly bring forward the \textit{black} jacket of Anakin's outfit, and \textit{white} clothes which correspond to the numerous clones' armor. We may also notice the introduction of the \textit{brown} outfit that is representative of Jedi knights.

The original series introduces \textit{helmets} or \textit{hat} wearing people, which often matches the outfit of Darth Vader and all the different military people in both the Empire and Rebel armies. Top colors are greatly focused on \textit{black}, which is most represented by Vader, and \textit{white} which is the main color of Luke's outfit. The last episode introduces \textit{green} outfits that are the ones worn by the Rebels in all actions happening in the forests of Endor moon.

\begin{table}[]
\begin{adjustbox}{max width=1.2\textwidth}
\centering
\begin{tabular}{cc cc cc}
\toprule
\multicolumn{6}{c}{\textbf{CAPTIONS} $G_{CaCa}$}                                              \\ 
\multicolumn{2}{c}{SW1}                &
\multicolumn{2}{c}{SW2}                & \multicolumn{2}{c}{SW3}                      \\ 
\midrule
a,black,jacket,wearing & 1,67  & a,black,shirt,wearing  & 1,00  & a,black,shirt,wearing  & 1,00  \\
a,black,wearing,woman  & 2,00  & a,shirt,wearing,white  & 2,33  & a,black,jacket,wearing & 2,33  \\
a,red,wearing,woman    & 3,33  & a,black,jacket,wearing & 2,67  & a,shirt,wearing,white  & 3,00  \\
a,red,shirt,wearing    & 4,33  & a,black,man,wearing    & 4,00  & a,black,man,wearing    & 3,67  \\
a,shirt,wearing,white  & 4,67  & a,black,wearing,woman  & 5,67  & a,black,wearing,woman  & 5,00  \\
a,jacket,red,wearing   & 6,67  & a,wearing,white,woman  & 6,67  & a,man,wearing,white    & 6,00  \\
a,black,man,wearing    & 7,33  & a,brown,shirt,wearing  & 7,00  & a,red,shirt,wearing    & 7,33  \\
a,blue,wearing,woman   & 10,33 & a,red,shirt,wearing    & 7,00  & a,blue,shirt,wearing   & 8,00  \\
a,man,red,wearing      & 10,67 & a,brown,chair,wooden   & 9,33  & a,brown,shirt,wearing  & 9,67  \\
a,gray,shirt,wearing   & 12,00 & a,blue,shirt,wearing   & 10,33 & a,hat,man,wearing      & 11,33 \\ \midrule
\multicolumn{2}{c}{SW4}                   & \multicolumn{2}{c}{SW5}                & \multicolumn{2}{c}{SW6}                      \\ \midrule
a,shirt,wearing,white  & 1,00  & a,shirt,wearing,white  & 1,00  & a,shirt,wearing,white  & 1,33  \\
a,man,wearing,white    & 2,00  & a,man,wearing,white    & 2,00  & a,black,jacket,wearing & 3,00  \\
a,black,shirt,wearing  & 3,33  & a,black,shirt,wearing  & 3,33  & a,man,wearing,white    & 3,33  \\
a,black,man,wearing    & 4,33  & a,black,man,wearing    & 4,33  & and,red,sign,white     & 4,33  \\
a,red,shirt,wearing    & 5,00  & a,red,shirt,wearing    & 5,00  & a,black,shirt,wearing  & 4,33  \\
a,black,jacket,wearing & 5,33  & a,black,jacket,wearing & 5,33  & a,black,man,wearing    & 5,67  \\
a,wearing,white,woman  & 7,33  & a,wearing,white,woman  & 7,33  & a,hat,man,wearing      & 7,33  \\
a,helmet,man,wearing   & 9,67  & a,helmet,man,wearing   & 9,67  & a,green,shirt,wearing  & 7,67  \\
and,bag,black,white    & 10,00 & and,bag,black,white    & 10,00 & a,black,wearing,woman  & 9,67  \\
a,black,wearing,woman  & 11,33 & a,black,wearing,woman  & 11,33 & a,man,shirt,wearing    & 9,67   \\ \bottomrule
\end{tabular}

\end{adjustbox}
\captionof{table}{
Top 10 nodes sorted and their influence score of the caption layer $G_{CaCa}$ for each of the 6 SW movies.
}

\label{tab:caption}
\end{table}

\subsection{Node influence in the multilayer network}

We now analyze node influence score from the multilayer networks as reported in Table~\ref{tab:multilayer}. Interesting nodes in this network highlight and associate different key elements of the story. As illustrated in the global topological analysis of Section~\ref{sec:topology}, the caption layer has order of magnitude differences with all other layers in terms of size, hence strongly influencing the ranking. Our multilayer model allows for investigating this difference by simply checking rankings in the multilayer network $\mathbb G'=\mathbb G - G_{CaCa}$ with all layers except the caption layer (in Table~\ref{tab:MULTI-WITHOUT-CAPTION}). 

Recalling topological properties as displayed in Figure~\ref{fig:statistics2}, we may notice that the whole multilayer $\mathbb G$ behaves similarly to the caption layer $G_{CaCa}$, except for diameter which becomes significantly smaller. The multilayer without captions $\mathbb G'$ shows a rather low density, but a high clustering coefficient suggesting a of a community structure organization. Most interestingly, it displays a negative assortativity, meaning that high degree nodes tend to connect preferably with low degree nodes. This is probably an effect of the association to location nodes within the graph.

\subsubsection{Multilayer network, all layers}

The first thing we may notice is that face $G_{FF}$ and character $G_{CC}$ layers are prominent in the results, then comes the caption layer $G_{CaCa}$ and the keyword layer $G_{KK}$. The fact that captions are not only numerous but cliques generated for each scene reinforces their influence score. However, we have a good amount of redundancy between people over face, script, and keyword detections, confirming these stories are centred around the narration of characters' adventures. 

The first movie bring forward all the top characters we may find everywhere, the main protagonists, Qui-Gon, Obi-Wan, with Amidala (through her doppelgangers) and Anakin. The very controversial Jar Jar is often felt as over-represented by the fandom, and we can only confirm this in this ranking. Anakin and Amidala/Padme make the top of the next movie, which revolves over their relationship, and the development of the Jedi training of Anakin, hence the prominent keywords \textit{Master} and \textit{Jedi}. For the last episode of the prequel trilogy, Anakin and Obi-Wan are the top most represented characters (since this episode will lead them to a fight), and their master/Jedi relationship is taking prominence from the keywords. We may notice the introduction of the Jedi master Yoda in the top ranking, a highly central character of the whole series, who is leading the Jedi council in this episode. One main character that was most influential in the face and character layers was Palpatine, but he is absent from the top ranking in the multilayer. This is indicative of his strong connection with a few characters and places in the plot of SW3 for instance with Anakin and mostly on Coruscant. Amidala/Padme is also a central character in SW2 and SW3 but she is stranded on Coruscant for most of the latter film, whereas her and Anakin where travelling a lot in the former. There is no specific conclusion from the captions' perspective, other than black outfits are dominating this series.

The two first episodes of the original series see much more captions being brought forward. Beyond the black and white outfits we discussed in the previous section, we may notice the introduction of \textit{red shirts} which are none other than the uniform of the Rebels. Luke, Leia, and Han Solo are the most represented characters, following the cast distribution. We may also notice in SW5 the mention to \textit{comlink} because the characters and separated in different sites throughout the movie, and communicates a lot through this device. The last episode unifies subplot in which secondary characters also play more important roles (such as delivering Solo, or cutting the power from Endor) and we see this in the introduction of other charismatic characters: C-3PO, Chewbacca, and Lando. 

Although the location layer nodes are not represented, the influence of the layer through links to characters may be observed. Prominent character nodes (whichever the layer) that are brought forward often correspond to those traveling a lot between locations. For example, although Amidala is central in SW3, she enters the top in SW2 where she travels a lot, and the other around is true for Yoda who travels a lot in SW3. 

\begin{table}[]
\begin{adjustbox}{max width=1.2\textwidth}
\centering
\begin{tabular}{ccc ccc ccc}
\toprule
\multicolumn{9}{c}{\textbf{MULTILAYER, ALL LAYERS} $\mathbb G$}                                                                                                         \\ 
\multicolumn{3}{c}{SW1}                   & \multicolumn{3}{c}{SW2}                & \multicolumn{3}{c}{SW3}                      \\ 
\midrule
QUI-GON& $G_{FF}$                   & 1,00  & ANAKIN& $G_{FF}$                  & 1,67  & ANAKIN& $G_{FF}$                  & 2,33  \\
A.DOPPELGANGER& $G_{FF}$          & 2,33  & AMIDALA& $G_{FF}$                 & 3,00  & OBI-WAN& $G_{FF}$                   & 3,33  \\
ANAKIN& $G_{FF}$                  & 2,67  & master& $G_{KK}$                  & 3,33  & jedi& $G_{KK}$                    & 5,00  \\
OBI-WAN& $G_{FF}$                   & 4,00  & OBI-WAN& $G_{FF}$                   & 4,33  & anakin& $G_{KK}$                  & 5,33  \\
QUI-GON& $G_{CC}$                   & 5,33  & a,black,shirt,wearing& $G_{CaCa}$  & 6,00  & a,black,shirt,wearing& $G_{CaCa}$  & 6,67  \\
JAR JAR& $G_{FF}$                   & 5,67  & jedi& $G_{KK}$                    & 6,33  & YODA& $G_{FF}$                    & 8,00  \\
ANAKIN& $G_{CC}$                  & 7,00  & continuing& $G_{KK}$              & 8,00  & a,black,jacket,wearing& $G_{CaCa}$ & 8,33  \\
a,black,jacket,wearing& $G_{CaCa}$ & 9,33  & a,black,jacket,wearing& $G_{CaCa}$ & 8,33  & master& $G_{KK}$                  & 8,67  \\
a,black,wearing,woman& $G_{CaCa}$  & 9,67  & PADME& $G_{CC}$                   & 9,67  & a,black,man,wearing& $G_{CaCa}$    & 9,33  \\
JAR JAR& $G_{CC}$                   & 11,67 & a,black,wearing,woman& $G_{CaCa}$  & 10,33 & anakin& $G_{KK}$                  & 11,67 \\ \midrule
\multicolumn{3}{c}{SW4}                   & \multicolumn{3}{c}{SW5}                & \multicolumn{3}{c}{SW6}                      \\ \midrule
a,shirt,wearing,white& $G_{CaCa}$  & 1,33  & LEIA& $G_{FF}$                    & 1,00  & LUKE& $G_{FF}$                    & 1,00  \\
LUKE& $G_{FF}$                    & 1,67  & HAN SOLO& $G_{FF}$                  & 3,00  & HAN SOLO& $G_{FF}$                  & 2,00  \\
a,man,wearing,white& $G_{CaCa}$    & 3,00  & a,shirt,wearing,white& $G_{CaCa}$  & 3,00  & LEIA& $G_{FF}$                    & 3,00  \\
LEIA& $G_{FF}$                    & 5,33  & a,black,shirt,wearing& $G_{CaCa}$  & 4,33  & HAN SOLO& $G_{CC}$                  & 5,33  \\
a,black,shirt,wearing& $G_{CaCa}$  & 5,67  & LUKE& $G_{FF}$                    & 6,00  & luke& $G_{KK}$                    & 6,33  \\
a,black,man,wearing& $G_{CaCa}$    & 7,00  & a,black,man,wearing& $G_{CaCa}$    & 7,00  & C-3PO& $G_{CC}$                   & 6,67  \\
a,black,jacket,wearing& $G_{CaCa}$ & 7,67  & a,black,jacket,wearing& $G_{CaCa}$ & 7,33  & C-3PO& $G_{FF}$                   & 7,67  \\
a,red,shirt,wearing& $G_{CaCa}$    & 8,33  & a,wearing,white,woman& $G_{CaCa}$  & 9,33  & CHEWBACCA& $G_{FF}$               & 10,33 \\
LUKE& $G_{CC}$                    & 10,00 & comlink& $G_{KK}$                 & 9,67  & a,shirt,wearing,white& $G_{CaCa}$  & 11,67 \\
HAN SOLO& $G_{FF}$                  & 10,33 & a,black,wearing,woman& $G_{CaCa}$  & 10,00 & LANDO& $G_{FF}$                   & 12,00 \\  \bottomrule
\end{tabular}

\end{adjustbox}
\captionof{table}{Top 10 nodes sorted, with their layer and influence score of the overall multilayer network $\mathbb G$ for each of the 6 SW movies. 
}

\label{tab:multilayer}
\end{table}

\begin{table}[]

\begin{adjustbox}{max width=1.2\textwidth}

\begin{tabular}{ccc ccc ccc}
\toprule
\multicolumn{9}{c}{\textbf{MULTILAYER, WITHOUT CAPTIONS} $\mathbb G'$}                                                                                                         \\ 
\multicolumn{3}{c}{SW1}                   & \multicolumn{3}{c}{SW2}                & \multicolumn{3}{c}{SW3}                      \\ 
\midrule

QUI-GON& $G_{FF}$          & 1,00  & ANAKIN& $G_{FF}$      & 1,00  & OBI-WAN& $G_{FF}$          & 1,00  \\
A.DOPPELGANGER& $G_{FF}$ & 2,33  & OBI-WAN& $G_{FF}$       & 2,67  & ANAKIN& $G_{FF}$         & 2,00  \\
ANAKIN& $G_{FF}$         & 2,67  & AMIDALA& $G_{FF}$     & 2,67  & YODA& $G_{FF}$           & 3,00  \\
JAR JAR& $G_{FF}$          & 4,33  & PADME& $G_{CC}$  & 3,67  & PALPATINE& $G_{FF}$      & 5,33  \\
OBI-WAN& $G_{FF}$          & 4,67  & ANAKIN& $G_{CC}$ & 5,00  & B.B.ORGANA& $G_{FF}$     & 5,33  \\
QUI-GON& $G_{CC}$     & 6,00  & OBI-WAN& $G_{CC}$  & 6,67  & OBI-WAN& $G_{CC}$     & 5,67  \\
ANAKIN& $G_{CC}$    & 7,00  & M.WINDU& $G_{FF}$     & 7,33  & ANAKIN& $G_{CC}$    & 6,67  \\
PANAKA& $G_{FF}$         & 8,00  & YODA& $G_{FF}$        & 9,00  & DVQSD& $G_{LL}$      & 9,67  \\
SHMI& $G_{FF}$           & 9,00  & jedi& $G_{KK}$     & 9,00  & M.WINDU& $G_{FF}$        & 11,00 \\
PADME& $G_{FF}$          & 10,33 & master& $G_{KK}$   & 10,33 & PALPATINE& $G_{CC}$ & 11,00  \\ \midrule
\multicolumn{3}{c}{SW4}                   & \multicolumn{3}{c}{SW5}                & \multicolumn{3}{c}{SW6}                      \\ \midrule
LUKE        & $G_{FF}$ & 1,00  & LEIA      & $G_{FF}$ & 1,00  & HAN SOLO   & $G_{FF}$ & 1,33  \\
LEIA       & $G_{FF}$ & 2,33  & LUKE      & $G_{FF}$ & 2,00  & LUKE       & $G_{FF}$ & 2,67  \\
LUKE   & $G_{CC}$ & 2,67  & HAN SOLO  & $G_{FF}$ & 3,67  & LEIA       & $G_{FF}$ & 3,00  \\
H.SOLO    & $G_{FF}$ & 4,00  & HAN SOLO  & $G_{CC}$ & 5,33  & C-3PO      & $G_{FF}$ & 3,67  \\
C-3PO      & $G_{FF}$ & 5,00  & MHMFC     & $G_{LL}$ & 6,33  & CHEWBACCA  & $G_{FF}$ & 5,67  \\
C-3PO  & $G_{CC}$ & 6,67  & HRBCC     & $G_{LL}$ & 6,67  & HAN SOLO   & $G_{CC}$ & 6,00  \\
O.WAN       & $G_{FF}$ & 8,00  & CHEWBACCA & $G_{FF}$ & 7,00  & C-3PO      & $G_{CC}$ & 7,67  \\
CHEWBACCA   & $G_{FF}$ & 8,67  & LUKE      & $G_{CC}$ & 8,67  & LANDO      & $G_{FF}$ & 8,33  \\
H.SOLO & $G_{CC}$ & 10,00 & C-3PO     & $G_{FF}$ & 10,33 & LUKE       & $G_{CC}$ & 9,33  \\
MFC    & $G_{LL}$ & 10,67 & YODA      & $G_{FF}$ & 10,67 & HAN SOLO   & $G_{KK}$ & 10,00 \\  \bottomrule
\end{tabular}

\end{adjustbox}
\captionof{table}{Top 10 nodes sorted, with their layer and influence score of the overall multilayer network $\mathbb G'$ for each of the 6 SW movies. 
}
\label{tab:MULTI-WITHOUT-CAPTION}

\end{table}

\subsubsection{Multilayer network, without the caption layer}

The ranking of nodes in $\mathbb G'$ (Table~\ref{tab:MULTI-WITHOUT-CAPTION}) is very close to those of the full multilayer $\mathbb G$ (Table~\ref{tab:multilayer}), with the exception of all captions being taken out of the top. We can however observe a few locations making their place into the top ranking, but less keywords.

From the first episode in the prequel series, the main changes are the following. The ranking of Jar Jar has increased a bit, but we can mostly notice the inclusion of \textit{Shmi}, who is Anakin's mother, a central character in the whole segment concerning Tatooine. Queen Amidala, under her name \textit{Padme}, is also entering the ranking. \textit{Panaka} is the guard who accompanies Amidala/Padme all along to protect her, and take a long participation in most action scenes. In SW2, \textit{Obi-Wan} gains a few ranks, probably for his numerous travels (checking on the clone army). The leaders of the Jedi council, \textit{Mace Windu} and \textit{Yoda} enter the ranking too, and for the next movie. The keywords \textit{Jedi} and \textit{master} are still maintained, underlining the other thema of this movie which revolves around the Jedi training of Anakin. The last movie of the prequel does not show the persistence of these keywords in the top ranking, but sees major introductions of first \textit{Palpatine} who corrupted Anakin, and of \textit{Bail Organa}, a senator organizing the resistance against Palpatine, who will harbour one child of Anakin after his turning to the dark side. A location appears in this movie rankings, which is Darth Vader's Quarter Star Destroyer, in a scene at the ending that exists only in the script, and was finally deleted. 


The original series also sees a lot new nodes replacing captions, above all, \textit{Chewbacca} and \textit{C-3PO}, companions of the main characters, entering all top rankings. In SW4, \textit{Obi-Wan} also enters the ranking, since he guides the young Luke all along this adventure. Most importantly, the \textit{Millenium Falcon Cockpit} (\textit{MFC}) the vessel which caries all characters through their adventure is the main location which enters this ranking. The \textit{comlink} keyword disappears of SW5 but \textit{Yoda} appears in this ranking, since Luke makes the trip to receive training from him during this episode. Two locations enter in the ranking, \textit{Main Hangar - Millenium Falcon - Cockpit} (\textit{MHMFC}) and \textit{Hoth - Rebel Base - Command Center} (\textit{HRBCC}) where most characters regroup during the first part of the movie, before being separated then. In the last episode, nothing changes much except that \textit{Han Solo} takes the leadership of the ranking.

\begin{figure}[t!]
 \centering
 
\begin{subfigure}[]{0.45\textwidth}
   \centering 
    \includegraphics[width=\textwidth]{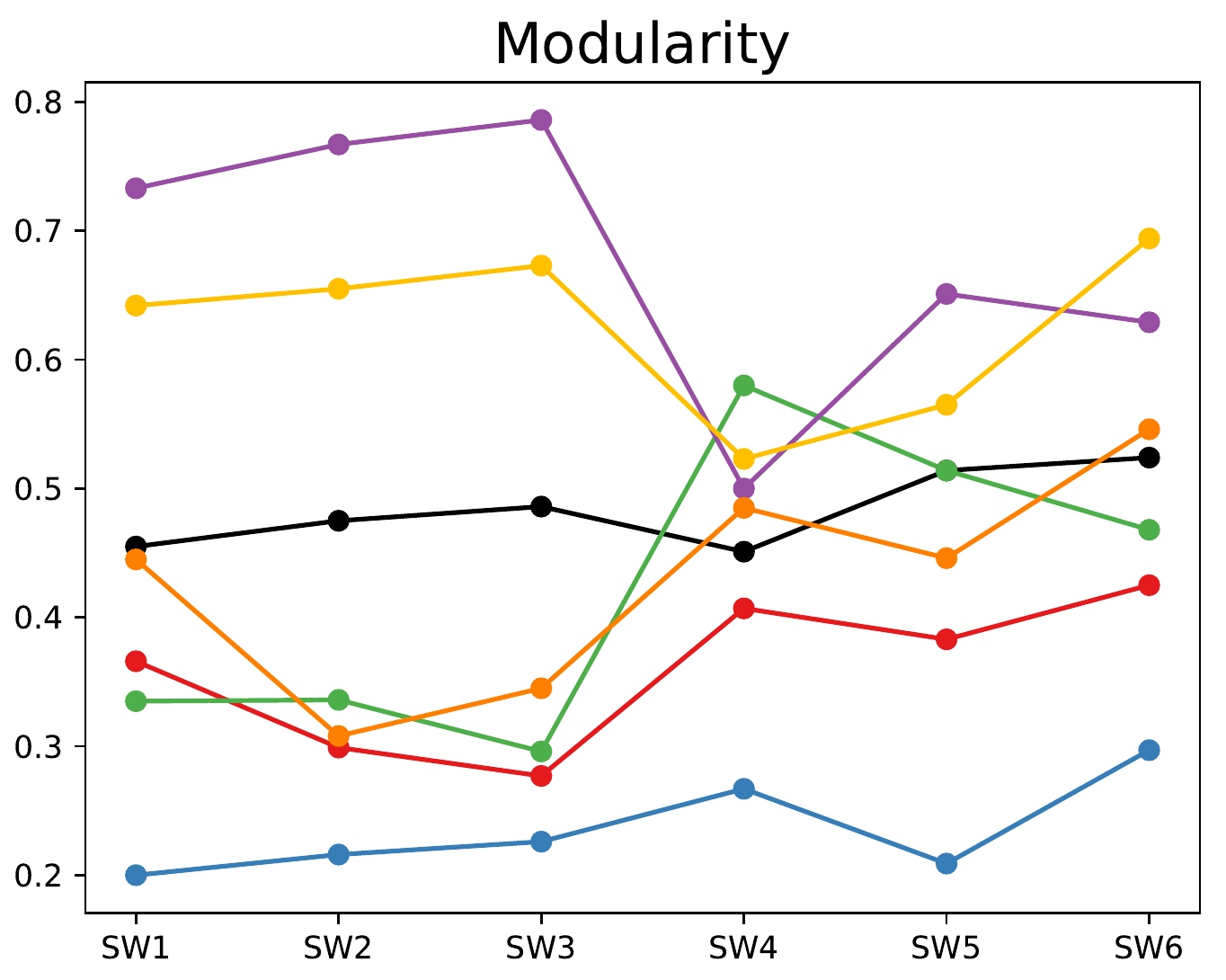}
     \label{fig:modularityy}
\end{subfigure}
\begin{subfigure}[]{0.45\textwidth}  
    \centering 
    \includegraphics[width=\textwidth]{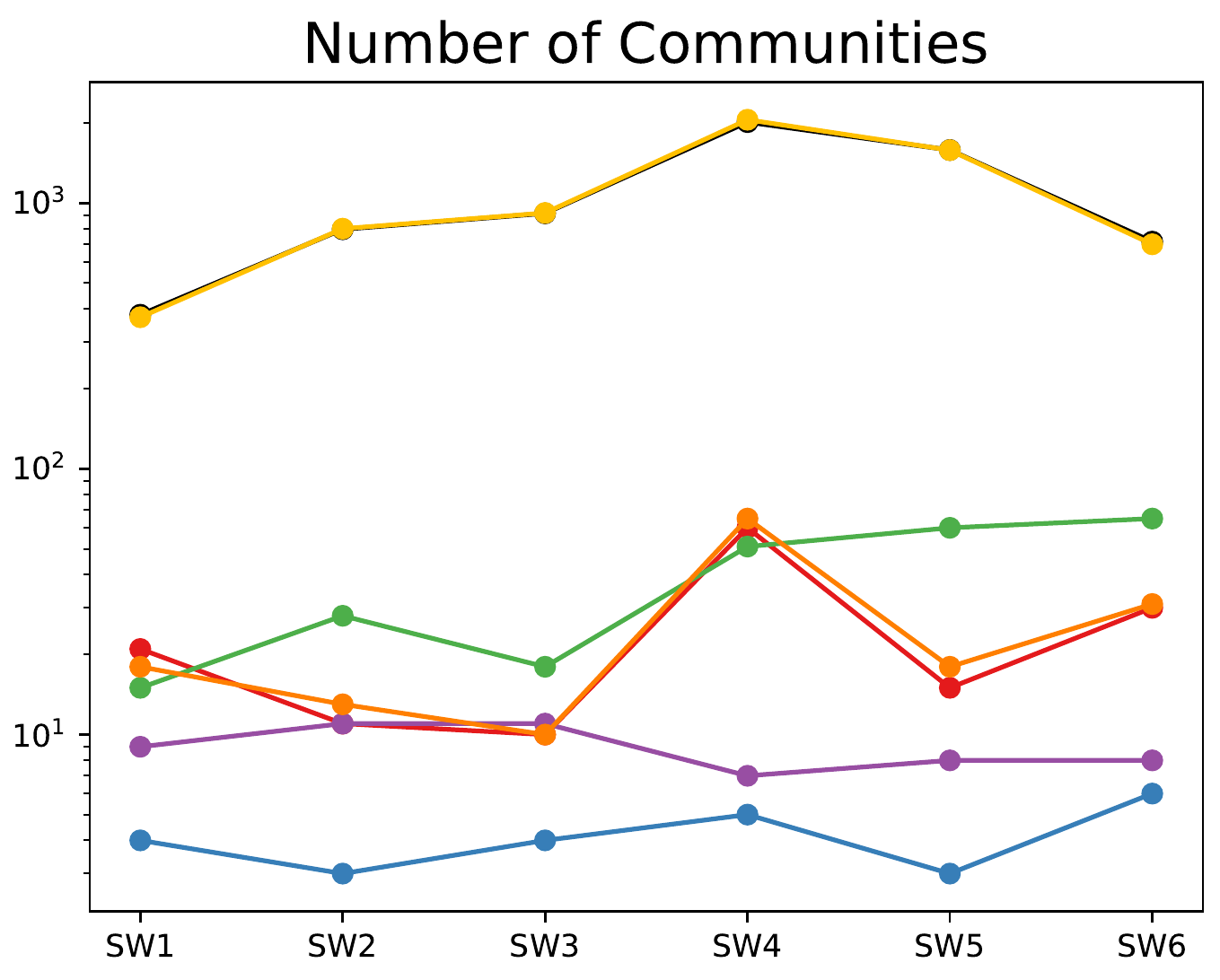}
     \label{fig:communitiess}
\end{subfigure}
\begin{subfigure}[b]{0.8\textwidth}   
            \centering 
            \includegraphics[width=\textwidth]{legende.pdf}

        \end{subfigure}

\caption{Number of communities and modularity per layer for each movie of the saga.}
\label{fig:modularity}
\end{figure}

\begin{figure}[pht]
\centering
        \begin{subfigure}[b]{0.48\textwidth}
            \centering
            \includegraphics[width=\textwidth]{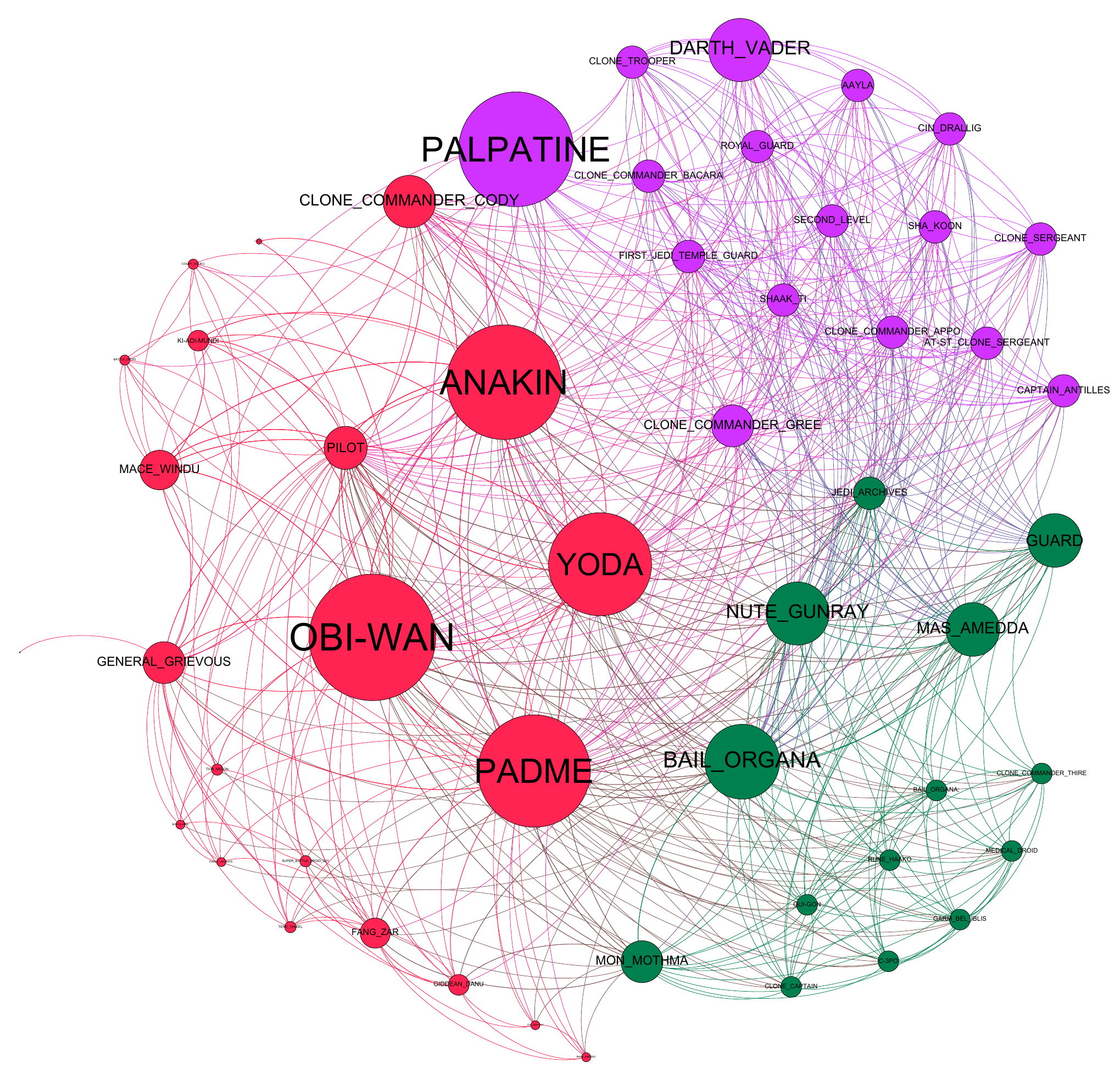}
            \caption[  $G_{CC}$] {$G_{CC}$}
            
            \label{fig:SW3C}
        \end{subfigure}
        \qquad        
        \begin{subfigure}[b]{0.475\textwidth}  
            \centering 
            \includegraphics[width=\textwidth]{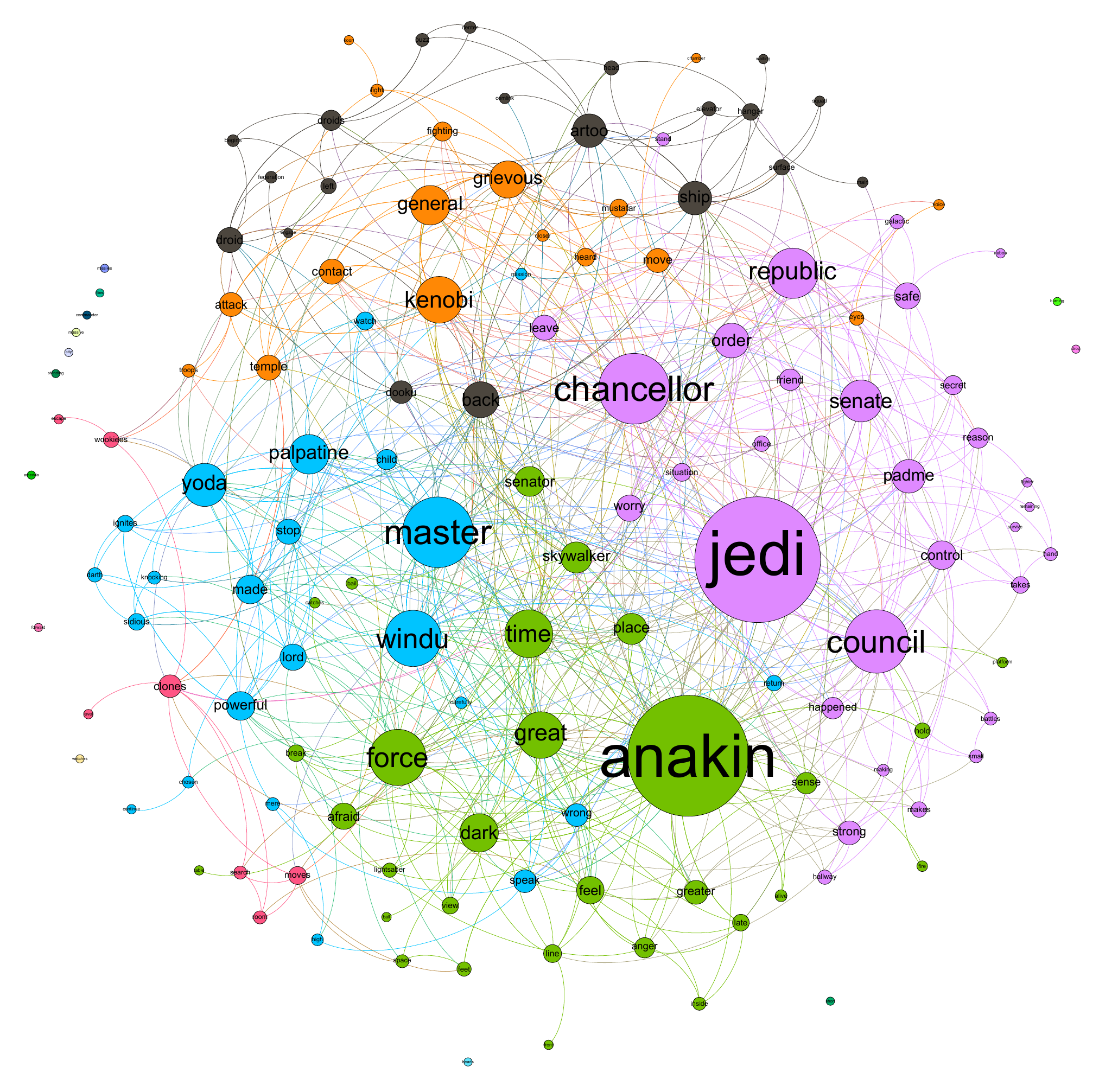}
            \caption[  $G_{KK}$]{$G_{KK}$}
            
            \label{fig:SW3K}
        \end{subfigure}
         \vskip\baselineskip    
        \begin{subfigure}[b]{0.475\textwidth}  
            \centering 
            \includegraphics[width=\textwidth]{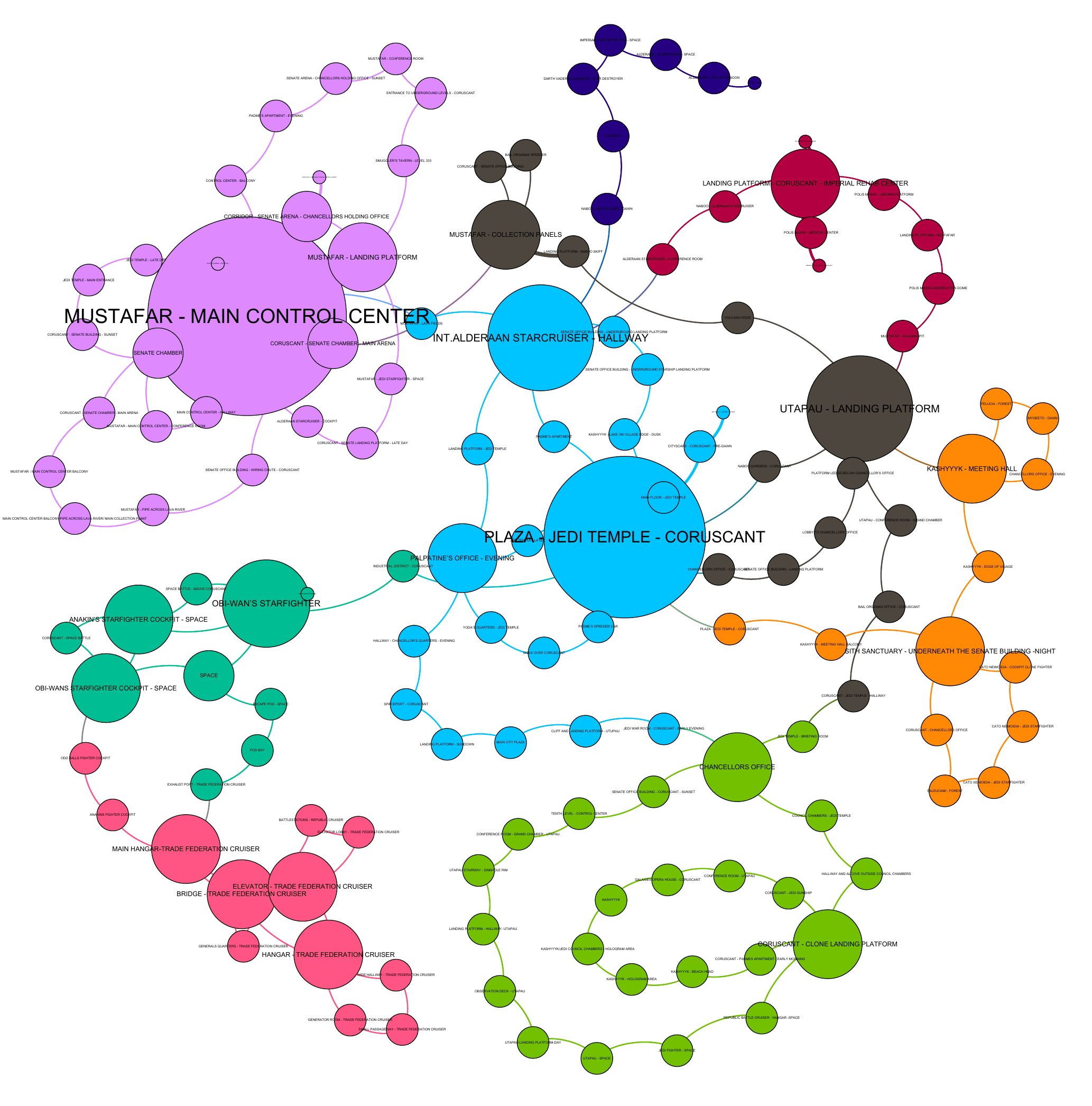}
            \caption[  $G_{LL}$]{$G_{LL}$} 
            
            \label{fig:SW3L}

        \end{subfigure}

\caption{The networks are better seen zoomed on the digital version of this \newline document. Visualization of communities in different layers of Episode III - Revenge \newline of the Sith (2005)~\cite{starwars2005episode}. The size of each node corresponds to its degree. (a) The \newline character layer $G_{CC}$. (b) The keyword layer $G_{KK}$. (c) The location layer $G_{LL}$.}
\label{fig:communities00}
\end{figure}

\begin{figure}[pht]
\centering
        \begin{subfigure}[b]{0.48\textwidth}   
            \centering 
            \includegraphics[width=\textwidth]{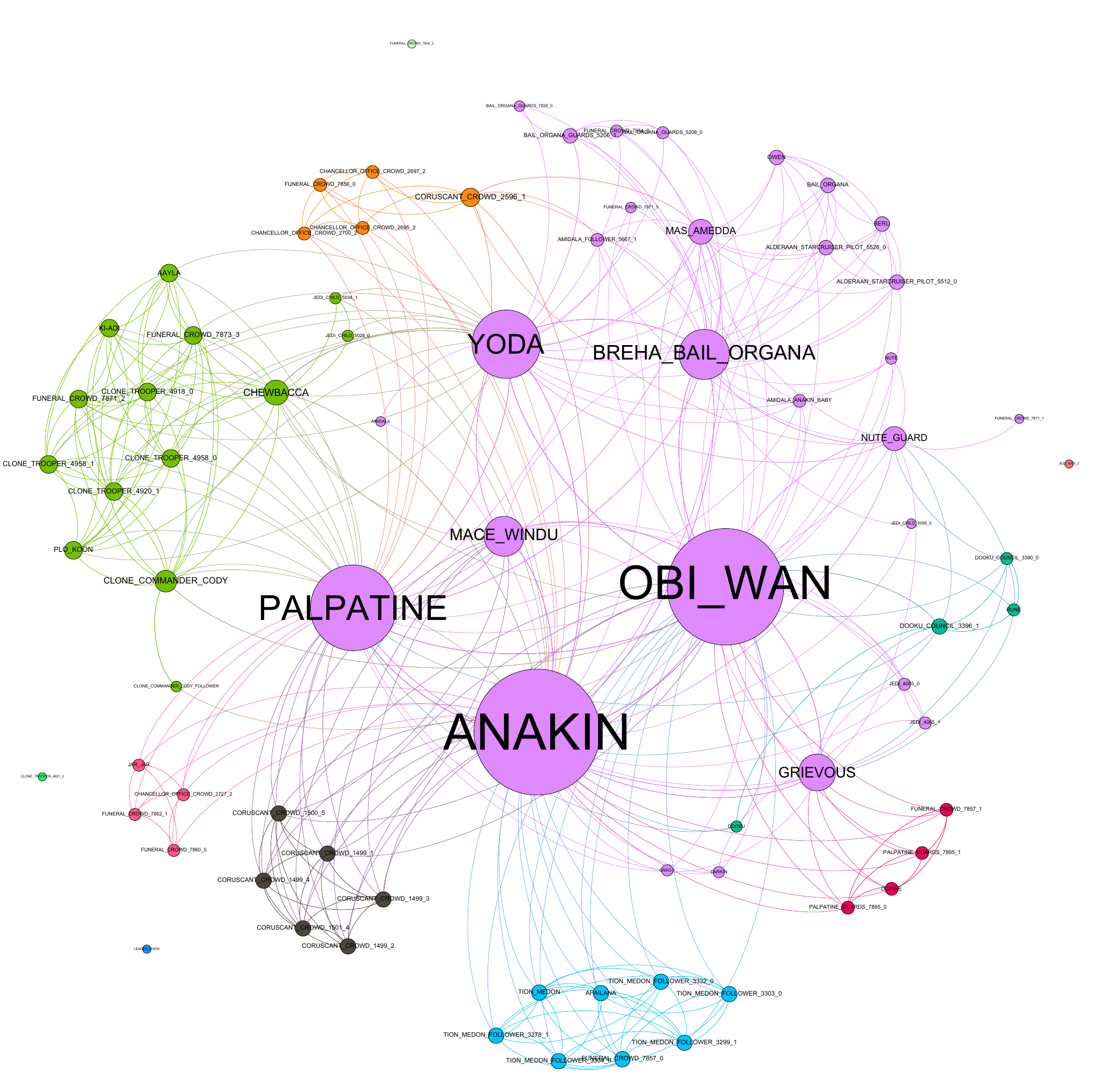}
            \caption[  $G_{FF}$]{$G_{FF}$}
            
            \label{fig:SW3F}
        \end{subfigure}
        \vskip\baselineskip  
        \begin{subfigure}[b]{0.475\textwidth}   
            \centering 
            \includegraphics[width=\textwidth]{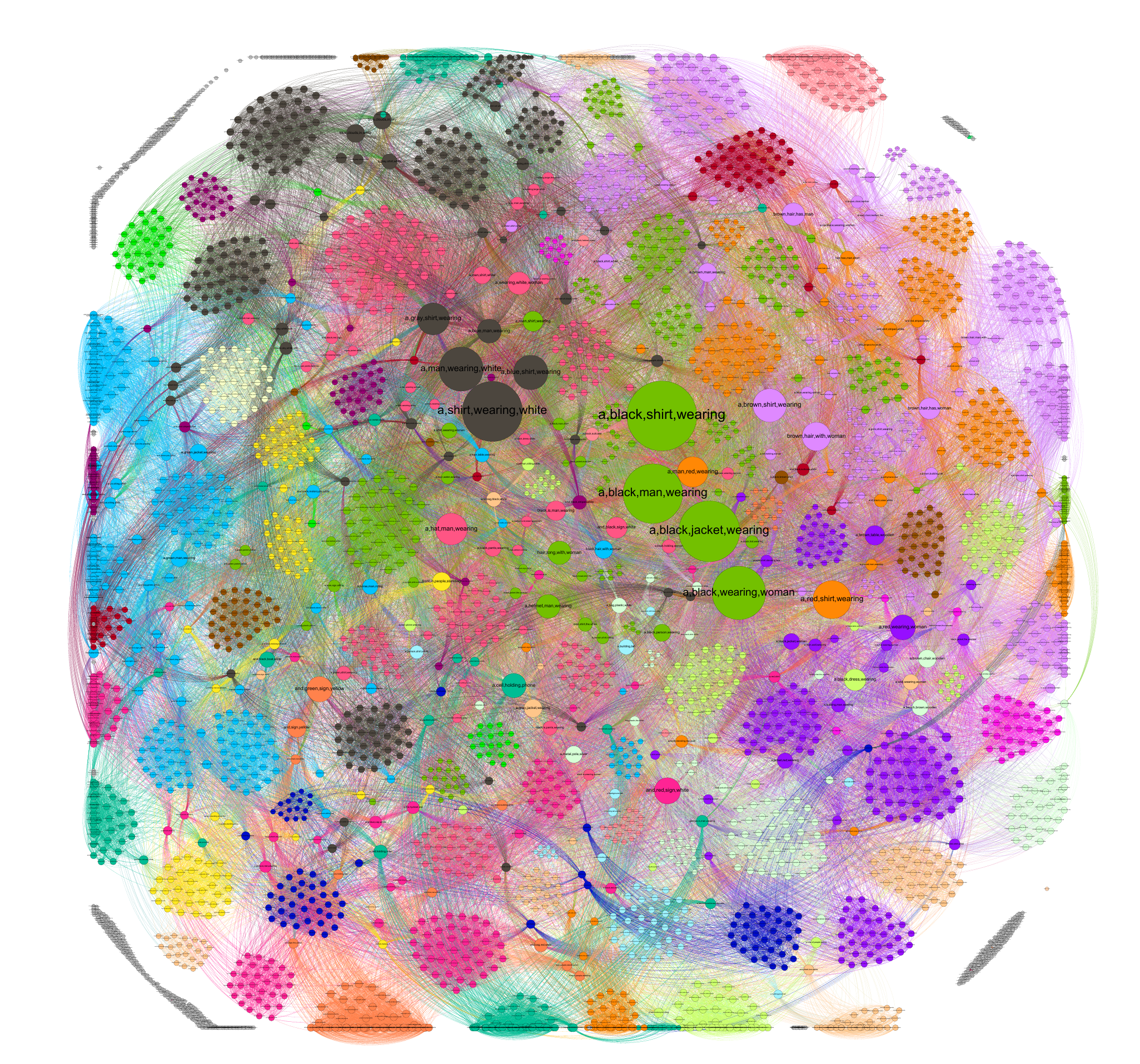}
            \caption[  $G_{CaCa}$]{$G_{CaCa}$}
            
            \label{fig:SW3CA}
        \end{subfigure}
        \qquad 
        \begin{subfigure}[b]{0.475\textwidth}   
            \centering 
            \includegraphics[width=\textwidth]{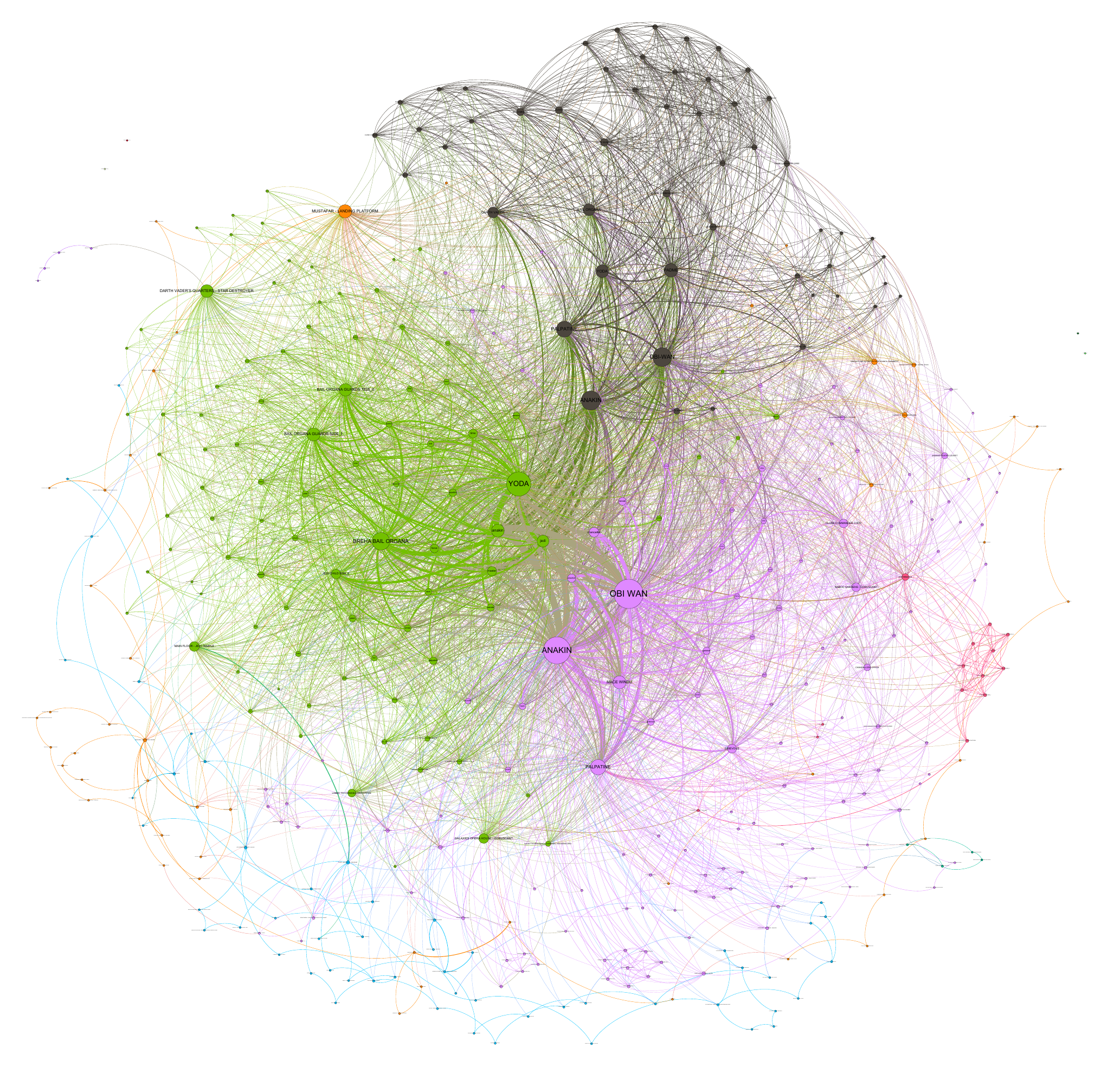}
            \caption[  $\mathbb G'$] {$\mathbb G'$}
            
            \label{fig:SW3M}
        \end{subfigure}

\caption{The networks are better seen zoomed on the digital version of this \newline document. Visualization of communities in different layers of Episode III - Revenge \newline of the Sith (2005)~\cite{starwars2005episode}. The size of each node corresponds to its degree. (a) The face \newline layer $G_{FF}$. (b) The caption layer $G_{CaCa}$. (c) The multilayer without captions $\mathbb G'$, with \newline the node label encoding: CHARACTER(C), FACE(F), keyword,  and LOCATION-.}
\label{fig:communities01}
\end{figure}

\subsection{Community detection}
Our preliminary results on global topological properties in Section~\ref{sec:topology} suggest the existence of communities especially given the clustering coefficient of the different layers \cite{gunce4}. To study clustering in the individual layers and the overall network, we use the modularity-based~\cite{girvan2002community} community detection algorithm often referred to as the Louvain method~\cite{blondel2008fast}, which has been generalized to multilayer networks too~\cite{domenico2014multilayer}.
Figure \ref{fig:modularity} reports the number of communities with the modularity per layer for each movie of the saga. Not surprisingly, the captions have the highest number of communities and highest modularity due to their definition which are cliques on each scene. It is however more surprising to see a high modularity for locations. Keywords best clusterize during SW5. Character and faces layers are social networks, displaying some potential for clustering. Captions also have a very high modularity, due to their nature as a collection of cliques.
Despite receiving a strong influence from the caption layer with comparable number of communities, the multilayer graph $\mathbb G$ shows overall modularity close to the keywords and faces. Without the caption layer, the multilayer graph $\mathbb G'$ seems very close to the community structures induced by faces, association to locations through cut order of the movie probably reinforces the importance of face co-occurrences.

The third episode of the prequel trilogy~\cite{starwars2005episode} is an interesting point in the series, where we can observe the main character of the whole saga, Anakin, turning into the dark version of himself that will be known as Darth Vader. We will observe how the different communities we measure may reflect this division. Communities of this episode are illustrated in Figure~\ref{fig:communities00} and ~\ref{fig:communities01} and with Gephi~\cite{heymann2014gephi} for SW3 only, all other episodes are also illustrated in the supplementary materials.

Starting with the character layer, we may notice three major communities. One community (pink) is centred around \textit{Padme} and \textit{Obi-Wan} and would correspond to the Jedi council that is represented by \textit{Yoda}, \textit{Mace Windu}, \textit{Ki-Adi}, together with the clone army they are leading, represented by \textit{Clone Commander Cody}. Their antagonist, \textit{General Grevious} is also put in this community, because one major plot of this episode is the fight of the Jedi against Grevious. 
A second community (green) is centred on politics and revolves around the senate on Coruscant, with \textit{Bail Organa}, and \textit{Mas Amedda}. A last community (purple) regroups the Sith side, with the major characters \textit{Palpatine} and \textit{Darth Vader}. 

On the contrary, the face layer does not make the distinction between Anakin and Vader. It shows 7 communities, with the main one (purple) formed from the main actors who are constantly interacting during the movie (\textit{Obi-Wan}, \textit{Anakin}, \textit{Palpatine}, \textit{Yoda}, \textit{Mace Windu}, \emph{etc.}). Other communities are formed around secondary characters or crowds such as the clone army together with \textit{Cody}. We also find as smaller tight communities such the Jedi council as a community, Coruscant politicians, crowds and followers. These minor characters are often presented together in one same scene creating such cliques.

Although the location layer gets a total of 10 communities, a few stand out. The locations are often connected by geographical proximity, as a sequence of scenes will follow a particular character or action that evolve in a small, continuous environment. On a larger scale, this is the temporal proximity that emerges. Sequences of events taking place at the same time but in different places connect the related locations.
In particular, one community (purple) relates to the end of the film. At this point, the action is concentrated on the duel between Anakin and Obi-Wan on \textit{Mustafar} and the one between Yoda and The Emperor at the \emph{Senate} and \emph{Palpatine's office}. The \emph{Mustafar main control center} is one key location of the fight but is also cut while Jedis are shown being executed by clones all across the galaxy, and Anakin is killing the last separatist leaders. This community also includes the \emph{Alderaan starcruiser}, the protagonists last stand at the end of the movie.
Another community (green) consists of locations used to showcase the battle at the beginning of the movie in \emph{space} while cutting to the inside of \emph{Obi-Wan's starfighter cockpit} as well as \emph{Anakin's starfighter cockpit}. In the film, once they localize the \emph{Trade Federation cruiser} where \emph{Palpatine} is held hostage, they head inside. We can see this transition occur via the \emph{hangar} of the ship. The next community exposes the inside of the cruiser, such as the \emph{bridge} and the \emph{elevator} that lead the protagonists to the \emph{Senator's room} and eventually \emph{General quarters}. At the end of their confrontation, General Grievous escapes through the \emph{pod bay}, returning the action to \emph{space}. The sequence ends with Anakin navigating a damaged ship through the \emph{skies above Coruscant}.
From this point on, the characters go on different adventure which is why the other communities are not as geographically focused. Yoda is on \emph{Kashyyyk}, Obi-Wan goes to \emph{Utapau} and Anakin remains on \emph{Coruscant}.

The keyword layer presents 10 communities corresponding to different topics. The largest community (light green) may be related to Anakin's emotional journey with words such as  \textit{anakin}, \textit{kill}, \textit{padme}, \textit{obiwan}, \textit{love}, \textit{destroy}, \textit{save}, \textit{lost}, \textit{etc.}
A second community (pink) groups around the political intrigue with \textit{jedi}, \textit{chancellor}, \textit{senate}, \textit{dooku}, \textit{etc.}
Confirming our observations on the character layer, another community (purple) is on the organization of the Jedi council
\textit{master}, \textit{kenobi}, \textit{windu}, \textit{etc.}, and of course another one is focused on the dark side with \textit{force}, \textit{power}, \textit{sith}, \textit{apprentice}, \textit{darth}, \textit{etc.}

Captions are clustered by scene in a large number of communities. Each scene has a number of captions 
which describe what happened in this scene. Observing communities does not offer much more 
interpretation beyond the colors clothing community. Since it impacts a lot the multilayer structure, 
we are more interesting in observing communities in the multilayer network $\mathbb G'$ that excludes 
this layer. There is a total of 12 communities. Four major communities regroup from 52 to 140 nodes, 
with very little overlaps between layers. In a first community (green), we have 52 main and secondary 
characters (all from $G_{CC}$) interacting together during the movie. In a second community (light 
green), we have 77 locations mostly from the end of the movie, with a handful of keywords related to 
the last dual (\textit{fight}, \textit{late}, \textit{inside}, \textit{chamber}, \textit{burning}, 
\textit{koon}), and two extra characters. In a community (purple) of 86 nodes that combines all layers 
and regroups vocabulary attached to the force from both Sith and Jedi sides (\textit{e.g.} 
\textit{master}, \textit{force}, \textit{afraid}, \textit{feel}, \textit{great}, \textit{lord}, 
\textit{powerful}, \textit{order}, \textit{dark}, \textit{control}, \textit{strong}, \textit{anger}, 
\textit{etc.}) and the locations where Anakin is turned \textit{Opera} and \textit{Lobby to 
Chancellor's Office}. A last community of 140 nodes also regroup most layers, with just a little bit 
of characters, a lot of faces of people in situation with battles and crowds, with people from crowds, 
such as \textit{Obi-Wan}, \textit{Grevious}, \textit{Cody}, \textit{etc.} The locations are very 
varied, and the vocabulary attached tends to be more technical of battles, including 
\textit{droids, clones, contact, move, platform, hold, attack, break, hangar, squad, commander, troops, escape, fire, mission, surface, front, engage, missiles, fighter,} \textit{etc.}
All in all, we can see a difference between the last two major communities that underline the two worlds, centred on Anakin, and that clash at the end of the movie. One is closer to the world of Padme/Amidala, with the senate politics and Organa, the other is closer to the Palpatine side, fights and adventure. The main reason might be the very little interactions between Anakin and Organa on one side, and between Padme and Palpatine on the other side.

\section{Conclusion}
\label{sec:conclusion}
In this paper, we introduce a multilayer model with movie elements  \emph{characters}, \emph{locations}, \emph{keywords}, \emph{faces} and \emph{captions} are in interaction. Unlike single layer networks which usually focus only on \emph{characters} or \emph{scenes}, this model is much more informative. It completes the single character network analysis with a new topological analysis made of more semantic elements that brings us a global broad picture of the movie story. We also propose an automatic method to extract the multilayer network elements from the script, subtitles, and movie content. In order to enrich the previous model, additional multimedia elements are included, such as face recognition, dense captioning and subtitle information. We have publicly released all our multilayer network datasets and made them available at {\small \url{github.com/UCEFM/Multilayer-Networks-Data-Star-Wars-Saga}.

On a model side, we have not fully discussed another contribution of Kivel\"{a}'s model~\cite{kivela2014multilayer} which are \textit{aspects}. Aspects could be understood as another discrete dimension of the multilayer network model, and this completely captures the notion of time depicted by the different episodes of the saga. In addition, one could consider furthermore the media modality from which we extract information to be another \textit{aspect} dimension, this is actually, what we are doing when separating the faces network from the character network. In future work, we will focus on questioning the coupling across these aspects.

So far we have not proposed any fusion of nodes through layers, such as face and characters, but we considered them separately, especially since some characters correspond to different personas (Anakin/Vader, Padme/Amidala/Doppelgangers). This alignment will show its usefulness in further studies. The locations are typically hierarchical in the way they are depicted (\textit{e.g.} \textit{planet - location - room}) and would deserve further treatment. This will be necessary to propose one full analysis at the level of the 6 movies taken at once.

We have deployed the model on the popular 6-movies of the Star Wars saga. Results of a brief analysis of the extracted networks confirmed the effectiveness of the model. So far, we have considered the succession of scenes to be the time granularity. We may however extend this notion and attempt to recover time as represented \emph{in} the movie world. This will require more complex processing of the events in the movie, and would help untangle complex movies like \textit{Memento} or \textit{Pulp Fiction} which have complex timelines, or like \textit{the Lord of the Rings} which has many parallel plots. It could be used as a support to study the location of characters along the plot and to enable a better transition between places: imagine a plot divided into multiple parts with parallel actions, we wish to recover this parallel nature (currently the location network may only form looping chains by definition). Note that much more information can be gained by a deeper topological analysis, for example, deriving a co-occurrence network of characters in the same location, a directed network of conversations, or mention of characters, \emph{etc.} As for the time granularity, we wish to get done to the level of shots and even seconds, to help deploy dynamic analysis. Our future work will also include a larger set of multilayer dedicated metrics, such as node entanglement~\cite{renoust2014entanglement}, and centrality measures designed for modular networks~\cite{comm,notre3}. Furthermore, in the future, we plan to deploy our tool on larger collections, such as tv-series, or even a larger collection of movies so we may obtain a higher view at collection level of artistic styles~\cite{sigaki2018history}.

Apart from movie representation for network analysis purposes, we believe that the model opens a numerous of new research directions. Indeed, it can also be used to characterize movie genres, or directors, and even correlate with acting careers from public databases such as IMDB. Furthermore, we can imagine automatically generate the movie trailer by searching important scenes where all movie characters are present. We also are working on including another layer to this multilayer network through emotions, which could help characterize characters and movie genres. Other layers from different media are left so far to explore, such as the actual sound component, the DVD chapter decomposition, and even language comparison if we consider different languages of the subtitle tracks. Fusing all sources of information like the proposed model does should come handy in supporting machine learning tasks, such as face recognizers, and movie  classification~\cite{gorinski2018s,viard2018movie}.

\nocite{*}


\section*{Availability of data and materials}
Not applicable.

\section*{Competing interests}
The authors declare that they have no competing interests.
  
\section*{Funding}
Not applicable.

\section*{Author's contributions}

YM is the main author of this paper, he has implemented most of the experiments and wrote the original draft . LV is responsible for the implementation regarding the face detection and tracking. OR has led the use case analysis. BR, HC, MEH designed the model, the framework and the experiments. BR participated to the experiments implementation, and the writing of the original draft. HC and MEH  did the review and editing of the first draft. They also proposed additional units of analysis. All the authors have read and approved the final manuscript.

\section*{Acknowledgements}
Not applicable.

\section*{Authors' information}
Not applicable.

\bibliographystyle{bmc-mathphys} 
\bibliography{7-bibliography}     


\begin{thebibliography}{78}
\ifx \bisbn   \undefined \def \bisbn  #1{ISBN #1}\fi
\ifx \binits  \undefined \def \binits#1{#1}\fi
\ifx \bauthor  \undefined \def \bauthor#1{#1}\fi
\ifx \batitle  \undefined \def \batitle#1{#1}\fi
\ifx \bjtitle  \undefined \def \bjtitle#1{#1}\fi
\ifx \bvolume  \undefined \def \bvolume#1{\textbf{#1}}\fi
\ifx \byear  \undefined \def \byear#1{#1}\fi
\ifx \bissue  \undefined \def \bissue#1{#1}\fi
\ifx \bfpage  \undefined \def \bfpage#1{#1}\fi
\ifx \blpage  \undefined \def \blpage #1{#1}\fi
\ifx \burl  \undefined \def \burl#1{\textsf{#1}}\fi
\ifx \doiurl  \undefined \def \doiurl#1{\textsf{#1}}\fi
\ifx \betal  \undefined \def \betal{\textit{et al.}}\fi
\ifx \binstitute  \undefined \def \binstitute#1{#1}\fi
\ifx \binstitutionaled  \undefined \def \binstitutionaled#1{#1}\fi
\ifx \bctitle  \undefined \def \bctitle#1{#1}\fi
\ifx \beditor  \undefined \def \beditor#1{#1}\fi
\ifx \bpublisher  \undefined \def \bpublisher#1{#1}\fi
\ifx \bbtitle  \undefined \def \bbtitle#1{#1}\fi
\ifx \bedition  \undefined \def \bedition#1{#1}\fi
\ifx \bseriesno  \undefined \def \bseriesno#1{#1}\fi
\ifx \blocation  \undefined \def \blocation#1{#1}\fi
\ifx \bsertitle  \undefined \def \bsertitle#1{#1}\fi
\ifx \bsnm \undefined \def \bsnm#1{#1}\fi
\ifx \bsuffix \undefined \def \bsuffix#1{#1}\fi
\ifx \bparticle \undefined \def \bparticle#1{#1}\fi
\ifx \barticle \undefined \def \barticle#1{#1}\fi
\ifx \bconfdate \undefined \def \bconfdate #1{#1}\fi
\ifx \botherref \undefined \def \botherref #1{#1}\fi
\ifx \url \undefined \def \url#1{\textsf{#1}}\fi
\ifx \bchapter \undefined \def \bchapter#1{#1}\fi
\ifx \bbook \undefined \def \bbook#1{#1}\fi
\ifx \bcomment \undefined \def \bcomment#1{#1}\fi
\ifx \oauthor \undefined \def \oauthor#1{#1}\fi
\ifx \citeauthoryear \undefined \def \citeauthoryear#1{#1}\fi
\ifx \endbibitem  \undefined \def \endbibitem {}\fi
\ifx \bconflocation  \undefined \def \bconflocation#1{#1}\fi
\ifx \arxivurl  \undefined \def \arxivurl#1{\textsf{#1}}\fi
\csname PreBibitemsHook\endcsname

\bibitem{rital2005weighted}
\begin{bchapter}
\bauthor{\bsnm{Rital}, \binits{S.}},
\bauthor{\bsnm{Cherifi}, \binits{H.}},
\bauthor{\bsnm{Miguet}, \binits{S.}}:
\bctitle{Weighted adaptive neighborhood hypergraph partitioning for image
  segmentation}.
In: \bbtitle{International Conference on Pattern Recognition and Image
  Analysis},
pp. \bfpage{522}--\blpage{531}
(\byear{2005}).
\bcomment{Springer}
\end{bchapter}
\endbibitem

\bibitem{park2012social}
\begin{barticle}
\bauthor{\bsnm{Park}, \binits{S.-B.}},
\bauthor{\bsnm{Oh}, \binits{K.-J.}},
\bauthor{\bsnm{Jo}, \binits{G.-S.}}:
\batitle{Social network analysis in a movie using character-net}.
\bjtitle{Multimedia Tools and Applications}
\bvolume{59}(\bissue{2}),
\bfpage{601}--\blpage{627}
(\byear{2012})
\end{barticle}
\endbibitem

\bibitem{waumans2015topology}
\begin{barticle}
\bauthor{\bsnm{Waumans}, \binits{M.C.}},
\bauthor{\bsnm{Nicod{\`e}me}, \binits{T.}},
\bauthor{\bsnm{Bersini}, \binits{H.}}:
\batitle{Topology analysis of social networks extracted from literature}.
\bjtitle{PloS one}
\bvolume{10}(\bissue{6}),
\bfpage{0126470}
(\byear{2015})
\end{barticle}
\endbibitem

\bibitem{tan2014character}
\begin{bchapter}
\bauthor{\bsnm{Tan}, \binits{M.S.}},
\bauthor{\bsnm{Ujum}, \binits{E.A.}},
\bauthor{\bsnm{Ratnavelu}, \binits{K.}}:
\bctitle{A character network study of two sci-fi tv series},
vol. \bseriesno{1588},
pp. \bfpage{246}--\blpage{251}
(\byear{2014}).
\bcomment{AIP}
\end{bchapter}
\endbibitem

\bibitem{renoust2015social}
\begin{barticle}
\bauthor{\bsnm{Renoust}, \binits{B.}},
\bauthor{\bsnm{Kobayashi}, \binits{T.}},
\bauthor{\bsnm{Ngo}, \binits{T.D.}},
\bauthor{\bsnm{Le}, \binits{D.-D.}},
\bauthor{\bsnm{Satoh}, \binits{S.}}:
\batitle{When face-tracking meets social networks: a story of politics in news
  videos}.
\bjtitle{Applied Network Science}
\bvolume{1}(\bissue{1}),
\bfpage{4}
(\byear{2016})
\end{barticle}
\endbibitem

\bibitem{renoust2016visual}
\begin{barticle}
\bauthor{\bsnm{Renoust}, \binits{B.}},
\bauthor{\bsnm{Le}, \binits{D.-D.}},
\bauthor{\bsnm{Satoh}, \binits{S.}}:
\batitle{Visual analytics of political networks from face-tracking of news
  video}.
\bjtitle{IEEE Transactions on Multimedia}
\bvolume{18}(\bissue{11}),
\bfpage{2184}--\blpage{2195}
(\byear{2016})
\end{barticle}
\endbibitem

\bibitem{mish2016game}
\begin{botherref}
\oauthor{\bsnm{Mish}, \binits{B.}}:
Game of Nodes: A Social Network Analysis of Game of Thrones.
https://gameofnodes.wordpress.com
(2016)
\end{botherref}
\endbibitem

\bibitem{mourchid_multilayer18}
\begin{bchapter}
\bauthor{\bsnm{Mourchid}, \binits{Y.}},
\bauthor{\bsnm{Renoust}, \binits{B.}},
\bauthor{\bsnm{Cherifi}, \binits{H.}},
\bauthor{\bsnm{El~Hassouni}, \binits{M.}}:
\bctitle{Multilayer network model of movie script},
pp. \bfpage{782}--\blpage{796}
(\byear{2018}).
\bcomment{Springer}
\end{bchapter}
\endbibitem

\bibitem{viard2018movie}
\begin{botherref}
\oauthor{\bsnm{Viard}, \binits{T.}},
\oauthor{\bsnm{Fournier{-}S'niehotta}, \binits{R.}}:
Movie rating prediction using content-based and link stream features.
CoRR
\textbf{abs/1805.02893}
(2018).
\arxivurl{1805.02893}
\end{botherref}
\endbibitem

\bibitem{markovivc2018applying}
\begin{barticle}
\bauthor{\bsnm{Markovi{\v{c}}}, \binits{R.}},
\bauthor{\bsnm{Gosak}, \binits{M.}},
\bauthor{\bsnm{Perc}, \binits{M.}},
\bauthor{\bsnm{Marhl}, \binits{M.}},
\bauthor{\bsnm{Grubelnik}, \binits{V.}}:
\batitle{Applying network theory to fables: complexity in slovene
  belles-lettres for different age groups}.
\bjtitle{Journal of Complex Networks}
\bvolume{7}(\bissue{1}),
\bfpage{114}--\blpage{127}
(\byear{2018})
\end{barticle}
\endbibitem

\bibitem{chen2009novel}
\begin{barticle}
\bauthor{\bsnm{Chen}, \binits{B.-W.}},
\bauthor{\bsnm{Wang}, \binits{J.-C.}},
\bauthor{\bsnm{Wang}, \binits{J.-F.}}:
\batitle{A novel video summarization based on mining the story-structure and
  semantic relations among concept entities}.
\bjtitle{IEEE Transactions on Multimedia}
\bvolume{11}(\bissue{2}),
\bfpage{295}--\blpage{312}
(\byear{2009})
\end{barticle}
\endbibitem

\bibitem{kipling1998just}
\begin{botherref}
\oauthor{\bsnm{Kipling}}:
Just so stories for little children
(1909)
\end{botherref}
\endbibitem

\bibitem{kurzhals2016visual}
\begin{barticle}
\bauthor{\bsnm{Kurzhals}, \binits{K.}},
\bauthor{\bsnm{John}, \binits{M.}},
\bauthor{\bsnm{Heimerl}, \binits{F.}},
\bauthor{\bsnm{Kuznecov}, \binits{P.}},
\bauthor{\bsnm{Weiskopf}, \binits{D.}}:
\batitle{Visual movie analytics}.
\bjtitle{IEEE Transactions on Multimedia}
\bvolume{18}(\bissue{11}),
\bfpage{2149}--\blpage{2160}
(\byear{2016})
\end{barticle}
\endbibitem

\bibitem{sekara2016fundamental}
\begin{barticle}
\bauthor{\bsnm{Sekara}, \binits{V.}},
\bauthor{\bsnm{Stopczynski}, \binits{A.}},
\bauthor{\bsnm{Lehmann}, \binits{S.}}:
\batitle{Fundamental structures of dynamic social networks}.
\bjtitle{Proceedings of the national academy of sciences}
\bvolume{113}(\bissue{36}),
\bfpage{9977}--\blpage{9982}
(\byear{2016})
\end{barticle}
\endbibitem

\bibitem{renoust2014entanglement}
\begin{botherref}
\oauthor{\bsnm{Renoust}, \binits{B.}},
\oauthor{\bsnm{Melan{\c{c}}on}, \binits{G.}},
\oauthor{\bsnm{Viaud}, \binits{M.-L.}}:
Entanglement in multiplex networks: understanding group cohesion in homophily
  networks.
Social Network Analysis-Community Detection and Evolution,
89--117
(2014)
\end{botherref}
\endbibitem

\bibitem{bao2015recommendations}
\begin{barticle}
\bauthor{\bsnm{Bao}, \binits{J.}},
\bauthor{\bsnm{Zheng}, \binits{Y.}},
\bauthor{\bsnm{Wilkie}, \binits{D.}},
\bauthor{\bsnm{Mokbel}, \binits{M.}}:
\batitle{Recommendations in location-based social networks: a survey}.
\bjtitle{GeoInformatica}
\bvolume{19}(\bissue{3}),
\bfpage{525}--\blpage{565}
(\byear{2015})
\end{barticle}
\endbibitem

\bibitem{latapy2018stream}
\begin{barticle}
\bauthor{\bsnm{Latapy}, \binits{M.}},
\bauthor{\bsnm{Viard}, \binits{T.}},
\bauthor{\bsnm{Magnien}, \binits{C.}}:
\batitle{Stream graphs and link streams for the modeling of interactions over
  time}.
\bjtitle{Social Network Analysis and Mining}
\bvolume{8}(\bissue{1}),
\bfpage{61}
(\byear{2018})
\end{barticle}
\endbibitem

\bibitem{jhala2008exploiting}
\begin{botherref}
\oauthor{\bsnm{Jhala}, \binits{A.}}:
Exploiting structure and conventions of movie scripts for information retrieval
  and text mining.
Joint International Conference on Interactive Digital Storytelling,
210--213
(2008).
Springer
\end{botherref}
\endbibitem

\bibitem{guo2016deep}
\begin{barticle}
\bauthor{\bsnm{Guo}, \binits{Y.}},
\bauthor{\bsnm{Liu}, \binits{Y.}},
\bauthor{\bsnm{Oerlemans}, \binits{A.}},
\bauthor{\bsnm{Lao}, \binits{S.}},
\bauthor{\bsnm{Wu}, \binits{S.}},
\bauthor{\bsnm{Lew}, \binits{M.S.}}:
\batitle{Deep learning for visual understanding: A review}.
\bjtitle{Neurocomputing}
\bvolume{187},
\bfpage{27}--\blpage{48}
(\byear{2016})
\end{barticle}
\endbibitem

\bibitem{demirkesen2008comparison}
\begin{bchapter}
\bauthor{\bsnm{Demirkesen}, \binits{C.}},
\bauthor{\bsnm{Cherifi}, \binits{H.}}:
\bctitle{A comparison of multiclass svm methods for real world natural scenes}.
In: \bbtitle{International Conference on Advanced Concepts for Intelligent
  Vision Systems},
pp. \bfpage{752}--\blpage{763}
(\byear{2008}).
\bcomment{Springer}
\end{bchapter}
\endbibitem

\bibitem{pastrana2006predicting}
\begin{barticle}
\bauthor{\bsnm{Pastrana-Vidal}, \binits{R.R.}},
\bauthor{\bsnm{Gicquel}, \binits{J.C.}},
\bauthor{\bsnm{Blin}, \binits{J.L.}},
\bauthor{\bsnm{Cherifi}, \binits{H.}}:
\batitle{Predicting subjective video quality from separated spatial and
  temporal assessment}.
\bjtitle{Human Vision and Electronic Imaging XI}
\bvolume{6057},
\bfpage{60570}
(\byear{2006}).
\bcomment{International Society for Optics and Photonics}
\end{barticle}
\endbibitem

\bibitem{jiang2017face}
\begin{botherref}
\oauthor{\bsnm{Jiang}, \binits{H.}},
\oauthor{\bsnm{Learned-Miller}, \binits{E.}}:
Face detection with the faster r-cnn.
Automatic Face \& Gesture Recognition (FG 2017), 2017 12th IEEE International
  Conference on,
650--657
(2017).
IEEE
\end{botherref}
\endbibitem

\bibitem{cao2018vggface2}
\begin{botherref}
\oauthor{\bsnm{Cao}, \binits{Q.}},
\oauthor{\bsnm{Shen}, \binits{L.}},
\oauthor{\bsnm{Xie}, \binits{W.}},
\oauthor{\bsnm{Parkhi}, \binits{O.M.}},
\oauthor{\bsnm{Zisserman}, \binits{A.}}:
Vggface2: A dataset for recognising faces across pose and age.
Automatic Face \& Gesture Recognition (FG 2018), 2018 13th IEEE International
  Conference on,
67--74
(2018).
IEEE
\end{botherref}
\endbibitem

\bibitem{johnson2016densecap}
\begin{botherref}
\oauthor{\bsnm{Johnson}, \binits{J.}},
\oauthor{\bsnm{Karpathy}, \binits{A.}},
\oauthor{\bsnm{Fei-Fei}, \binits{L.}}:
Densecap: Fully convolutional localization networks for dense captioning.
Proceedings of the IEEE Conference on Computer Vision and Pattern Recognition,
4565--4574
(2016)
\end{botherref}
\endbibitem

\bibitem{yang2017dense}
\begin{botherref}
\oauthor{\bsnm{Yang}, \binits{L.}},
\oauthor{\bsnm{Tang}, \binits{K.}},
\oauthor{\bsnm{Yang}, \binits{J.}},
\oauthor{\bsnm{Li}, \binits{L.-J.}}:
Dense captioning with joint inference and visual context.
Proceedings of the IEEE Conference on Computer Vision and Pattern Recognition
  (CVPR)
\textbf{2}
(2017)
\end{botherref}
\endbibitem

\bibitem{domenico2014multilayer}
\begin{botherref}
\oauthor{\bsnm{Domenico}, \binits{M.}},
\oauthor{\bsnm{Porter}, \binits{M.}},
\oauthor{\bsnm{Arenas}, \binits{A.}}:
Multilayer analysis and visualization of networks.
J. Complex Netw
\textbf{10}
(2014)
\end{botherref}
\endbibitem

\bibitem{kivela2014multilayer}
\begin{barticle}
\bauthor{\bsnm{Kivel{\"a}}, \binits{M.}},
\bauthor{\bsnm{Arenas}, \binits{A.}},
\bauthor{\bsnm{Barthelemy}, \binits{M.}},
\bauthor{\bsnm{Gleeson}, \binits{J.P.}},
\bauthor{\bsnm{Moreno}, \binits{Y.}},
\bauthor{\bsnm{Porter}, \binits{M.A.}}:
\batitle{Multilayer networks}.
\bjtitle{Journal of complex networks}
\bvolume{2}(\bissue{3}),
\bfpage{203}--\blpage{271}
(\byear{2014})
\end{barticle}
\endbibitem

\bibitem{starwars1977episode}
\begin{botherref}
\oauthor{\bsnm{Lucas}, \binits{G.}}:
{Star Wars: Episode IV - A New Hope}.
Twentieth Century Fox Film Corporation
(1977)
\end{botherref}
\endbibitem

\bibitem{starwars1980episode}
\begin{botherref}
\oauthor{\bsnm{Lucas}, \binits{G.}}:
{Star Wars: Episode V - The Empire Strikes Back}.
Twentieth Century Fox Film Corporation
(1980)
\end{botherref}
\endbibitem

\bibitem{starwars1983episode}
\begin{botherref}
\oauthor{\bsnm{Lucas}, \binits{G.}}:
{Star Wars: Episode VI - Return of the Jedi}.
Twentieth Century Fox Film Corporation
(1983)
\end{botherref}
\endbibitem

\bibitem{starwars1999episode}
\begin{botherref}
\oauthor{\bsnm{Lucas}, \binits{G.}}:
{Star Wars: Episode I - The Phantom Menace}.
Twentieth Century Fox Film Corporation
(1999)
\end{botherref}
\endbibitem

\bibitem{starwars2002episode}
\begin{botherref}
\oauthor{\bsnm{Lucas}, \binits{G.}}:
{Star Wars: Episode II - Attack of the Clones}.
Twentieth Century Fox Film Corporation
(2002)
\end{botherref}
\endbibitem

\bibitem{starwars2005episode}
\begin{botherref}
\oauthor{\bsnm{Lucas}, \binits{G.}}:
{Star Wars: Episode III - Revenge of the Sith}.
Twentieth Century Fox Film Corporation
(2005)
\end{botherref}
\endbibitem

\bibitem{kadushin2012understanding}
\begin{botherref}
\oauthor{\bsnm{Kadushin}, \binits{C.}}:
Understanding social networks: Theories, concepts, and findings
(2012)
\end{botherref}
\endbibitem

\bibitem{yeung1996extracting}
\begin{botherref}
\oauthor{\bsnm{Yeung}, \binits{M.}},
\oauthor{\bsnm{Yeo}, \binits{B.-L.}},
\oauthor{\bsnm{Liu}, \binits{B.}}:
Extracting story units from long programs for video browsing and navigation.
Multimedia Computing and Systems, 1996., Proceedings of the Third IEEE
  International Conference on,
296--305
(1996).
IEEE
\end{botherref}
\endbibitem

\bibitem{jung2004narrative}
\begin{botherref}
\oauthor{\bsnm{Jung}, \binits{B.}},
\oauthor{\bsnm{Kwak}, \binits{T.}},
\oauthor{\bsnm{Song}, \binits{J.}},
\oauthor{\bsnm{Lee}, \binits{Y.}}:
Narrative abstraction model for story-oriented video.
Proceedings of the 12th annual ACM international conference on Multimedia,
828--835
(2004).
ACM
\end{botherref}
\endbibitem

\bibitem{correa2019semantic}
\begin{botherref}
\oauthor{\bsnm{Jr.}, \binits{E.A.C.}},
\oauthor{\bsnm{Marinho}, \binits{V.Q.}},
\oauthor{\bsnm{Amancio}, \binits{D.R.}}:
Semantic flow in language networks.
CoRR
\textbf{abs/1905.07595}
(2019).
\arxivurl{1905.07595}
\end{botherref}
\endbibitem

\bibitem{knuth1993stanford}
\begin{botherref}
\oauthor{\bsnm{Knuth}, \binits{D.E.}}:
The stanford graphbase: a platform for combinatorial computing.
AcM Press New York
(1993)
\end{botherref}
\endbibitem

\bibitem{chen2019unsupervised}
\begin{barticle}
\bauthor{\bsnm{Chen}, \binits{R.-G.}},
\bauthor{\bsnm{Chen}, \binits{C.-C.}},
\bauthor{\bsnm{Chen}, \binits{C.-M.}}:
\batitle{Unsupervised cluster analyses of character networks in fiction:
  Community structure and centrality}.
\bjtitle{Knowledge-Based Systems}
\bvolume{163},
\bfpage{800}--\blpage{810}
(\byear{2019})
\end{barticle}
\endbibitem

\bibitem{weng2009rolenet}
\begin{barticle}
\bauthor{\bsnm{Weng}, \binits{C.-Y.}},
\bauthor{\bsnm{Chu}, \binits{W.-T.}},
\bauthor{\bsnm{Wu}, \binits{J.-L.}}:
\batitle{Rolenet: Movie analysis from the perspective of social networks}.
\bjtitle{IEEE Transactions on Multimedia}
\bvolume{11}(\bissue{2}),
\bfpage{256}--\blpage{271}
(\byear{2009})
\end{barticle}
\endbibitem

\bibitem{tran2015cocharnet}
\begin{barticle}
\bauthor{\bsnm{Tran}, \binits{Q.D.}},
\bauthor{\bsnm{Jung}, \binits{J.E.}}:
\batitle{Cocharnet: Extracting social networks using character co-occurrence in
  movies.}
\bjtitle{J. UCS}
\bvolume{21}(\bissue{6}),
\bfpage{796}--\blpage{815}
(\byear{2015})
\end{barticle}
\endbibitem

\bibitem{he2018srn}
\begin{botherref}
\oauthor{\bsnm{He}, \binits{J.}},
\oauthor{\bsnm{Xie}, \binits{Y.}},
\oauthor{\bsnm{Luan}, \binits{X.}},
\oauthor{\bsnm{Zhang}, \binits{L.}},
\oauthor{\bsnm{Zhang}, \binits{X.}}:
Srn: The movie character relationship analysis via social network.
International Conference on Multimedia Modeling,
289--301
(2018).
Springer
\end{botherref}
\endbibitem

\bibitem{gorinski2018s}
\begin{botherref}
\oauthor{\bsnm{Gorinski}, \binits{P.J.}},
\oauthor{\bsnm{Lapata}, \binits{M.}}:
What’s this movie about? a joint neural network architecture for movie
  content analysis.
Proceedings of the 2018 Conference of the North American Chapter of the
  Association for Computational Linguistics: Human Language Technologies,
  Volume 1 (Long Papers),
1770--1781
(2018)
\end{botherref}
\endbibitem

\bibitem{lv2018storyrolenet}
\begin{barticle}
\bauthor{\bsnm{Lv}, \binits{J.}},
\bauthor{\bsnm{Wu}, \binits{B.}},
\bauthor{\bsnm{Zhou}, \binits{L.}},
\bauthor{\bsnm{Wang}, \binits{H.}}:
\batitle{Storyrolenet: Social network construction of role relationship in
  video}.
\bjtitle{IEEE Access}
\bvolume{6},
\bfpage{25958}--\blpage{25969}
(\byear{2018})
\end{barticle}
\endbibitem

\bibitem{ren2018generating}
\begin{botherref}
\oauthor{\bsnm{Ren}, \binits{H.}},
\oauthor{\bsnm{Renoust}, \binits{B.}},
\oauthor{\bsnm{Viaud}, \binits{M.-L.}},
\oauthor{\bsnm{Melan{\c{c}}on}, \binits{G.}},
\oauthor{\bsnm{Satoh}, \binits{S.}}:
Generating “visual clouds” from multiplex networks for tv news archive
  query visualization.
2018 International Conference on Content-Based Multimedia Indexing (CBMI),
1--6
(2018).
IEEE
\end{botherref}
\endbibitem

\bibitem{flint1917newspaper}
\begin{botherref}
\oauthor{\bsnm{Flint}, \binits{L.N.}}:
Newspaper writing in high schools: Containing an outline for the use of
  teachers
(1917)
\end{botherref}
\endbibitem

\bibitem{nadeau2007survey}
\begin{barticle}
\bauthor{\bsnm{Nadeau}, \binits{D.}},
\bauthor{\bsnm{Sekine}, \binits{S.}}:
\batitle{A survey of named entity recognition and classification}.
\bjtitle{Lingvisticae Investigationes}
\bvolume{30}(\bissue{1}),
\bfpage{3}--\blpage{26}
(\byear{2007})
\end{barticle}
\endbibitem

\bibitem{al2017choosing}
\begin{botherref}
\oauthor{\bsnm{Al~Omran}, \binits{F.N.A.}},
\oauthor{\bsnm{Treude}, \binits{C.}}:
Choosing an nlp library for analyzing software documentation: a systematic
  literature review and a series of experiments.
Proceedings of the 14th International Conference on Mining Software
  Repositories,
187--197
(2017).
IEEE Press
\end{botherref}
\endbibitem

\bibitem{salton1975vector}
\begin{barticle}
\bauthor{\bsnm{Salton}, \binits{G.}},
\bauthor{\bsnm{Wong}, \binits{A.}},
\bauthor{\bsnm{Yang}, \binits{C.-S.}}:
\batitle{A vector space model for automatic indexing}.
\bjtitle{Communications of the ACM}
\bvolume{18}(\bissue{11}),
\bfpage{613}--\blpage{620}
(\byear{1975})
\end{barticle}
\endbibitem

\bibitem{li2007keyword}
\begin{barticle}
\bauthor{\bsnm{Li}, \binits{J.}},
\bauthor{\bsnm{Zhang}, \binits{K.}}, \betal:
\batitle{Keyword extraction based on tf/idf for chinese news document}.
\bjtitle{Wuhan University Journal of Natural Sciences}
\bvolume{12}(\bissue{5}),
\bfpage{917}--\blpage{921}
(\byear{2007})
\end{barticle}
\endbibitem

\bibitem{blei2003latent}
\begin{barticle}
\bauthor{\bsnm{Blei}, \binits{D.M.}},
\bauthor{\bsnm{Ng}, \binits{A.Y.}},
\bauthor{\bsnm{Jordan}, \binits{M.I.}}:
\batitle{Latent dirichlet allocation}.
\bjtitle{Journal of machine Learning research}
\bvolume{3}(\bissue{Jan}),
\bfpage{993}--\blpage{1022}
(\byear{2003})
\end{barticle}
\endbibitem

\bibitem{yuepeng2015keyword}
\begin{barticle}
\bauthor{\bsnm{Yuepeng}, \binits{L.}},
\bauthor{\bsnm{Cui}, \binits{J.}},
\bauthor{\bsnm{Junchuan}, \binits{J.}}:
\batitle{A keyword extraction algorithm based on word2vec}.
\bjtitle{e-Science Technology \& Application}
\bvolume{4},
\bfpage{54}--\blpage{59}
(\byear{2015})
\end{barticle}
\endbibitem

\bibitem{castellano2012pyscenedetect}
\begin{botherref}
\oauthor{\bsnm{Castellano}, \binits{B.}}:
{PySceneDetect}.
Last accessed: 2019-06-20
(2012).
\url{github.com/Breakthrough/PySceneDetect}
\end{botherref}
\endbibitem

\bibitem{yang2016wider}
\begin{botherref}
\oauthor{\bsnm{Yang}, \binits{S.}},
\oauthor{\bsnm{Luo}, \binits{P.}},
\oauthor{\bsnm{Loy}, \binits{C.C.}},
\oauthor{\bsnm{Tang}, \binits{X.}}:
Wider face: A face detection benchmark.
IEEE Conference on Computer Vision and Pattern Recognition (CVPR)
(2016)
\end{botherref}
\endbibitem

\bibitem{he2016deep}
\begin{botherref}
\oauthor{\bsnm{He}, \binits{K.}},
\oauthor{\bsnm{Zhang}, \binits{X.}},
\oauthor{\bsnm{Ren}, \binits{S.}},
\oauthor{\bsnm{Sun}, \binits{J.}}:
Deep residual learning for image recognition.
Proceedings of the IEEE conference on computer vision and pattern recognition,
770--778
(2016)
\end{botherref}
\endbibitem

\bibitem{gisbrecht2015parametric}
\begin{barticle}
\bauthor{\bsnm{Gisbrecht}, \binits{A.}},
\bauthor{\bsnm{Schulz}, \binits{A.}},
\bauthor{\bsnm{Hammer}, \binits{B.}}:
\batitle{Parametric nonlinear dimensionality reduction using kernel t-sne}.
\bjtitle{Neurocomputing}
\bvolume{147},
\bfpage{71}--\blpage{82}
(\byear{2015})
\end{barticle}
\endbibitem

\bibitem{auber2017tulip}
\begin{botherref}
\oauthor{\bsnm{Auber}, \binits{D.}},
\oauthor{\bsnm{Archambault}, \binits{D.}},
\oauthor{\bsnm{Bourqui}, \binits{R.}},
\oauthor{\bsnm{Delest}, \binits{M.}},
\oauthor{\bsnm{Dubois}, \binits{J.}},
\oauthor{\bsnm{Lambert}, \binits{A.}},
\oauthor{\bsnm{Mary}, \binits{P.}},
\oauthor{\bsnm{Mathiaut}, \binits{M.}},
\oauthor{\bsnm{M{\'e}lan{\c{c}}on}, \binits{G.}},
\oauthor{\bsnm{Pinaud}, \binits{B.}},
\oauthor{\bsnm{Renoust}, \binits{B.}},
\oauthor{\bsnm{Vallet}, \binits{J.}}:
Tulip 5,
1--28
(2017)
\end{botherref}
\endbibitem

\bibitem{ester1996density}
\begin{botherref}
\oauthor{\bsnm{Ester}, \binits{M.}},
\oauthor{\bsnm{Kriegel}, \binits{H.-P.}},
\oauthor{\bsnm{Sander}, \binits{J.}},
\oauthor{\bsnm{Xu}, \binits{X.}}:
Density-based spatial clustering of applications with noise.
Int. Conf. Knowledge Discovery and Data Mining
\textbf{240}
(1996)
\end{botherref}
\endbibitem

\bibitem{Krishna2017}
\begin{barticle}
\bauthor{\bsnm{Krishna}, \binits{R.}},
\bauthor{\bsnm{Zhu}, \binits{Y.}},
\bauthor{\bsnm{Groth}, \binits{O.}},
\bauthor{\bsnm{Johnson}, \binits{J.}},
\bauthor{\bsnm{Hata}, \binits{K.}},
\bauthor{\bsnm{Kravitz}, \binits{J.}},
\bauthor{\bsnm{Chen}, \binits{S.}},
\bauthor{\bsnm{Kalantidis}, \binits{Y.}},
\bauthor{\bsnm{Li}, \binits{L.-J.}},
\bauthor{\bsnm{Shamma}, \binits{D.A.}},
\bauthor{\bsnm{Bernstein}, \binits{M.S.}},
\bauthor{\bsnm{Fei-Fei}, \binits{L.}}:
\batitle{Visual genome: Connecting language and vision using crowdsourced dense
  image annotations}.
\bjtitle{International Journal of Computer Vision}
\bvolume{123}(\bissue{1}),
\bfpage{32}--\blpage{73}
(\byear{2017})
\end{barticle}
\endbibitem

\bibitem{cavnar1994n}
\begin{botherref}
\oauthor{\bsnm{Cavnar}, \binits{W.B.}},
\oauthor{\bsnm{Trenkle}, \binits{J.M.}}, et al.:
N-gram-based text categorization.
Proceedings of SDAIR-94, 3rd annual symposium on document analysis and
  information retrieval
\textbf{161175}
(1994).
Citeseer
\end{botherref}
\endbibitem

\bibitem{bioglio2017movie}
\begin{botherref}
\oauthor{\bsnm{Bioglio}, \binits{L.}},
\oauthor{\bsnm{Pensa}, \binits{R.G.}}:
Is this movie a milestone? identification of the most influential movies in the
  history of cinema.
International Workshop on Complex Networks and their Applications,
921--934
(2017).
Springer
\end{botherref}
\endbibitem

\bibitem{domenico2013centrality}
\begin{botherref}
\oauthor{\bsnm{Domenico}, \binits{M.}},
\oauthor{\bsnm{Sol-Ribalta}, \binits{A.}},
\oauthor{\bsnm{Omodei}, \binits{E.}},
\oauthor{\bsnm{Gmez}, \binits{S.}},
\oauthor{\bsnm{Arenas}, \binits{A.}}:
Centrality in interconnected multilayer networks.
CoRR
(2013)
\end{botherref}
\endbibitem

\bibitem{notre2}
\begin{barticle}
\bauthor{\bsnm{Ghalmane}, \binits{Z.}},
\bauthor{\bsnm{El~Hassouni}, \binits{M.}},
\bauthor{\bsnm{Cherifi}, \binits{C.}},
\bauthor{\bsnm{Cherifi}, \binits{H.}}:
\batitle{Centrality in modular networks}.
\bjtitle{EPJ Data Science}
\bvolume{8}(\bissue{1}),
\bfpage{15}
(\byear{2019})
\end{barticle}
\endbibitem

\bibitem{imsdb2019}
\begin{botherref}
{The Internet Movie Script Database (IMSDb)}.
Last accessed: 2019-06-20.
\url{www.imsdb.com}
\end{botherref}
\endbibitem

\bibitem{simply2019}
\begin{botherref}
{Simply Scripts}.
Last accessed: 2019-06-20.
\url{www.simplyscripts.com}
\end{botherref}
\endbibitem

\bibitem{gunce4}
\begin{botherref}
\oauthor{\bsnm{Orman}, \binits{K.}},
\oauthor{\bsnm{Labatut}, \binits{V.}},
\oauthor{\bsnm{Cherifi}, \binits{H.}}:
An empirical study of the relation between community structure and
  transitivity.
Complex Networks,
99--110
(2013)
\end{botherref}
\endbibitem

\bibitem{girvan2002community}
\begin{barticle}
\bauthor{\bsnm{Girvan}, \binits{M.}},
\bauthor{\bsnm{Newman}, \binits{M.E.}}:
\batitle{Community structure in social and biological networks}.
\bjtitle{Proceedings of the national academy of sciences}
\bvolume{99}(\bissue{12}),
\bfpage{7821}--\blpage{7826}
(\byear{2002})
\end{barticle}
\endbibitem

\bibitem{blondel2008fast}
\begin{barticle}
\bauthor{\bsnm{Blondel}, \binits{V.D.}},
\bauthor{\bsnm{Guillaume}, \binits{J.-L.}},
\bauthor{\bsnm{Lambiotte}, \binits{R.}},
\bauthor{\bsnm{Lefebvre}, \binits{E.}}:
\batitle{Fast unfolding of communities in large networks}.
\bjtitle{Journal of statistical mechanics: theory and experiment}
\bvolume{2008}(\bissue{10}),
\bfpage{10008}
(\byear{2008})
\end{barticle}
\endbibitem

\bibitem{heymann2014gephi}
\begin{botherref}
\oauthor{\bsnm{Heymann}, \binits{S.}}:
Gephi.
Encyclopedia of social network analysis and mining,
612--625
(2014)
\end{botherref}
\endbibitem

\bibitem{comm}
\begin{barticle}
\bauthor{\bsnm{Gupta}, \binits{N.}},
\bauthor{\bsnm{Singh}, \binits{A.}},
\bauthor{\bsnm{Cherifi}, \binits{H.}}:
\batitle{Centrality measures for networks with community structure}.
\bjtitle{Physica A: Statistical Mechanics and its Applications}
\bvolume{452},
\bfpage{46}--\blpage{59}
(\byear{2016})
\end{barticle}
\endbibitem

\bibitem{notre3}
\begin{botherref}
\oauthor{\bsnm{Ghalmane}, \binits{Z.}},
\oauthor{\bsnm{El~Hassouni}, \binits{M.}},
\oauthor{\bsnm{Cherifi}, \binits{C.}},
\oauthor{\bsnm{Cherifi}, \binits{H.}}:
Centrality in complex networks with overlapping community structure.
Scientific Reports
\textbf{9}(10133)
(2019)
\end{botherref}
\endbibitem

\bibitem{sigaki2018history}
\begin{barticle}
\bauthor{\bsnm{Sigaki}, \binits{H.Y.}},
\bauthor{\bsnm{Perc}, \binits{M.}},
\bauthor{\bsnm{Ribeiro}, \binits{H.V.}}:
\batitle{History of art paintings through the lens of entropy and complexity}.
\bjtitle{Proceedings of the National Academy of Sciences}
\bvolume{115}(\bissue{37}),
\bfpage{8585}--\blpage{8594}
(\byear{2018})
\end{barticle}
\endbibitem

\bibitem{newman2006modularity}
\begin{barticle}
\bauthor{\bsnm{Newman}, \binits{M.E.}}:
\batitle{Modularity and community structure in networks}.
\bjtitle{Proceedings of the national academy of sciences}
\bvolume{103}(\bissue{23}),
\bfpage{8577}--\blpage{8582}
(\byear{2006})
\end{barticle}
\endbibitem

\bibitem{gupta2016centrality}
\begin{barticle}
\bauthor{\bsnm{Gupta}, \binits{N.}},
\bauthor{\bsnm{Singh}, \binits{A.}},
\bauthor{\bsnm{Cherifi}, \binits{H.}}:
\batitle{Centrality measures for networks with community structure}.
\bjtitle{Physica A: Statistical Mechanics and its Applications}
\bvolume{452},
\bfpage{46}--\blpage{59}
(\byear{2016})
\end{barticle}
\endbibitem

\bibitem{eude1994statistical}
\begin{bchapter}
\bauthor{\bsnm{Eude}, \binits{T.}},
\bauthor{\bsnm{Cherifi}, \binits{H.}},
\bauthor{\bsnm{Grisel}, \binits{R.}}:
\bctitle{Statistical distribution of dct coefficients and their application to
  an adaptive compression algorithm}.
In: \bbtitle{Proceedings of TENCON'94-1994 IEEE Region 10's 9th Annual
  International Conference on:'Frontiers of Computer Technology'},
pp. \bfpage{427}--\blpage{430}
(\byear{1994}).
\bcomment{IEEE}
\end{bchapter}
\endbibitem

\bibitem{orman2013empirical}
\begin{bchapter}
\bauthor{\bsnm{Orman}, \binits{K.}},
\bauthor{\bsnm{Labatut}, \binits{V.}},
\bauthor{\bsnm{Cherifi}, \binits{H.}}:
\bctitle{An empirical study of the relation between community structure and
  transitivity}.
In: \bbtitle{Complex Networks},
pp. \bfpage{99}--\blpage{110}.
\bpublisher{Springer}, \blocation{???}
(\byear{2013})
\end{bchapter}
\endbibitem

\bibitem{cherifi2019community}
\begin{botherref}
\oauthor{\bsnm{Cherifi}, \binits{H.}},
\oauthor{\bsnm{Palla}, \binits{G.}},
\oauthor{\bsnm{Szymanski}, \binits{B.K.}},
\oauthor{\bsnm{Lu}, \binits{X.}}:
On community structure in complex networks: challenges and opportunities.
arXiv preprint arXiv:1908.04901
(2019)
\end{botherref}
\endbibitem

\bibitem{ghalmane2019centrality}
\begin{barticle}
\bauthor{\bsnm{Ghalmane}, \binits{Z.}},
\bauthor{\bsnm{Cherifi}, \binits{C.}},
\bauthor{\bsnm{Cherifi}, \binits{H.}},
\bauthor{\bsnm{El~Hassouni}, \binits{M.}}:
\batitle{Centrality in complex networks with overlapping community structure}.
\bjtitle{Scientific reports}
\bvolume{9}(\bissue{1}),
\bfpage{15}
(\byear{2019})
\end{barticle}
\endbibitem

\end{thebibliography}

\newcommand{\BMCxmlcomment}[1]{}

\BMCxmlcomment{

<refgrp>

<bibl id="B1">
  <title><p>Weighted adaptive neighborhood hypergraph partitioning for image
  segmentation</p></title>
  <aug>
    <au><snm>Rital</snm><fnm>S</fnm></au>
    <au><snm>Cherifi</snm><fnm>H</fnm></au>
    <au><snm>Miguet</snm><fnm>S</fnm></au>
  </aug>
  <source>International Conference on Pattern Recognition and Image
  Analysis</source>
  <pubdate>2005</pubdate>
  <fpage>522</fpage>
  <lpage>-531</lpage>
</bibl>

<bibl id="B2">
  <title><p>Social network analysis in a movie using character-net</p></title>
  <aug>
    <au><snm>Park</snm><fnm>SB</fnm></au>
    <au><snm>Oh</snm><fnm>KJ</fnm></au>
    <au><snm>Jo</snm><fnm>GS</fnm></au>
  </aug>
  <source>Multimedia Tools and Applications</source>
  <publisher>Springer</publisher>
  <pubdate>2012</pubdate>
  <volume>59</volume>
  <issue>2</issue>
  <fpage>601</fpage>
  <lpage>-627</lpage>
</bibl>

<bibl id="B3">
  <title><p>Topology analysis of social networks extracted from
  literature</p></title>
  <aug>
    <au><snm>Waumans</snm><fnm>MC</fnm></au>
    <au><snm>Nicod{\`e}me</snm><fnm>T</fnm></au>
    <au><snm>Bersini</snm><fnm>H</fnm></au>
  </aug>
  <source>PloS one</source>
  <publisher>Public Library of Science</publisher>
  <pubdate>2015</pubdate>
  <volume>10</volume>
  <issue>6</issue>
  <fpage>e0126470</fpage>
</bibl>

<bibl id="B4">
  <title><p>A character network study of two sci-fi TV series</p></title>
  <aug>
    <au><snm>Tan</snm><fnm>MS</fnm></au>
    <au><snm>Ujum</snm><fnm>EA</fnm></au>
    <au><snm>Ratnavelu</snm><fnm>K</fnm></au>
  </aug>
  <source>AIP Conference Proceedings</source>
  <pubdate>2014</pubdate>
  <volume>1588</volume>
  <issue>1</issue>
  <fpage>246</fpage>
  <lpage>-251</lpage>
</bibl>

<bibl id="B5">
  <title><p>When face-tracking meets social networks: a story of politics in
  news videos</p></title>
  <aug>
    <au><snm>Renoust</snm><fnm>B</fnm></au>
    <au><snm>Kobayashi</snm><fnm>T</fnm></au>
    <au><snm>Ngo</snm><fnm>TD</fnm></au>
    <au><snm>Le</snm><fnm>DD</fnm></au>
    <au><snm>Satoh</snm><fnm>S</fnm></au>
  </aug>
  <source>Applied Network Science</source>
  <publisher>Springer</publisher>
  <pubdate>2016</pubdate>
  <volume>1</volume>
  <issue>1</issue>
  <fpage>4</fpage>
</bibl>

<bibl id="B6">
  <title><p>Visual analytics of political networks from face-tracking of news
  video</p></title>
  <aug>
    <au><snm>Renoust</snm><fnm>B</fnm></au>
    <au><snm>Le</snm><fnm>DD</fnm></au>
    <au><snm>Satoh</snm><fnm>S</fnm></au>
  </aug>
  <source>IEEE Transactions on Multimedia</source>
  <publisher>IEEE</publisher>
  <pubdate>2016</pubdate>
  <volume>18</volume>
  <issue>11</issue>
  <fpage>2184</fpage>
  <lpage>-2195</lpage>
</bibl>

<bibl id="B7">
  <title><p>Game of Nodes: A Social Network Analysis of Game of
  Thrones</p></title>
  <aug>
    <au><snm>Mish</snm><fnm>B</fnm></au>
  </aug>
  <publisher>https://gameofnodes.wordpress.com</publisher>
  <pubdate>2016</pubdate>
</bibl>

<bibl id="B8">
  <title><p>Multilayer Network Model of Movie Script</p></title>
  <aug>
    <au><snm>Mourchid</snm><fnm>Y</fnm></au>
    <au><snm>Renoust</snm><fnm>B</fnm></au>
    <au><snm>Cherifi</snm><fnm>H</fnm></au>
    <au><snm>El Hassouni</snm><fnm>M</fnm></au>
  </aug>
  <source>International Conference on Complex Networks and their
  Applications</source>
  <pubdate>2018</pubdate>
  <fpage>782</fpage>
  <lpage>-796</lpage>
</bibl>

<bibl id="B9">
  <title><p>Movie rating prediction using content-based and link stream
  features</p></title>
  <aug>
    <au><snm>Viard</snm><fnm>T</fnm></au>
    <au><snm>Fournier{-}S'niehotta</snm><fnm>R</fnm></au>
  </aug>
  <source>CoRR</source>
  <pubdate>2018</pubdate>
  <volume>abs/1805.02893</volume>
</bibl>

<bibl id="B10">
  <title><p>Applying network theory to fables: complexity in Slovene
  belles-lettres for different age groups</p></title>
  <aug>
    <au><snm>Markovi{\v{c}}</snm><fnm>R</fnm></au>
    <au><snm>Gosak</snm><fnm>M</fnm></au>
    <au><snm>Perc</snm><fnm>M</fnm></au>
    <au><snm>Marhl</snm><fnm>M</fnm></au>
    <au><snm>Grubelnik</snm><fnm>V</fnm></au>
  </aug>
  <source>Journal of Complex Networks</source>
  <publisher>Oxford University Press</publisher>
  <pubdate>2018</pubdate>
  <volume>7</volume>
  <issue>1</issue>
  <fpage>114</fpage>
  <lpage>-127</lpage>
</bibl>

<bibl id="B11">
  <title><p>A novel video summarization based on mining the story-structure and
  semantic relations among concept entities</p></title>
  <aug>
    <au><snm>Chen</snm><fnm>BW</fnm></au>
    <au><snm>Wang</snm><fnm>JC</fnm></au>
    <au><snm>Wang</snm><fnm>JF</fnm></au>
  </aug>
  <source>IEEE Transactions on Multimedia</source>
  <publisher>IEEE</publisher>
  <pubdate>2009</pubdate>
  <volume>11</volume>
  <issue>2</issue>
  <fpage>295</fpage>
  <lpage>-312</lpage>
</bibl>

<bibl id="B12">
  <title><p>Just so stories for little children</p></title>
  <aug>
    <au><cnm>Kipling</cnm></au>
  </aug>
  <publisher>New York: Doubleday</publisher>
  <pubdate>1909</pubdate>
</bibl>

<bibl id="B13">
  <title><p>Visual movie analytics</p></title>
  <aug>
    <au><snm>Kurzhals</snm><fnm>K</fnm></au>
    <au><snm>John</snm><fnm>M</fnm></au>
    <au><snm>Heimerl</snm><fnm>F</fnm></au>
    <au><snm>Kuznecov</snm><fnm>P</fnm></au>
    <au><snm>Weiskopf</snm><fnm>D</fnm></au>
  </aug>
  <source>IEEE Transactions on Multimedia</source>
  <publisher>IEEE</publisher>
  <pubdate>2016</pubdate>
  <volume>18</volume>
  <issue>11</issue>
  <fpage>2149</fpage>
  <lpage>-2160</lpage>
</bibl>

<bibl id="B14">
  <title><p>Fundamental structures of dynamic social networks</p></title>
  <aug>
    <au><snm>Sekara</snm><fnm>V</fnm></au>
    <au><snm>Stopczynski</snm><fnm>A</fnm></au>
    <au><snm>Lehmann</snm><fnm>S</fnm></au>
  </aug>
  <source>Proceedings of the national academy of sciences</source>
  <publisher>National Acad Sciences</publisher>
  <pubdate>2016</pubdate>
  <volume>113</volume>
  <issue>36</issue>
  <fpage>9977</fpage>
  <lpage>-9982</lpage>
</bibl>

<bibl id="B15">
  <title><p>Entanglement in multiplex networks: understanding group cohesion in
  homophily networks</p></title>
  <aug>
    <au><snm>Renoust</snm><fnm>B</fnm></au>
    <au><snm>Melan{\c{c}}on</snm><fnm>G</fnm></au>
    <au><snm>Viaud</snm><fnm>ML</fnm></au>
  </aug>
  <source>Social Network Analysis-Community Detection and Evolution</source>
  <publisher>Springer</publisher>
  <pubdate>2014</pubdate>
  <fpage>89</fpage>
  <lpage>-117</lpage>
</bibl>

<bibl id="B16">
  <title><p>Recommendations in location-based social networks: a
  survey</p></title>
  <aug>
    <au><snm>Bao</snm><fnm>J</fnm></au>
    <au><snm>Zheng</snm><fnm>Y</fnm></au>
    <au><snm>Wilkie</snm><fnm>D</fnm></au>
    <au><snm>Mokbel</snm><fnm>M</fnm></au>
  </aug>
  <source>GeoInformatica</source>
  <publisher>Springer</publisher>
  <pubdate>2015</pubdate>
  <volume>19</volume>
  <issue>3</issue>
  <fpage>525</fpage>
  <lpage>-565</lpage>
</bibl>

<bibl id="B17">
  <title><p>Stream graphs and link streams for the modeling of interactions
  over time</p></title>
  <aug>
    <au><snm>Latapy</snm><fnm>M</fnm></au>
    <au><snm>Viard</snm><fnm>T</fnm></au>
    <au><snm>Magnien</snm><fnm>C</fnm></au>
  </aug>
  <source>Social Network Analysis and Mining</source>
  <publisher>Springer</publisher>
  <pubdate>2018</pubdate>
  <volume>8</volume>
  <issue>1</issue>
  <fpage>61</fpage>
</bibl>

<bibl id="B18">
  <title><p>Exploiting Structure and Conventions of Movie Scripts for
  Information Retrieval and Text Mining</p></title>
  <aug>
    <au><snm>Jhala</snm><fnm>A</fnm></au>
  </aug>
  <source>Joint International Conference on Interactive Digital
  Storytelling</source>
  <pubdate>2008</pubdate>
  <fpage>210</fpage>
  <lpage>-213</lpage>
</bibl>

<bibl id="B19">
  <title><p>Deep learning for visual understanding: A review</p></title>
  <aug>
    <au><snm>Guo</snm><fnm>Y</fnm></au>
    <au><snm>Liu</snm><fnm>Y</fnm></au>
    <au><snm>Oerlemans</snm><fnm>A</fnm></au>
    <au><snm>Lao</snm><fnm>S</fnm></au>
    <au><snm>Wu</snm><fnm>S</fnm></au>
    <au><snm>Lew</snm><fnm>MS</fnm></au>
  </aug>
  <source>Neurocomputing</source>
  <publisher>Elsevier</publisher>
  <pubdate>2016</pubdate>
  <volume>187</volume>
  <fpage>27</fpage>
  <lpage>-48</lpage>
</bibl>

<bibl id="B20">
  <title><p>A comparison of multiclass SVM methods for real world natural
  scenes</p></title>
  <aug>
    <au><snm>Demirkesen</snm><fnm>C</fnm></au>
    <au><snm>Cherifi</snm><fnm>H</fnm></au>
  </aug>
  <source>International Conference on Advanced Concepts for Intelligent Vision
  Systems</source>
  <pubdate>2008</pubdate>
  <fpage>752</fpage>
  <lpage>-763</lpage>
</bibl>

<bibl id="B21">
  <title><p>Predicting subjective video quality from separated spatial and
  temporal assessment</p></title>
  <aug>
    <au><snm>Pastrana Vidal</snm><fnm>RR</fnm></au>
    <au><snm>Gicquel</snm><fnm>JC</fnm></au>
    <au><snm>Blin</snm><fnm>JL</fnm></au>
    <au><snm>Cherifi</snm><fnm>H</fnm></au>
  </aug>
  <source>Human Vision and Electronic Imaging XI</source>
  <pubdate>2006</pubdate>
  <volume>6057</volume>
  <fpage>60570S</fpage>
</bibl>

<bibl id="B22">
  <title><p>Face detection with the faster R-CNN</p></title>
  <aug>
    <au><snm>Jiang</snm><fnm>H</fnm></au>
    <au><snm>Learned Miller</snm><fnm>E</fnm></au>
  </aug>
  <source>Automatic Face \& Gesture Recognition (FG 2017), 2017 12th IEEE
  International Conference on</source>
  <pubdate>2017</pubdate>
  <fpage>650</fpage>
  <lpage>-657</lpage>
</bibl>

<bibl id="B23">
  <title><p>Vggface2: A dataset for recognising faces across pose and
  age</p></title>
  <aug>
    <au><snm>Cao</snm><fnm>Q</fnm></au>
    <au><snm>Shen</snm><fnm>L</fnm></au>
    <au><snm>Xie</snm><fnm>W</fnm></au>
    <au><snm>Parkhi</snm><fnm>OM</fnm></au>
    <au><snm>Zisserman</snm><fnm>A</fnm></au>
  </aug>
  <source>Automatic Face \& Gesture Recognition (FG 2018), 2018 13th IEEE
  International Conference on</source>
  <pubdate>2018</pubdate>
  <fpage>67</fpage>
  <lpage>-74</lpage>
</bibl>

<bibl id="B24">
  <title><p>Densecap: Fully convolutional localization networks for dense
  captioning</p></title>
  <aug>
    <au><snm>Johnson</snm><fnm>J</fnm></au>
    <au><snm>Karpathy</snm><fnm>A</fnm></au>
    <au><snm>Fei Fei</snm><fnm>L</fnm></au>
  </aug>
  <source>Proceedings of the IEEE Conference on Computer Vision and Pattern
  Recognition</source>
  <pubdate>2016</pubdate>
  <fpage>4565</fpage>
  <lpage>-4574</lpage>
</bibl>

<bibl id="B25">
  <title><p>Dense captioning with joint inference and visual
  context</p></title>
  <aug>
    <au><snm>Yang</snm><fnm>L</fnm></au>
    <au><snm>Tang</snm><fnm>K</fnm></au>
    <au><snm>Yang</snm><fnm>J</fnm></au>
    <au><snm>Li</snm><fnm>LJ</fnm></au>
  </aug>
  <source>Proceedings of the IEEE Conference on Computer Vision and Pattern
  Recognition (CVPR)</source>
  <pubdate>2017</pubdate>
  <volume>2</volume>
</bibl>

<bibl id="B26">
  <title><p>Multilayer analysis and visualization of networks</p></title>
  <aug>
    <au><snm>Domenico</snm><fnm>MD</fnm></au>
    <au><snm>Porter</snm><fnm>MA</fnm></au>
    <au><snm>Arenas</snm><fnm>A</fnm></au>
  </aug>
  <source>J. Complex Netw</source>
  <pubdate>2014</pubdate>
  <volume>10</volume>
</bibl>

<bibl id="B27">
  <title><p>Multilayer networks</p></title>
  <aug>
    <au><snm>Kivel{\"a}</snm><fnm>M</fnm></au>
    <au><snm>Arenas</snm><fnm>A</fnm></au>
    <au><snm>Barthelemy</snm><fnm>M</fnm></au>
    <au><snm>Gleeson</snm><fnm>JP</fnm></au>
    <au><snm>Moreno</snm><fnm>Y</fnm></au>
    <au><snm>Porter</snm><fnm>MA</fnm></au>
  </aug>
  <source>Journal of complex networks</source>
  <publisher>Oxford University Press</publisher>
  <pubdate>2014</pubdate>
  <volume>2</volume>
  <issue>3</issue>
  <fpage>203</fpage>
  <lpage>-271</lpage>
</bibl>

<bibl id="B28">
  <title><p>{Star Wars: Episode IV - A New Hope}</p></title>
  <aug>
    <au><snm>Lucas</snm><fnm>G</fnm></au>
  </aug>
  <source>Twentieth Century Fox Film Corporation</source>
  <pubdate>1977</pubdate>
</bibl>

<bibl id="B29">
  <title><p>{Star Wars: Episode V - The Empire Strikes Back}</p></title>
  <aug>
    <au><snm>Lucas</snm><fnm>G</fnm></au>
  </aug>
  <source>Twentieth Century Fox Film Corporation</source>
  <pubdate>1980</pubdate>
</bibl>

<bibl id="B30">
  <title><p>{Star Wars: Episode VI - Return of the Jedi}</p></title>
  <aug>
    <au><snm>Lucas</snm><fnm>G</fnm></au>
  </aug>
  <source>Twentieth Century Fox Film Corporation</source>
  <pubdate>1983</pubdate>
</bibl>

<bibl id="B31">
  <title><p>{Star Wars: Episode I - The Phantom Menace}</p></title>
  <aug>
    <au><snm>Lucas</snm><fnm>G</fnm></au>
  </aug>
  <source>Twentieth Century Fox Film Corporation</source>
  <pubdate>1999</pubdate>
</bibl>

<bibl id="B32">
  <title><p>{Star Wars: Episode II - Attack of the Clones}</p></title>
  <aug>
    <au><snm>Lucas</snm><fnm>G</fnm></au>
  </aug>
  <source>Twentieth Century Fox Film Corporation</source>
  <pubdate>2002</pubdate>
</bibl>

<bibl id="B33">
  <title><p>{Star Wars: Episode III - Revenge of the Sith}</p></title>
  <aug>
    <au><snm>Lucas</snm><fnm>G</fnm></au>
  </aug>
  <source>Twentieth Century Fox Film Corporation</source>
  <pubdate>2005</pubdate>
</bibl>

<bibl id="B34">
  <title><p>Understanding social networks: Theories, concepts, and
  findings</p></title>
  <aug>
    <au><snm>Kadushin</snm><fnm>C</fnm></au>
  </aug>
  <publisher>OUP USA</publisher>
  <pubdate>2012</pubdate>
</bibl>

<bibl id="B35">
  <title><p>Extracting story units from long programs for video browsing and
  navigation</p></title>
  <aug>
    <au><snm>Yeung</snm><fnm>M</fnm></au>
    <au><snm>Yeo</snm><fnm>BL</fnm></au>
    <au><snm>Liu</snm><fnm>B</fnm></au>
  </aug>
  <source>Multimedia Computing and Systems, 1996., Proceedings of the Third
  IEEE International Conference on</source>
  <pubdate>1996</pubdate>
  <fpage>296</fpage>
  <lpage>-305</lpage>
</bibl>

<bibl id="B36">
  <title><p>Narrative abstraction model for story-oriented video</p></title>
  <aug>
    <au><snm>Jung</snm><fnm>B</fnm></au>
    <au><snm>Kwak</snm><fnm>T</fnm></au>
    <au><snm>Song</snm><fnm>J</fnm></au>
    <au><snm>Lee</snm><fnm>Y</fnm></au>
  </aug>
  <source>Proceedings of the 12th annual ACM international conference on
  Multimedia</source>
  <pubdate>2004</pubdate>
  <fpage>828</fpage>
  <lpage>-835</lpage>
</bibl>

<bibl id="B37">
  <title><p>Semantic flow in language networks</p></title>
  <aug>
    <au><snm>Jr.</snm><fnm>EAC</fnm></au>
    <au><snm>Marinho</snm><fnm>VQ</fnm></au>
    <au><snm>Amancio</snm><fnm>DR</fnm></au>
  </aug>
  <source>CoRR</source>
  <pubdate>2019</pubdate>
  <volume>abs/1905.07595</volume>
</bibl>

<bibl id="B38">
  <title><p>The Stanford GraphBase: a platform for combinatorial
  computing</p></title>
  <aug>
    <au><snm>Knuth</snm><fnm>DE</fnm></au>
  </aug>
  <source>AcM Press New York</source>
  <pubdate>1993</pubdate>
</bibl>

<bibl id="B39">
  <title><p>Unsupervised cluster analyses of character networks in fiction:
  Community structure and centrality</p></title>
  <aug>
    <au><snm>Chen</snm><fnm>RH G</fnm></au>
    <au><snm>Chen</snm><fnm>C C</fnm></au>
    <au><snm>Chen</snm><fnm>C M</fnm></au>
  </aug>
  <source>Knowledge-Based Systems</source>
  <publisher>Elsevier</publisher>
  <pubdate>2019</pubdate>
  <volume>163</volume>
  <fpage>800</fpage>
  <lpage>-810</lpage>
</bibl>

<bibl id="B40">
  <title><p>Rolenet: Movie analysis from the perspective of social
  networks</p></title>
  <aug>
    <au><snm>Weng</snm><fnm>CY</fnm></au>
    <au><snm>Chu</snm><fnm>WT</fnm></au>
    <au><snm>Wu</snm><fnm>JL</fnm></au>
  </aug>
  <source>IEEE Transactions on Multimedia</source>
  <publisher>IEEE</publisher>
  <pubdate>2009</pubdate>
  <volume>11</volume>
  <issue>2</issue>
  <fpage>256</fpage>
  <lpage>-271</lpage>
</bibl>

<bibl id="B41">
  <title><p>CoCharNet: Extracting Social Networks using Character Co-occurrence
  in Movies.</p></title>
  <aug>
    <au><snm>Tran</snm><fnm>QD</fnm></au>
    <au><snm>Jung</snm><fnm>JE</fnm></au>
  </aug>
  <source>J. UCS</source>
  <pubdate>2015</pubdate>
  <volume>21</volume>
  <issue>6</issue>
  <fpage>796</fpage>
  <lpage>-815</lpage>
</bibl>

<bibl id="B42">
  <title><p>SRN: The movie character relationship analysis via social
  network</p></title>
  <aug>
    <au><snm>He</snm><fnm>J</fnm></au>
    <au><snm>Xie</snm><fnm>Y</fnm></au>
    <au><snm>Luan</snm><fnm>X</fnm></au>
    <au><snm>Zhang</snm><fnm>L</fnm></au>
    <au><snm>Zhang</snm><fnm>X</fnm></au>
  </aug>
  <source>International Conference on Multimedia Modeling</source>
  <pubdate>2018</pubdate>
  <fpage>289</fpage>
  <lpage>-301</lpage>
</bibl>

<bibl id="B43">
  <title><p>What’s This Movie About? A Joint Neural Network Architecture for
  Movie Content Analysis</p></title>
  <aug>
    <au><snm>Gorinski</snm><fnm>PJ</fnm></au>
    <au><snm>Lapata</snm><fnm>M</fnm></au>
  </aug>
  <source>Proceedings of the 2018 Conference of the North American Chapter of
  the Association for Computational Linguistics: Human Language Technologies,
  Volume 1 (Long Papers)</source>
  <pubdate>2018</pubdate>
  <fpage>1770</fpage>
  <lpage>-1781</lpage>
</bibl>

<bibl id="B44">
  <title><p>StoryRoleNet: Social Network Construction of Role Relationship in
  Video</p></title>
  <aug>
    <au><snm>Lv</snm><fnm>J</fnm></au>
    <au><snm>Wu</snm><fnm>B</fnm></au>
    <au><snm>Zhou</snm><fnm>L</fnm></au>
    <au><snm>Wang</snm><fnm>H</fnm></au>
  </aug>
  <source>IEEE Access</source>
  <publisher>IEEE</publisher>
  <pubdate>2018</pubdate>
  <volume>6</volume>
  <fpage>25958</fpage>
  <lpage>-25969</lpage>
</bibl>

<bibl id="B45">
  <title><p>Generating “visual clouds” from multiplex networks for tv news
  archive query visualization</p></title>
  <aug>
    <au><snm>Ren</snm><fnm>H</fnm></au>
    <au><snm>Renoust</snm><fnm>B</fnm></au>
    <au><snm>Viaud</snm><fnm>ML</fnm></au>
    <au><snm>Melan{\c{c}}on</snm><fnm>G</fnm></au>
    <au><snm>Satoh</snm><fnm>S</fnm></au>
  </aug>
  <source>2018 International Conference on Content-Based Multimedia Indexing
  (CBMI)</source>
  <pubdate>2018</pubdate>
  <fpage>1</fpage>
  <lpage>-6</lpage>
</bibl>

<bibl id="B46">
  <title><p>Newspaper writing in high schools: Containing an outline for the
  use of teachers</p></title>
  <aug>
    <au><snm>Flint</snm><fnm>LN</fnm></au>
  </aug>
  <publisher>Pub. from the Department of journalism Press in the University of
  Kansas</publisher>
  <pubdate>1917</pubdate>
</bibl>

<bibl id="B47">
  <title><p>A survey of named entity recognition and classification</p></title>
  <aug>
    <au><snm>Nadeau</snm><fnm>D</fnm></au>
    <au><snm>Sekine</snm><fnm>S</fnm></au>
  </aug>
  <source>Lingvisticae Investigationes</source>
  <publisher>John Benjamins</publisher>
  <pubdate>2007</pubdate>
  <volume>30</volume>
  <issue>1</issue>
  <fpage>3</fpage>
  <lpage>-26</lpage>
</bibl>

<bibl id="B48">
  <title><p>Choosing an NLP library for analyzing software documentation: a
  systematic literature review and a series of experiments</p></title>
  <aug>
    <au><snm>Al Omran</snm><fnm>FNA</fnm></au>
    <au><snm>Treude</snm><fnm>C</fnm></au>
  </aug>
  <source>Proceedings of the 14th International Conference on Mining Software
  Repositories</source>
  <pubdate>2017</pubdate>
  <fpage>187</fpage>
  <lpage>-197</lpage>
</bibl>

<bibl id="B49">
  <title><p>A vector space model for automatic indexing</p></title>
  <aug>
    <au><snm>Salton</snm><fnm>G</fnm></au>
    <au><snm>Wong</snm><fnm>A</fnm></au>
    <au><snm>Yang</snm><fnm>CS</fnm></au>
  </aug>
  <source>Communications of the ACM</source>
  <publisher>ACM</publisher>
  <pubdate>1975</pubdate>
  <volume>18</volume>
  <issue>11</issue>
  <fpage>613</fpage>
  <lpage>-620</lpage>
</bibl>

<bibl id="B50">
  <title><p>Keyword extraction based on tf/idf for Chinese news
  document</p></title>
  <aug>
    <au><snm>Li</snm><fnm>J</fnm></au>
    <au><snm>Zhang</snm><fnm>K</fnm></au>
    <au><cnm>others</cnm></au>
  </aug>
  <source>Wuhan University Journal of Natural Sciences</source>
  <publisher>Springer</publisher>
  <pubdate>2007</pubdate>
  <volume>12</volume>
  <issue>5</issue>
  <fpage>917</fpage>
  <lpage>-921</lpage>
</bibl>

<bibl id="B51">
  <title><p>Latent dirichlet allocation</p></title>
  <aug>
    <au><snm>Blei</snm><fnm>DM</fnm></au>
    <au><snm>Ng</snm><fnm>AY</fnm></au>
    <au><snm>Jordan</snm><fnm>MI</fnm></au>
  </aug>
  <source>Journal of machine Learning research</source>
  <pubdate>2003</pubdate>
  <volume>3</volume>
  <issue>Jan</issue>
  <fpage>993</fpage>
  <lpage>-1022</lpage>
</bibl>

<bibl id="B52">
  <title><p>A keyword extraction algorithm based on Word2vec</p></title>
  <aug>
    <au><snm>Yuepeng</snm><fnm>L</fnm></au>
    <au><snm>Cui</snm><fnm>J</fnm></au>
    <au><snm>Junchuan</snm><fnm>J</fnm></au>
  </aug>
  <source>e-Science Technology \& Application</source>
  <pubdate>2015</pubdate>
  <volume>4</volume>
  <fpage>54</fpage>
  <lpage>-59</lpage>
</bibl>

<bibl id="B53">
  <title><p>{PySceneDetect}</p></title>
  <aug>
    <au><snm>Castellano</snm><fnm>B</fnm></au>
  </aug>
  <pubdate>2012</pubdate>
  <url>github.com/Breakthrough/PySceneDetect</url>
  <note>Last accessed: 2019-06-20</note>
</bibl>

<bibl id="B54">
  <title><p>WIDER FACE: A Face Detection Benchmark</p></title>
  <aug>
    <au><snm>Yang</snm><fnm>S</fnm></au>
    <au><snm>Luo</snm><fnm>P</fnm></au>
    <au><snm>Loy</snm><fnm>CC</fnm></au>
    <au><snm>Tang</snm><fnm>X</fnm></au>
  </aug>
  <source>IEEE Conference on Computer Vision and Pattern Recognition
  (CVPR)</source>
  <pubdate>2016</pubdate>
</bibl>

<bibl id="B55">
  <title><p>Deep residual learning for image recognition</p></title>
  <aug>
    <au><snm>He</snm><fnm>K</fnm></au>
    <au><snm>Zhang</snm><fnm>X</fnm></au>
    <au><snm>Ren</snm><fnm>S</fnm></au>
    <au><snm>Sun</snm><fnm>J</fnm></au>
  </aug>
  <source>Proceedings of the IEEE conference on computer vision and pattern
  recognition</source>
  <pubdate>2016</pubdate>
  <fpage>770</fpage>
  <lpage>-778</lpage>
</bibl>

<bibl id="B56">
  <title><p>Parametric nonlinear dimensionality reduction using kernel
  t-SNE</p></title>
  <aug>
    <au><snm>Gisbrecht</snm><fnm>A</fnm></au>
    <au><snm>Schulz</snm><fnm>A</fnm></au>
    <au><snm>Hammer</snm><fnm>B</fnm></au>
  </aug>
  <source>Neurocomputing</source>
  <publisher>Elsevier</publisher>
  <pubdate>2015</pubdate>
  <volume>147</volume>
  <fpage>71</fpage>
  <lpage>-82</lpage>
</bibl>

<bibl id="B57">
  <title><p>TULIP 5</p></title>
  <aug>
    <au><snm>Auber</snm><fnm>D</fnm></au>
    <au><snm>Archambault</snm><fnm>D</fnm></au>
    <au><snm>Bourqui</snm><fnm>R</fnm></au>
    <au><snm>Delest</snm><fnm>M</fnm></au>
    <au><snm>Dubois</snm><fnm>J</fnm></au>
    <au><snm>Lambert</snm><fnm>A</fnm></au>
    <au><snm>Mary</snm><fnm>P</fnm></au>
    <au><snm>Mathiaut</snm><fnm>M</fnm></au>
    <au><snm>M{\'e}lan{\c{c}}on</snm><fnm>G</fnm></au>
    <au><snm>Pinaud</snm><fnm>B</fnm></au>
    <au><snm>Renoust</snm><fnm>B</fnm></au>
    <au><snm>Vallet</snm><fnm>J</fnm></au>
  </aug>
  <publisher>Springer</publisher>
  <pubdate>2017</pubdate>
  <fpage>1</fpage>
  <lpage>28</lpage>
</bibl>

<bibl id="B58">
  <title><p>Density-based spatial clustering of applications with
  noise</p></title>
  <aug>
    <au><snm>Ester</snm><fnm>M</fnm></au>
    <au><snm>Kriegel</snm><fnm>HP</fnm></au>
    <au><snm>Sander</snm><fnm>J</fnm></au>
    <au><snm>Xu</snm><fnm>X</fnm></au>
  </aug>
  <source>Int. Conf. Knowledge Discovery and Data Mining</source>
  <pubdate>1996</pubdate>
  <volume>240</volume>
</bibl>

<bibl id="B59">
  <title><p>Visual Genome: Connecting Language and Vision Using Crowdsourced
  Dense Image Annotations</p></title>
  <aug>
    <au><snm>Krishna</snm><fnm>R</fnm></au>
    <au><snm>Zhu</snm><fnm>Y</fnm></au>
    <au><snm>Groth</snm><fnm>O</fnm></au>
    <au><snm>Johnson</snm><fnm>J</fnm></au>
    <au><snm>Hata</snm><fnm>K</fnm></au>
    <au><snm>Kravitz</snm><fnm>J</fnm></au>
    <au><snm>Chen</snm><fnm>S</fnm></au>
    <au><snm>Kalantidis</snm><fnm>Y</fnm></au>
    <au><snm>Li</snm><fnm>LJ</fnm></au>
    <au><snm>Shamma</snm><fnm>DA</fnm></au>
    <au><snm>Bernstein</snm><fnm>MS</fnm></au>
    <au><snm>Fei Fei</snm><fnm>L</fnm></au>
  </aug>
  <source>International Journal of Computer Vision</source>
  <pubdate>2017</pubdate>
  <volume>123</volume>
  <issue>1</issue>
  <fpage>32</fpage>
  <lpage>-73</lpage>
</bibl>

<bibl id="B60">
  <title><p>N-gram-based text categorization</p></title>
  <aug>
    <au><snm>Cavnar</snm><fnm>WB</fnm></au>
    <au><snm>Trenkle</snm><fnm>JM</fnm></au>
    <au><cnm>others</cnm></au>
  </aug>
  <source>Proceedings of SDAIR-94, 3rd annual symposium on document analysis
  and information retrieval</source>
  <pubdate>1994</pubdate>
  <volume>161175</volume>
</bibl>

<bibl id="B61">
  <title><p>Is This Movie a Milestone? Identification of the Most Influential
  Movies in the History of Cinema</p></title>
  <aug>
    <au><snm>Bioglio</snm><fnm>L</fnm></au>
    <au><snm>Pensa</snm><fnm>RG</fnm></au>
  </aug>
  <source>International Workshop on Complex Networks and their
  Applications</source>
  <pubdate>2017</pubdate>
  <fpage>921</fpage>
  <lpage>-934</lpage>
</bibl>

<bibl id="B62">
  <title><p>Centrality in interconnected multilayer networks</p></title>
  <aug>
    <au><snm>Domenico</snm><fnm>MD</fnm></au>
    <au><snm>Sol Ribalta</snm><fnm>A</fnm></au>
    <au><snm>Omodei</snm><fnm>E</fnm></au>
    <au><snm>Gmez</snm><fnm>S</fnm></au>
    <au><snm>Arenas</snm><fnm>A</fnm></au>
  </aug>
  <source>CoRR</source>
  <pubdate>2013</pubdate>
</bibl>

<bibl id="B63">
  <title><p>Centrality in modular networks</p></title>
  <aug>
    <au><snm>Ghalmane</snm><fnm>Z</fnm></au>
    <au><snm>El Hassouni</snm><fnm>M</fnm></au>
    <au><snm>Cherifi</snm><fnm>C</fnm></au>
    <au><snm>Cherifi</snm><fnm>H</fnm></au>
  </aug>
  <source>EPJ Data Science</source>
  <publisher>SpringerOpen</publisher>
  <pubdate>2019</pubdate>
  <volume>8</volume>
  <issue>1</issue>
  <fpage>15</fpage>
</bibl>

<bibl id="B64">
  <title><p>{The Internet Movie Script Database (IMSDb)}</p></title>
  <url>www.imsdb.com</url>
  <note>Last accessed: 2019-06-20</note>
</bibl>

<bibl id="B65">
  <title><p>{Simply Scripts}</p></title>
  <url>www.simplyscripts.com</url>
  <note>Last accessed: 2019-06-20</note>
</bibl>

<bibl id="B66">
  <title><p>An empirical study of the relation between community structure and
  transitivity</p></title>
  <aug>
    <au><snm>Orman</snm><fnm>K</fnm></au>
    <au><snm>Labatut</snm><fnm>V</fnm></au>
    <au><snm>Cherifi</snm><fnm>H</fnm></au>
  </aug>
  <source>Complex Networks</source>
  <publisher>Springer</publisher>
  <pubdate>2013</pubdate>
  <fpage>99</fpage>
  <lpage>-110</lpage>
</bibl>

<bibl id="B67">
  <title><p>Community structure in social and biological networks</p></title>
  <aug>
    <au><snm>Girvan</snm><fnm>M</fnm></au>
    <au><snm>Newman</snm><fnm>ME</fnm></au>
  </aug>
  <source>Proceedings of the national academy of sciences</source>
  <publisher>National Academy of Sciences</publisher>
  <pubdate>2002</pubdate>
  <volume>99</volume>
  <issue>12</issue>
  <fpage>7821</fpage>
  <lpage>-7826</lpage>
</bibl>

<bibl id="B68">
  <title><p>Fast unfolding of communities in large networks</p></title>
  <aug>
    <au><snm>Blondel</snm><fnm>VD</fnm></au>
    <au><snm>Guillaume</snm><fnm>JL</fnm></au>
    <au><snm>Lambiotte</snm><fnm>R</fnm></au>
    <au><snm>Lefebvre</snm><fnm>E</fnm></au>
  </aug>
  <source>Journal of statistical mechanics: theory and experiment</source>
  <publisher>IOP Publishing</publisher>
  <pubdate>2008</pubdate>
  <volume>2008</volume>
  <issue>10</issue>
  <fpage>P10008</fpage>
</bibl>

<bibl id="B69">
  <title><p>Gephi</p></title>
  <aug>
    <au><snm>Heymann</snm><fnm>S</fnm></au>
  </aug>
  <source>Encyclopedia of social network analysis and mining</source>
  <publisher>Springer</publisher>
  <pubdate>2014</pubdate>
  <fpage>612</fpage>
  <lpage>-625</lpage>
</bibl>

<bibl id="B70">
  <title><p>Centrality measures for networks with community
  structure</p></title>
  <aug>
    <au><snm>Gupta</snm><fnm>N</fnm></au>
    <au><snm>Singh</snm><fnm>A</fnm></au>
    <au><snm>Cherifi</snm><fnm>H</fnm></au>
  </aug>
  <source>Physica A: Statistical Mechanics and its Applications</source>
  <publisher>Elsevier</publisher>
  <pubdate>2016</pubdate>
  <volume>452</volume>
  <fpage>46</fpage>
  <lpage>-59</lpage>
</bibl>

<bibl id="B71">
  <title><p>Centrality in complex networks with overlapping community
  structure</p></title>
  <aug>
    <au><snm>Ghalmane</snm><fnm>Z</fnm></au>
    <au><snm>El Hassouni</snm><fnm>M</fnm></au>
    <au><snm>Cherifi</snm><fnm>C</fnm></au>
    <au><snm>Cherifi</snm><fnm>H</fnm></au>
  </aug>
  <source>Scientific Reports</source>
  <publisher>Nature</publisher>
  <pubdate>2019</pubdate>
  <volume>9</volume>
  <issue>10133</issue>
</bibl>

<bibl id="B72">
  <title><p>History of art paintings through the lens of entropy and
  complexity</p></title>
  <aug>
    <au><snm>Sigaki</snm><fnm>HY</fnm></au>
    <au><snm>Perc</snm><fnm>M</fnm></au>
    <au><snm>Ribeiro</snm><fnm>HV</fnm></au>
  </aug>
  <source>Proceedings of the National Academy of Sciences</source>
  <publisher>National Acad Sciences</publisher>
  <pubdate>2018</pubdate>
  <volume>115</volume>
  <issue>37</issue>
  <fpage>E8585</fpage>
  <lpage>-E8594</lpage>
</bibl>

<bibl id="B73">
  <title><p>Modularity and community structure in networks</p></title>
  <aug>
    <au><snm>Newman</snm><fnm>ME</fnm></au>
  </aug>
  <source>Proceedings of the national academy of sciences</source>
  <publisher>National Acad Sciences</publisher>
  <pubdate>2006</pubdate>
  <volume>103</volume>
  <issue>23</issue>
  <fpage>8577</fpage>
  <lpage>-8582</lpage>
</bibl>

<bibl id="B74">
  <title><p>Centrality measures for networks with community
  structure</p></title>
  <aug>
    <au><snm>Gupta</snm><fnm>N</fnm></au>
    <au><snm>Singh</snm><fnm>A</fnm></au>
    <au><snm>Cherifi</snm><fnm>H</fnm></au>
  </aug>
  <source>Physica A: Statistical Mechanics and its Applications</source>
  <publisher>Elsevier</publisher>
  <pubdate>2016</pubdate>
  <volume>452</volume>
  <fpage>46</fpage>
  <lpage>-59</lpage>
</bibl>

<bibl id="B75">
  <title><p>Statistical distribution of DCT coefficients and their application
  to an adaptive compression algorithm</p></title>
  <aug>
    <au><snm>Eude</snm><fnm>T</fnm></au>
    <au><snm>Cherifi</snm><fnm>H</fnm></au>
    <au><snm>Grisel</snm><fnm>R</fnm></au>
  </aug>
  <source>Proceedings of TENCON'94-1994 IEEE Region 10's 9th Annual
  International Conference on:'Frontiers of Computer Technology'</source>
  <pubdate>1994</pubdate>
  <fpage>427</fpage>
  <lpage>-430</lpage>
</bibl>

<bibl id="B76">
  <title><p>An empirical study of the relation between community structure and
  transitivity</p></title>
  <aug>
    <au><snm>Orman</snm><fnm>K</fnm></au>
    <au><snm>Labatut</snm><fnm>V</fnm></au>
    <au><snm>Cherifi</snm><fnm>H</fnm></au>
  </aug>
  <source>Complex Networks</source>
  <publisher>Springer</publisher>
  <pubdate>2013</pubdate>
  <fpage>99</fpage>
  <lpage>-110</lpage>
</bibl>

<bibl id="B77">
  <title><p>On community structure in complex networks: challenges and
  opportunities</p></title>
  <aug>
    <au><snm>Cherifi</snm><fnm>H</fnm></au>
    <au><snm>Palla</snm><fnm>G</fnm></au>
    <au><snm>Szymanski</snm><fnm>BK</fnm></au>
    <au><snm>Lu</snm><fnm>X</fnm></au>
  </aug>
  <source>arXiv preprint arXiv:1908.04901</source>
  <pubdate>2019</pubdate>
</bibl>

<bibl id="B78">
  <title><p>Centrality in Complex Networks with overlapping Community
  structure</p></title>
  <aug>
    <au><snm>Ghalmane</snm><fnm>Z</fnm></au>
    <au><snm>Cherifi</snm><fnm>C</fnm></au>
    <au><snm>Cherifi</snm><fnm>H</fnm></au>
    <au><snm>El Hassouni</snm><fnm>M</fnm></au>
  </aug>
  <source>Scientific reports</source>
  <publisher>Nature Publishing Group</publisher>
  <pubdate>2019</pubdate>
  <volume>9</volume>
  <issue>1</issue>
  <fpage>15</fpage>
</bibl>

</refgrp>
} 

\section*{Supplementary Materials}

\begin{figure}[t!]
\centering
        \begin{subfigure}[b]{0.49\textwidth}
            \centering
            \includegraphics[width=\textwidth]{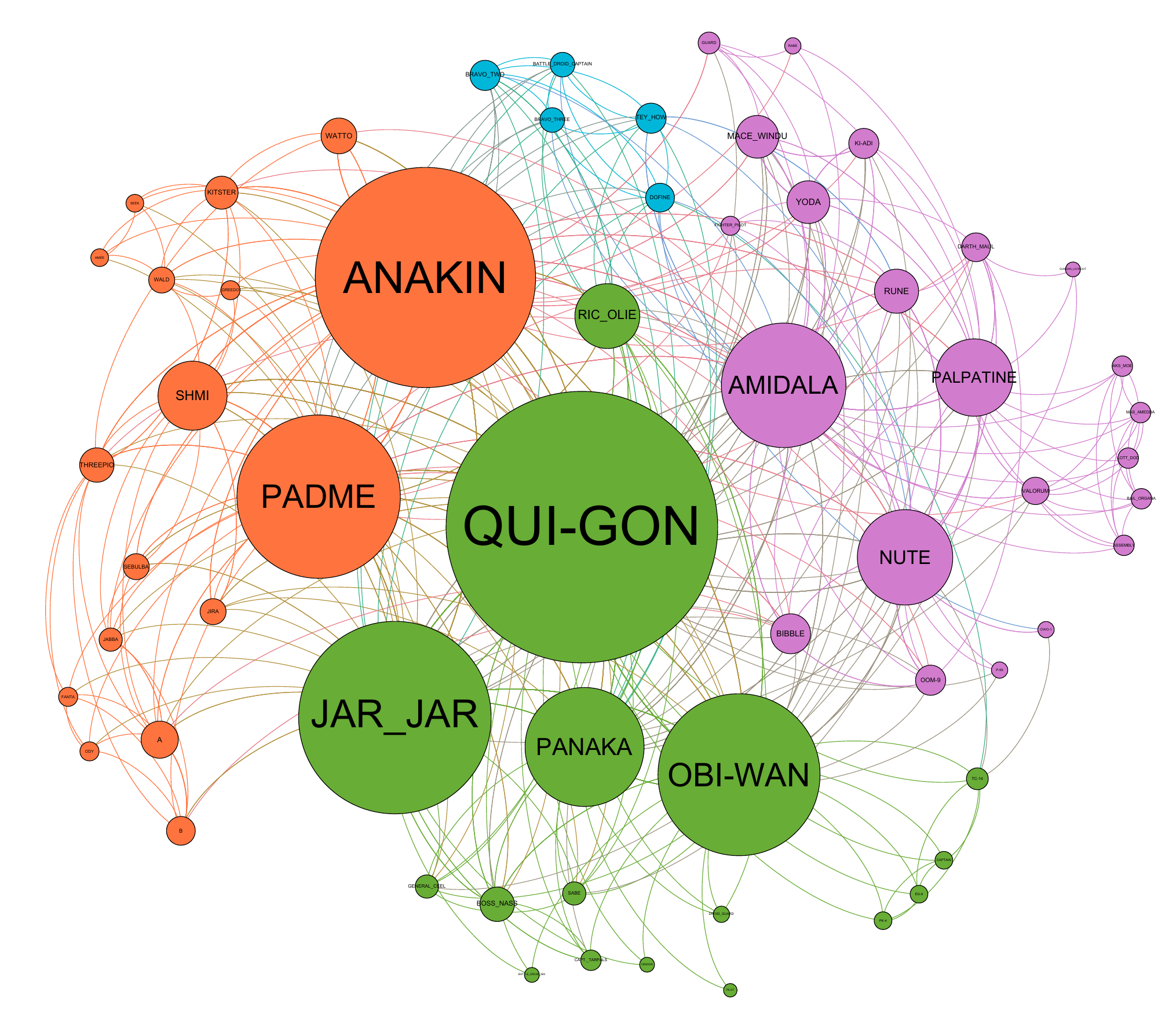}
            \caption[  $G_{CC}$] {$G_{CC}$}
            
            \label{fig:SW3C}
        \end{subfigure}
        \qquad        
        \begin{subfigure}[b]{0.48\textwidth}  
            \centering 
            \includegraphics[width=\textwidth]{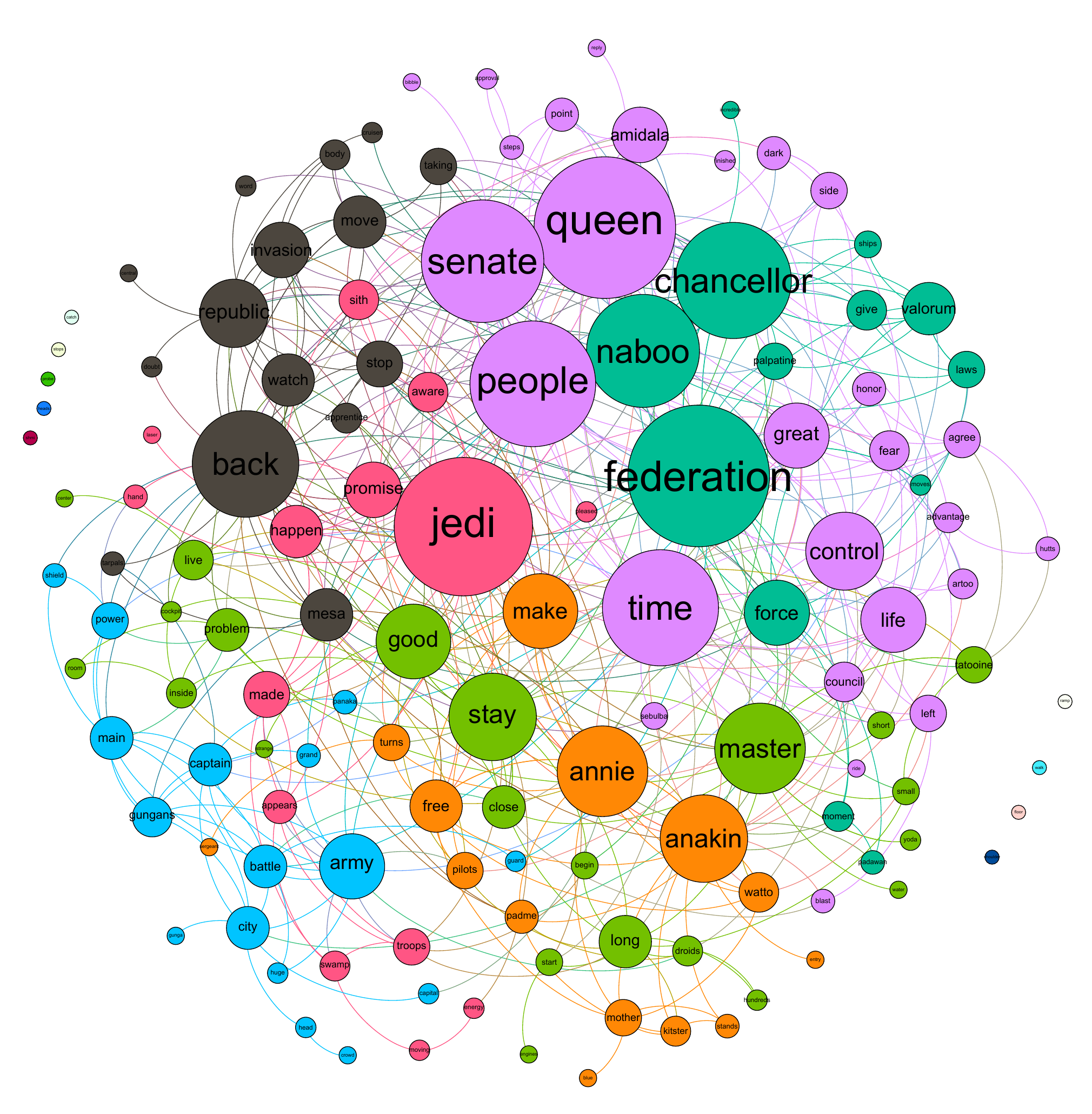}
            \caption[  $G_{KK}$]{$G_{KK}$}
            
            \label{fig:SW3K}
        \end{subfigure}
         \vskip\baselineskip    
        \begin{subfigure}[b]{0.48\textwidth}  
            \centering 
            \includegraphics[width=\textwidth]{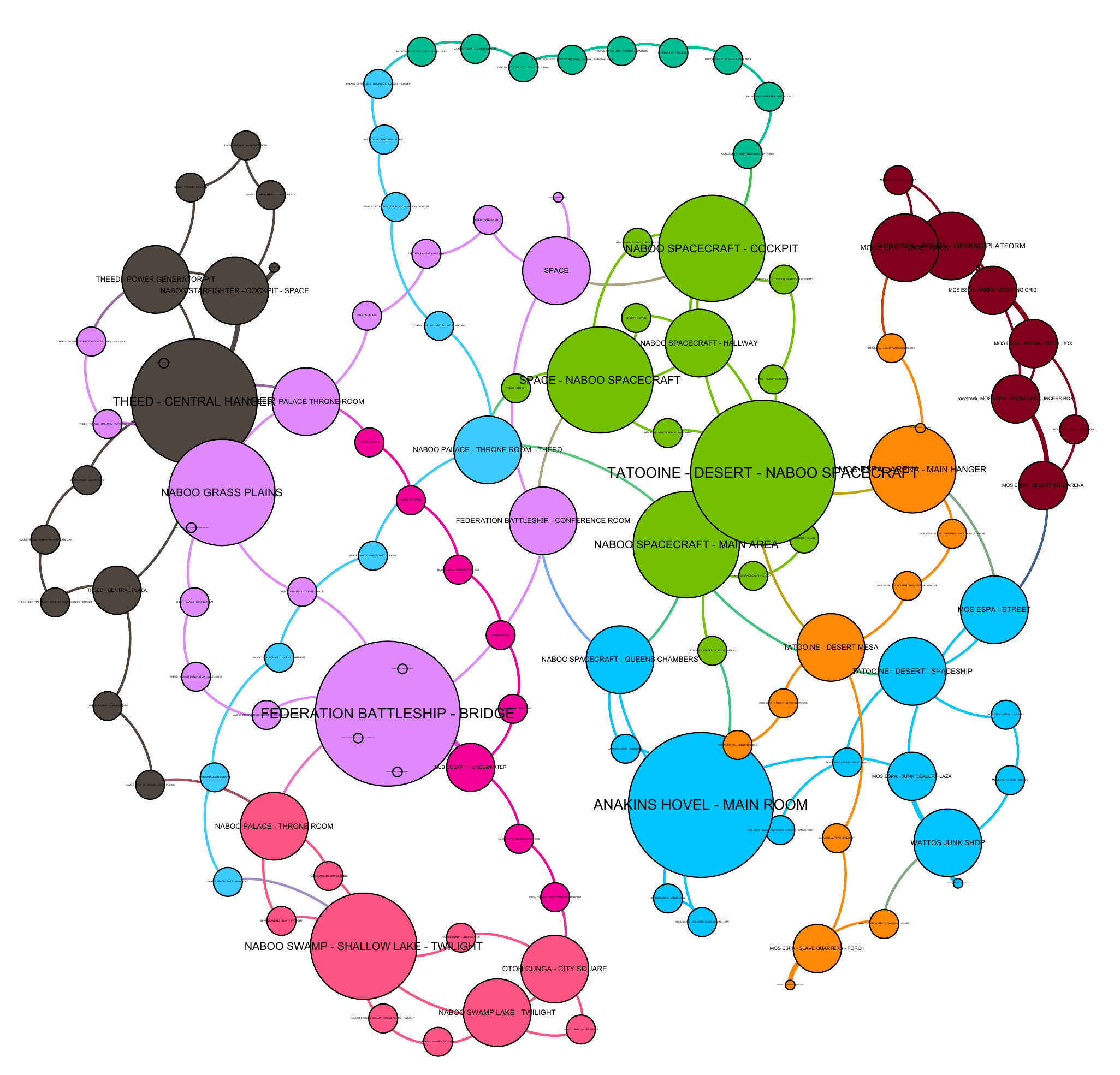}
            \caption[  $G_{LL}$]{$G_{LL}$} 
            
            \label{fig:SW3L}

        \end{subfigure}

\caption{The networks are better seen zoomed on the digital version of this \newline document. Visualization of communities in different layers of Episode I - The Phantom \newline Menace (1999)~\cite{starwars1999episode}. The size of each node corresponds to its degree. (a) The \newline character layer $G_{CC}$. (b) The keyword layer $G_{KK}$. (c) The location layer $G_{LL}$.}
\label{fig:communities}
\end{figure}

\begin{figure}[t!]
\centering
        \begin{subfigure}[b]{0.49\textwidth}   
            \centering 
            \includegraphics[width=\textwidth]{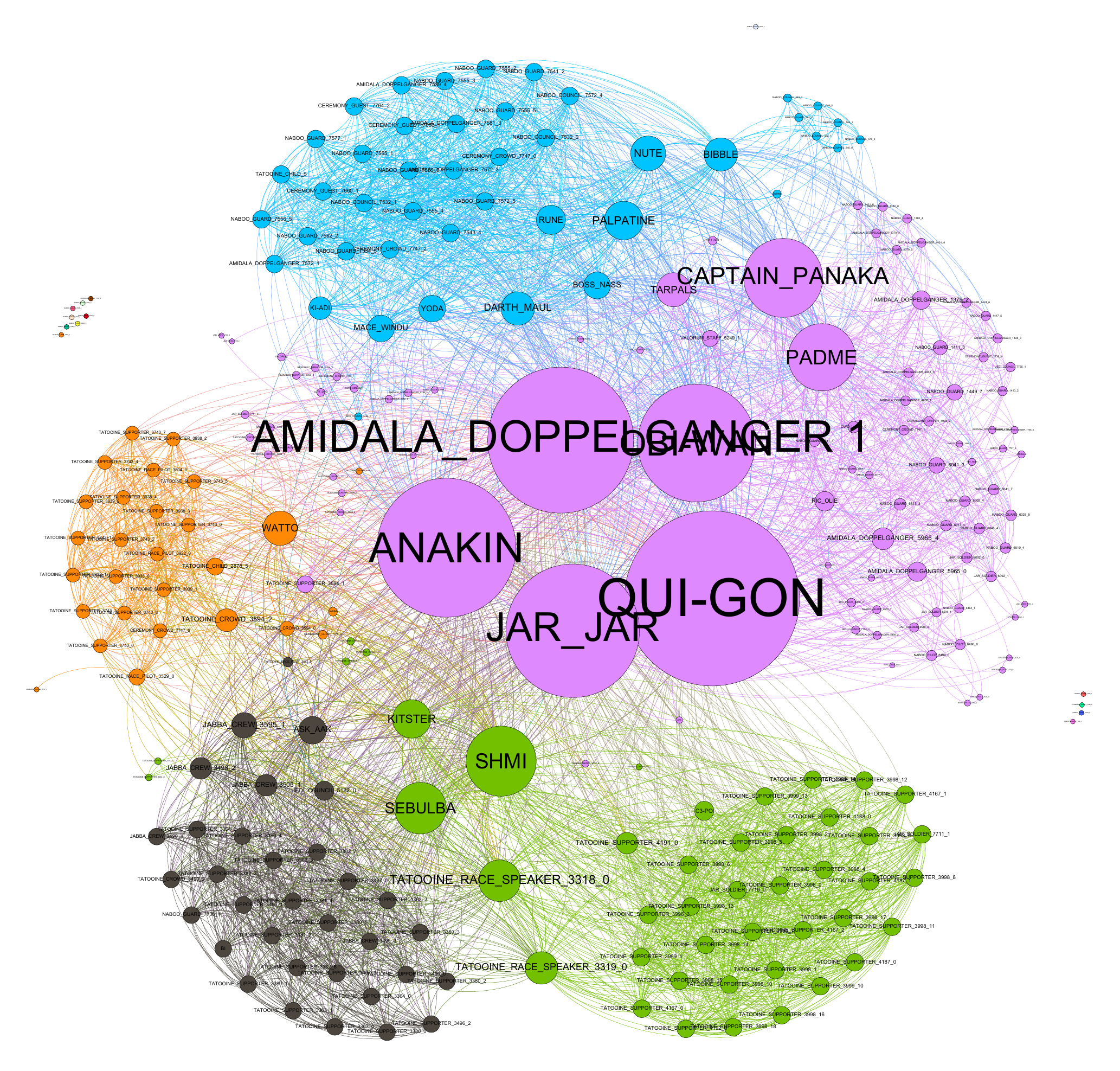}
            \caption[  $G_{FF}$]{$G_{FF}$}
            
            \label{fig:SW3F}
        \end{subfigure}
        \vskip\baselineskip  
        \begin{subfigure}[b]{0.48\textwidth}   
            \centering 
            \includegraphics[width=\textwidth]{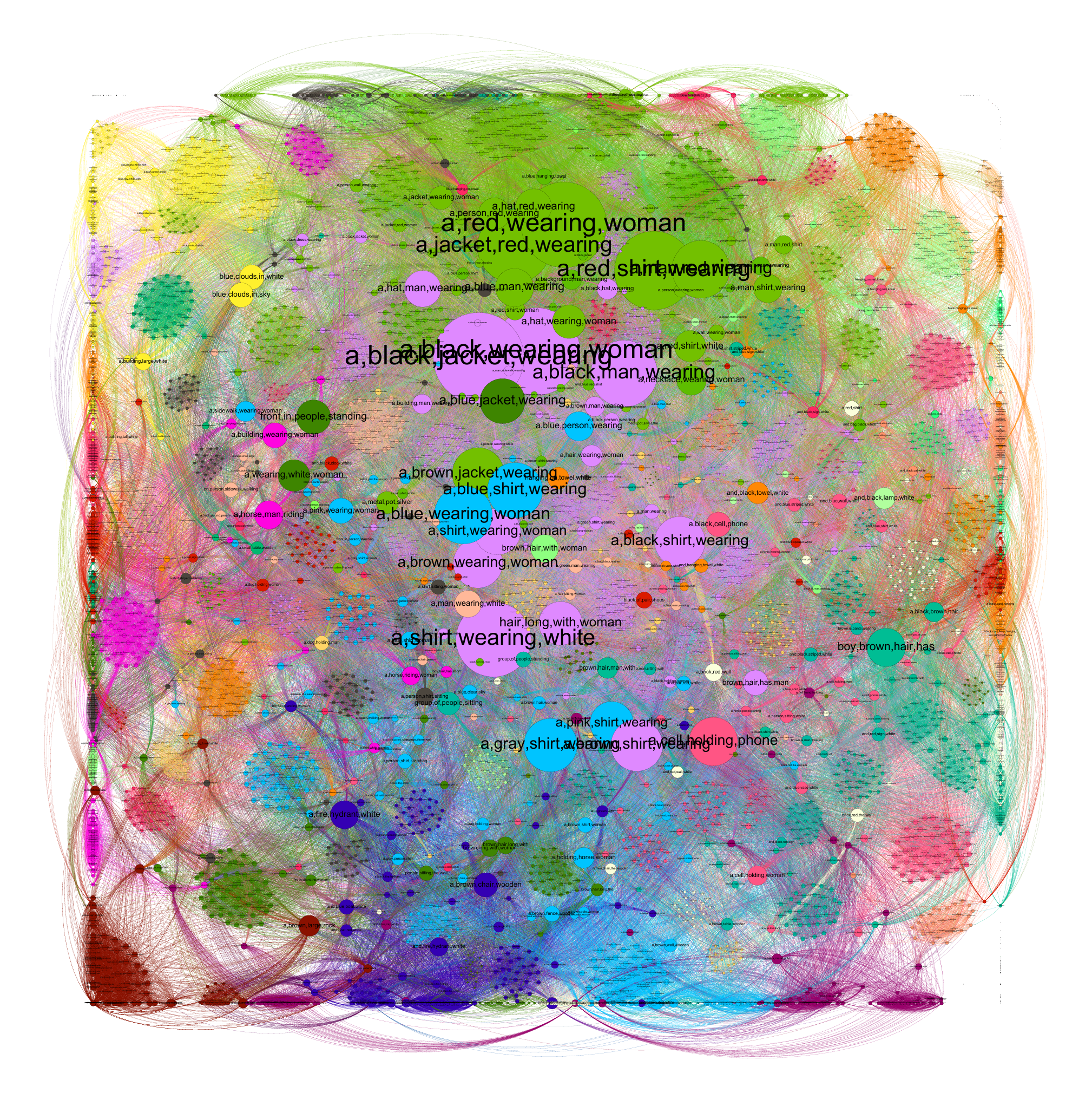}
            \caption[  $G_{CaCa}$]{$G_{CaCa}$}
            
            \label{fig:SW3CA}
        \end{subfigure}
        \qquad 
        \begin{subfigure}[b]{0.48\textwidth}   
            \centering 
            \includegraphics[width=\textwidth]{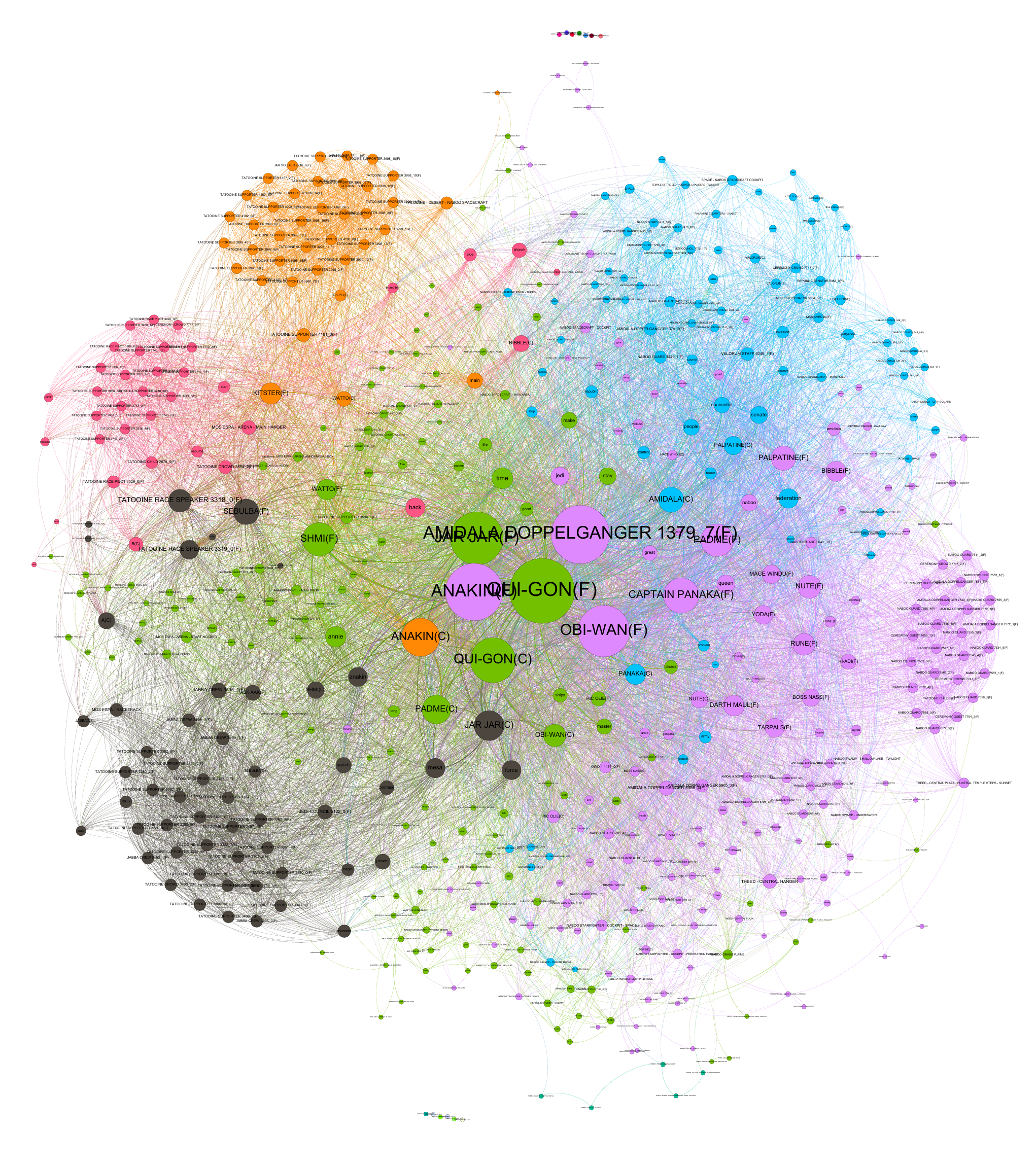}
            \caption[  $\mathbb G'$] {$\mathbb G'$}
            
            \label{fig:SW3M}
        \end{subfigure}

\caption{The networks are better seen zoomed on the digital version of this \newline document. Visualization of communities in different layers of Episode I - The Phantom \newline Menace (1999)~\cite{starwars1999episode} The size of each node corresponds to its degree. (a) The face \newline layer $G_{FF}$. (b) The caption layer $G_{CaCa}$. (c) The multilayer without captions $\mathbb G'$, with \newline the node label encoding: CHARACTER(C), FACE(F), keyword,  and LOCATION-.}
\label{fig:communities}
\end{figure}

\begin{figure}[t!]
\centering
        \begin{subfigure}[b]{0.49\textwidth}
            \centering
            \includegraphics[width=\textwidth]{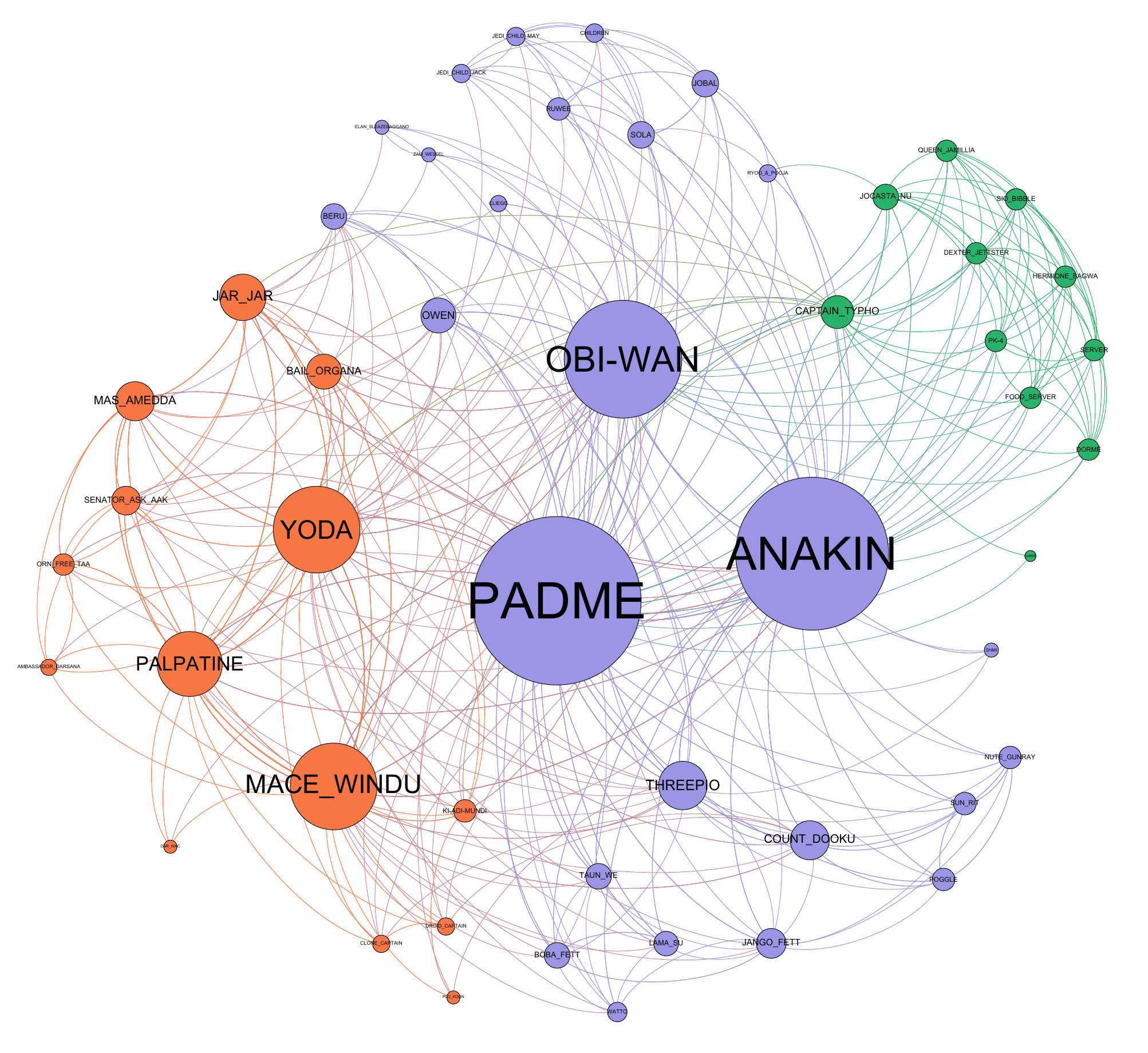}
            \caption[  $G_{CC}$] {$G_{CC}$}
            
            \label{fig:SW3C}
        \end{subfigure}
        \qquad        
        \begin{subfigure}[b]{0.48\textwidth}  
            \centering 
            \includegraphics[width=\textwidth]{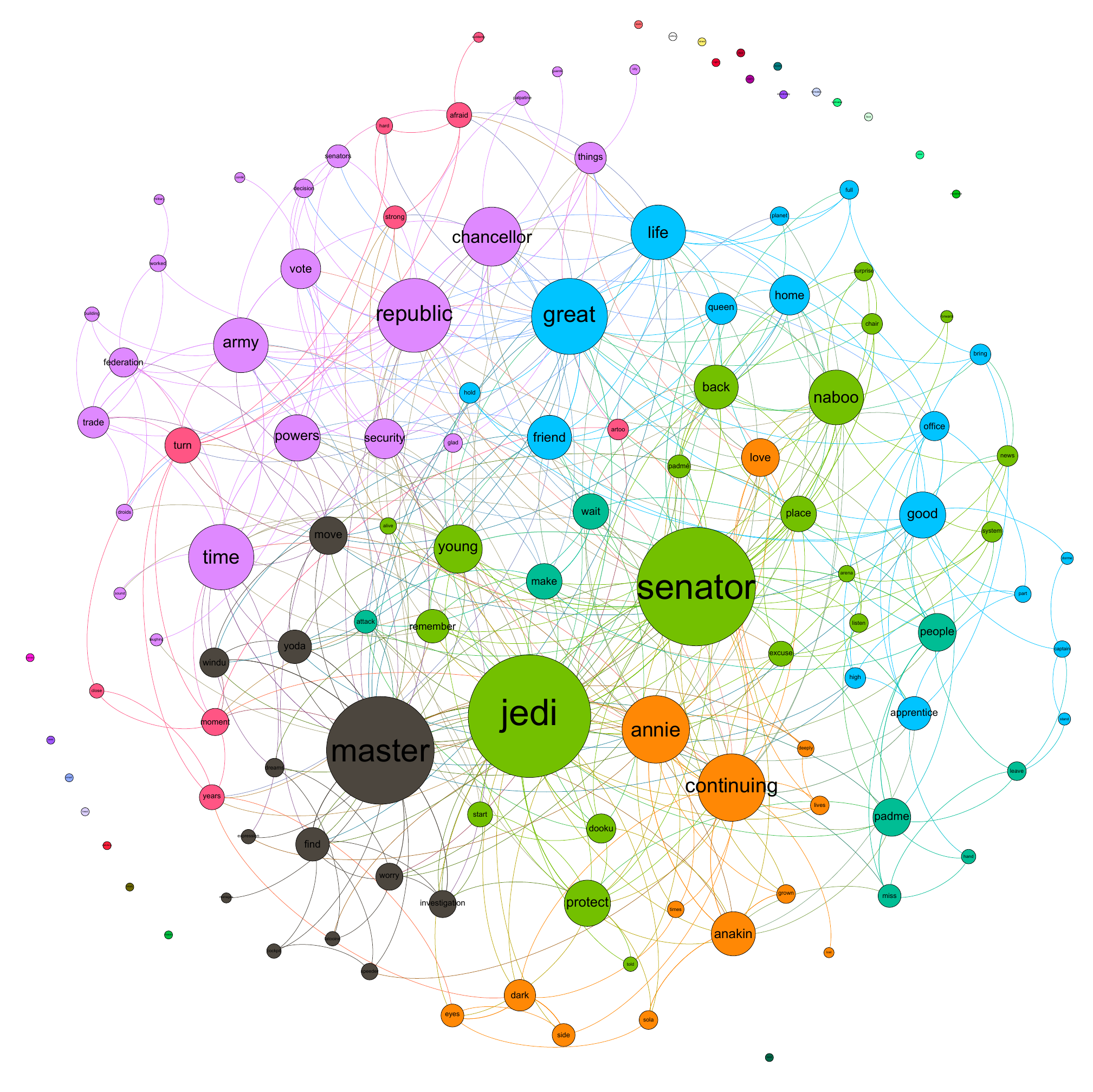}
            \caption[  $G_{KK}$]{$G_{KK}$}
            
            \label{fig:SW3K}
        \end{subfigure}
         \vskip\baselineskip    
        \begin{subfigure}[b]{0.48\textwidth}  
            \centering 
            \includegraphics[width=\textwidth]{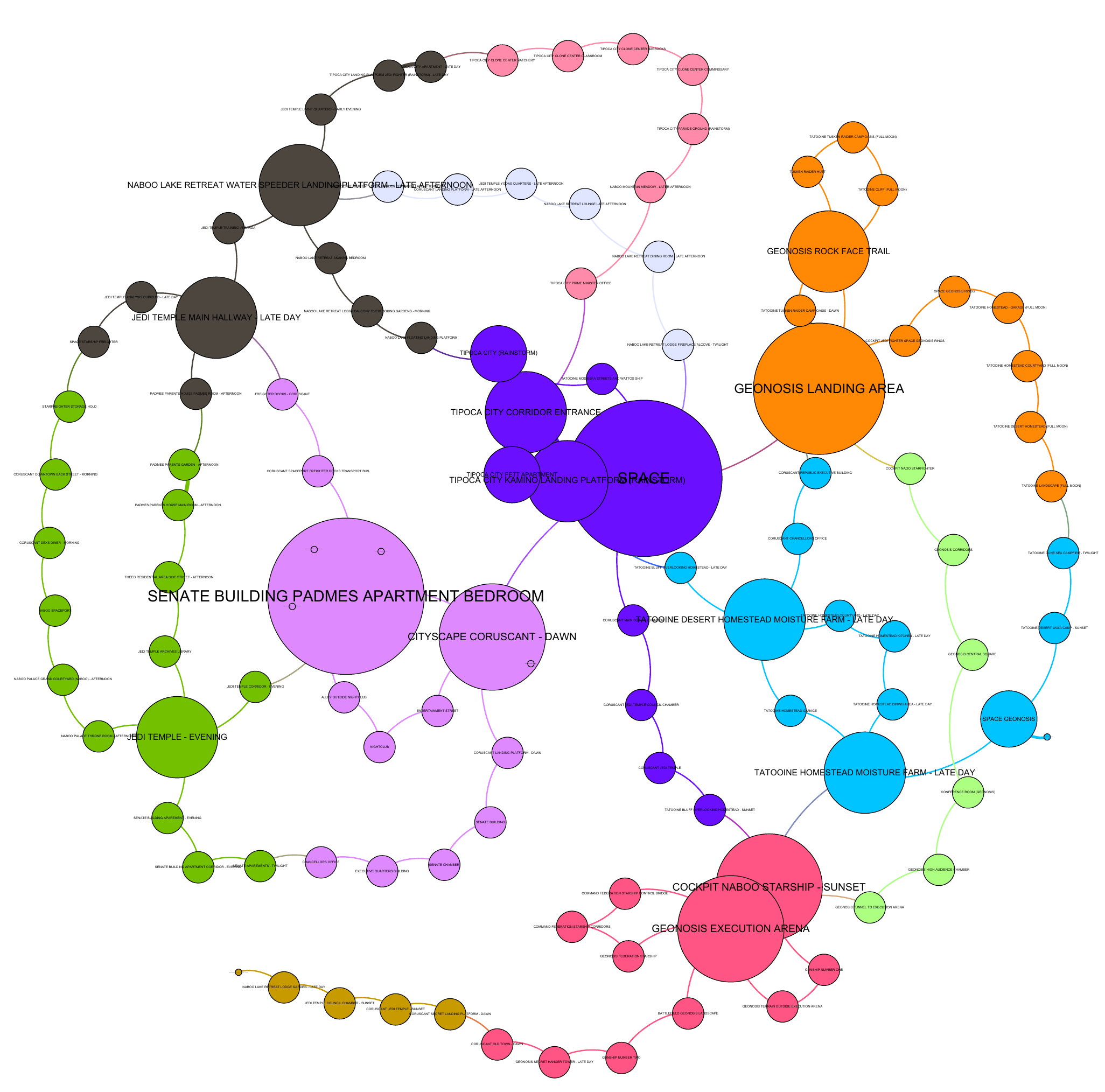}
            \caption[  $G_{LL}$]{$G_{LL}$} 
            
            \label{fig:SW3L}

        \end{subfigure}

\caption{The networks are better seen zoomed on the digital version of this \newline document. Visualization of communities in different layers of Episode II - Attack \newline of the Clones (2002)~\cite{starwars2002episode}. The size of each node corresponds to its degree. (a) The \newline character layer $G_{CC}$. (b) The keyword layer $G_{KK}$. (c) The location layer $G_{LL}$.}
\label{fig:communities}
\end{figure}

\begin{figure}[t!]
\centering
        \begin{subfigure}[b]{0.49\textwidth}   
            \centering 
            \includegraphics[width=\textwidth]{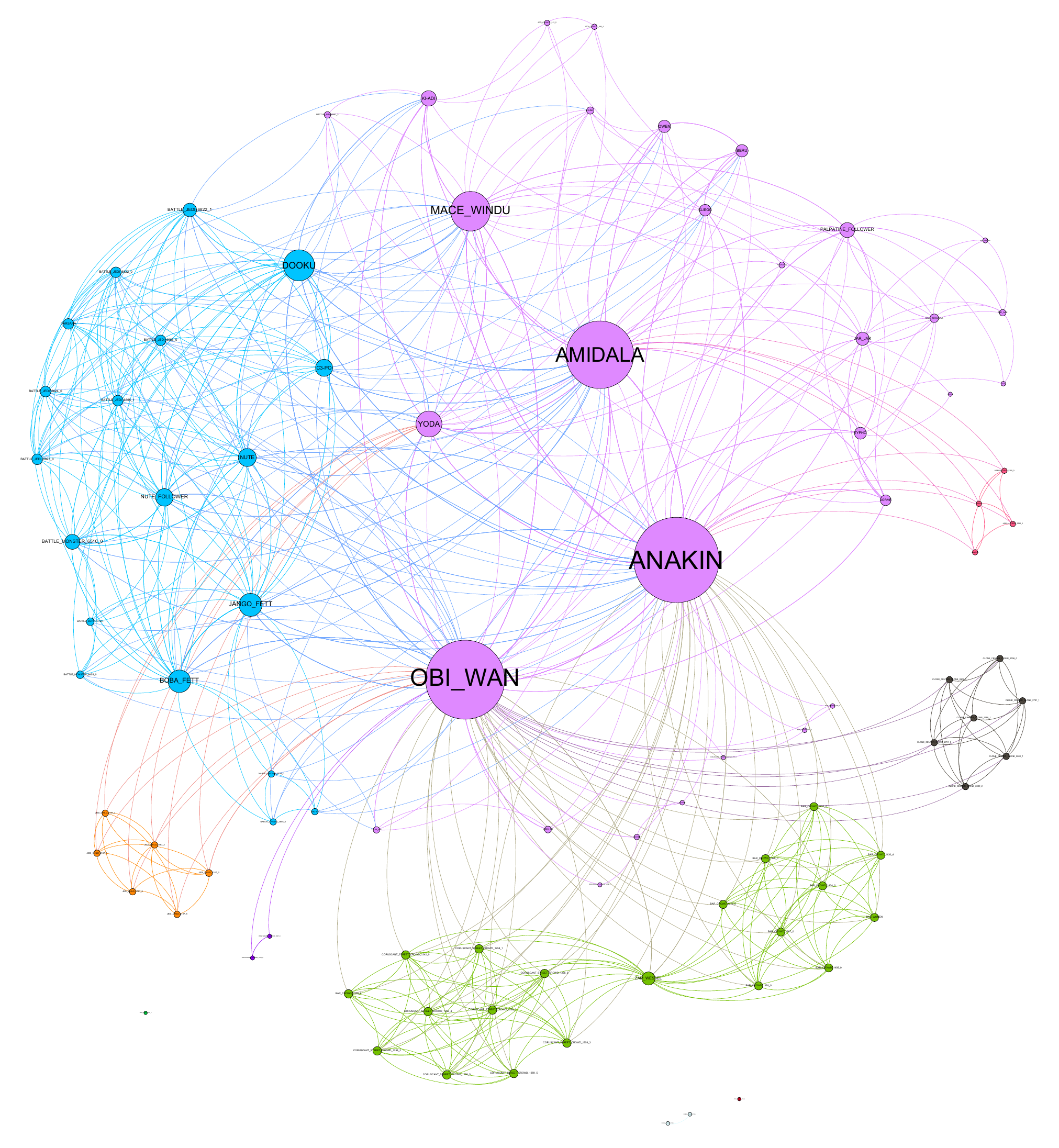}
            \caption[  $G_{FF}$]{$G_{FF}$}
            
            \label{fig:SW3F}
        \end{subfigure}
        \vskip\baselineskip  
        \begin{subfigure}[b]{0.48\textwidth}   
            \centering 
            \includegraphics[width=\textwidth]{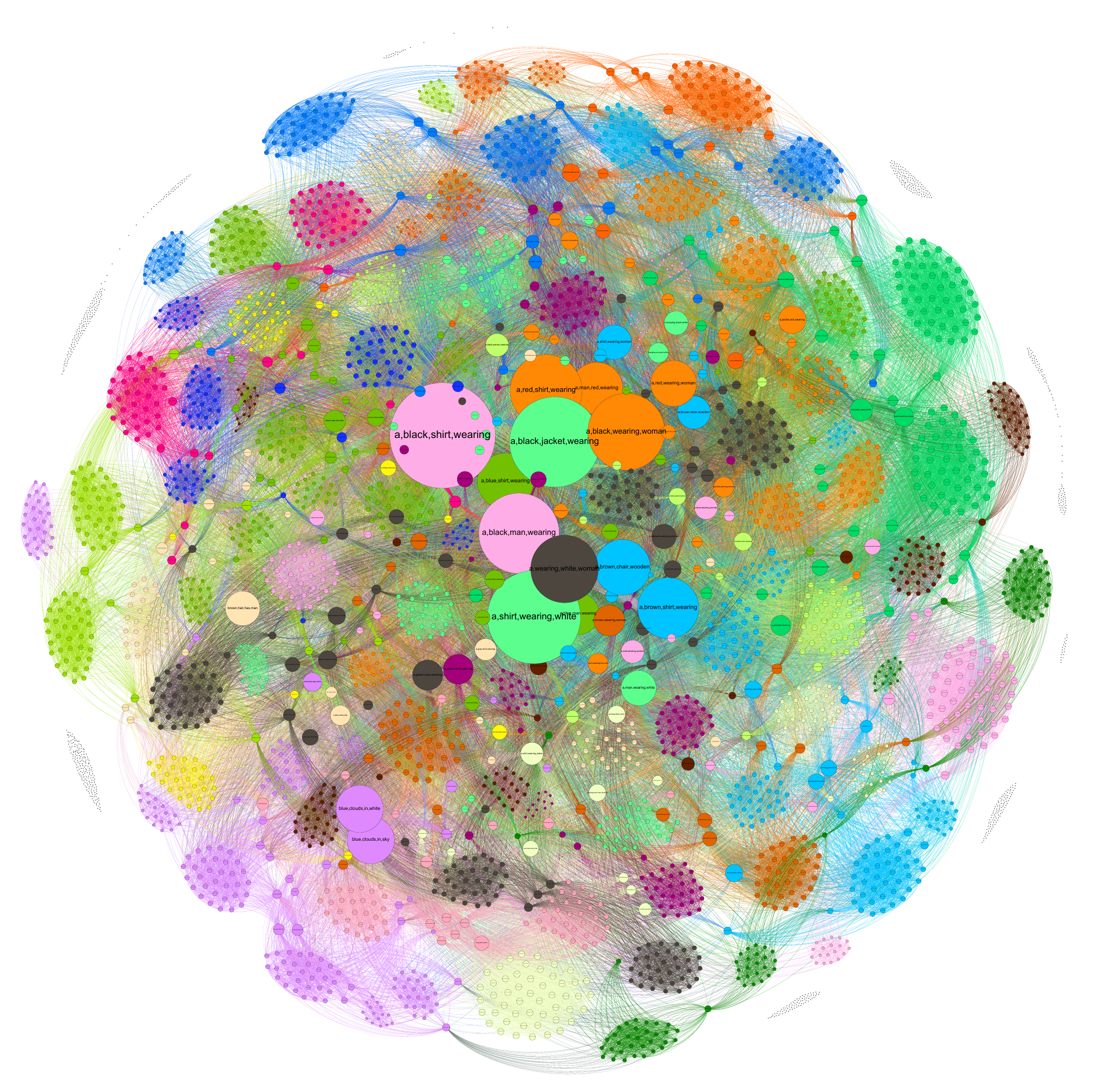}
            \caption[  $G_{CaCa}$]{$G_{CaCa}$}
            
            \label{fig:SW3CA}
        \end{subfigure}
        \qquad 
        \begin{subfigure}[b]{0.48\textwidth}   
            \centering 
            \includegraphics[width=\textwidth]{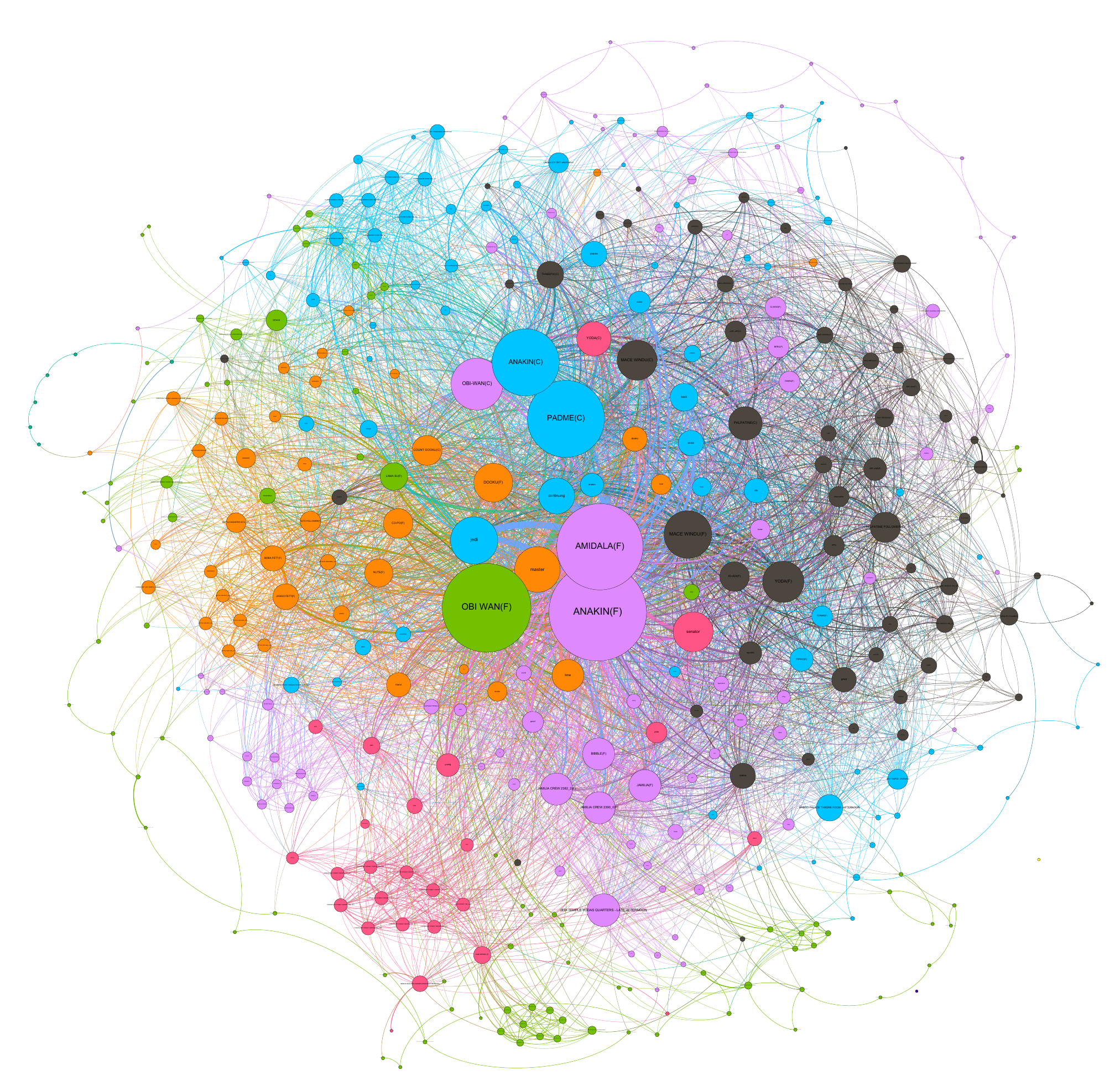}
            \caption[  $\mathbb G'$] {$\mathbb G'$}
            
            \label{fig:SW3M}
        \end{subfigure}

\caption{The networks are better seen zoomed on the digital version of this \newline document. Visualization of communities in different layers of Episode II - Attack \newline of the Clones (2002)~\cite{starwars2002episode}. The size of each node corresponds to its degree. (a) The face \newline layer $G_{FF}$. (b) The caption layer $G_{CaCa}$. (c) The multilayer without captions $\mathbb G'$, with \newline the node label encoding: CHARACTER(C), FACE(F), keyword,  and LOCATION-.}
\label{fig:communities}
\end{figure}

\begin{figure}[t!]
\centering
        \begin{subfigure}[b]{0.49\textwidth}
            \centering
            \includegraphics[width=\textwidth]{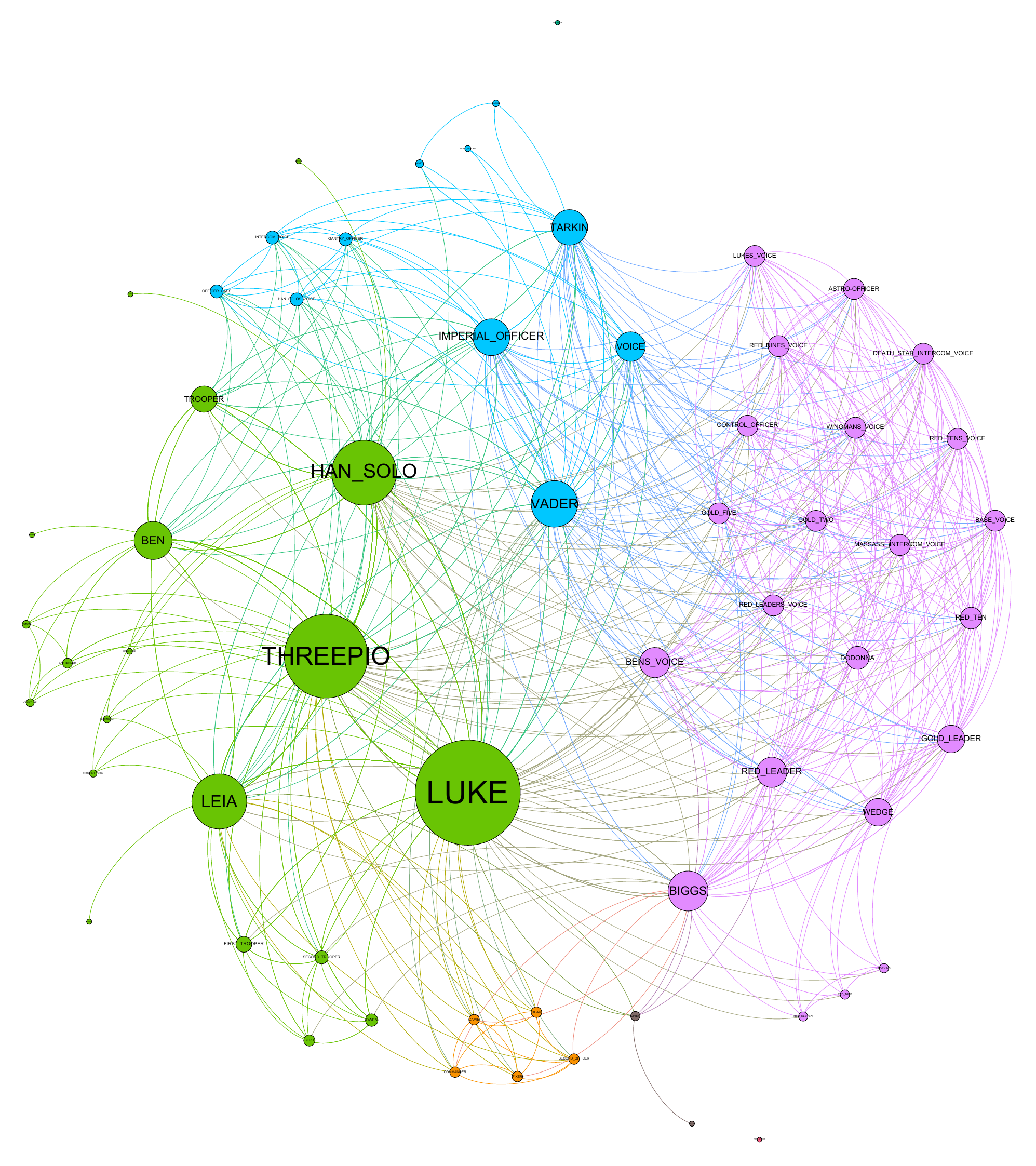}
            \caption[  $G_{CC}$] {$G_{CC}$}
            
            \label{fig:SW3C}
        \end{subfigure}
        \qquad        
        \begin{subfigure}[b]{0.48\textwidth}  
            \centering 
            \includegraphics[width=\textwidth]{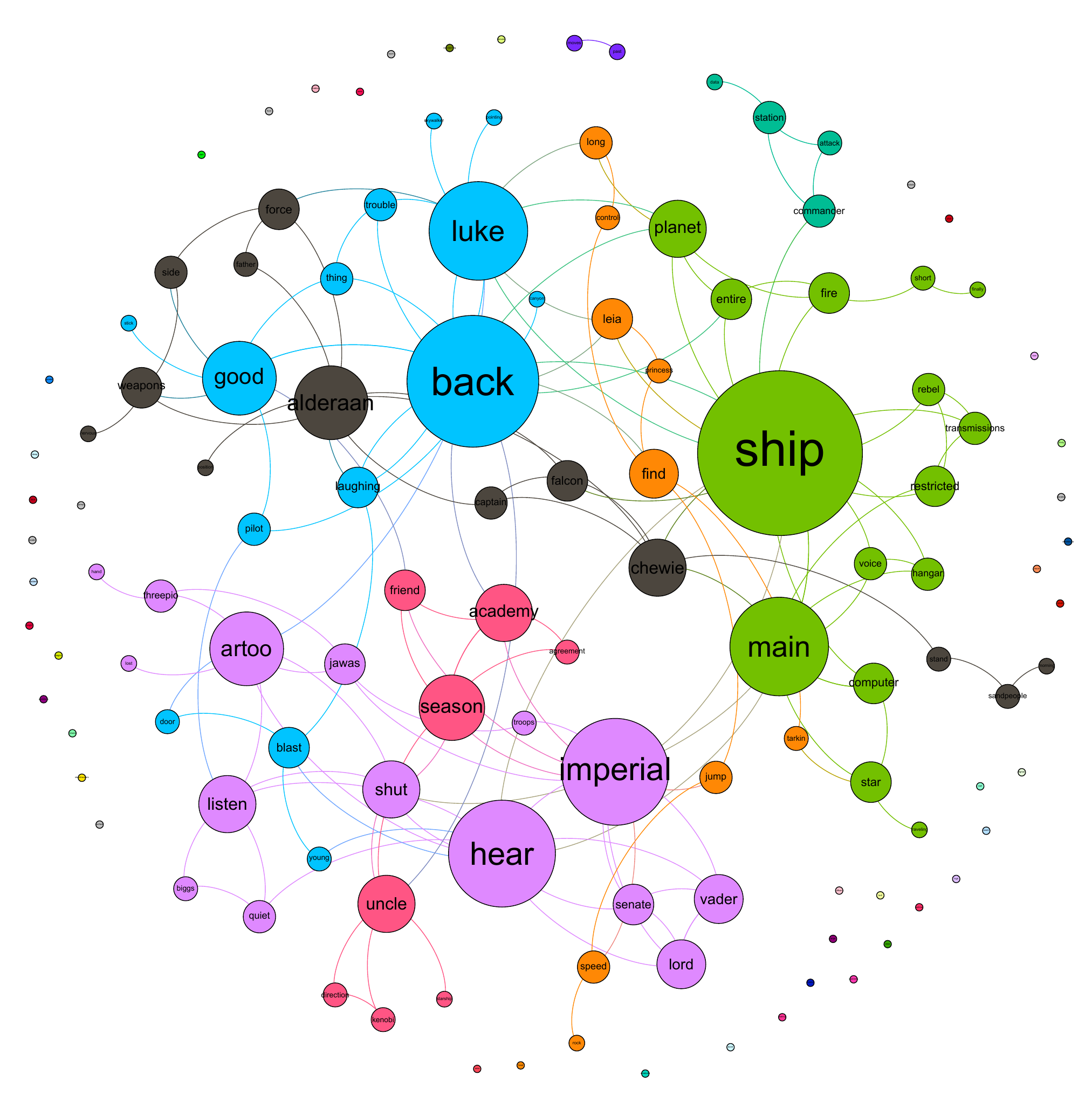}
            \caption[  $G_{KK}$]{$G_{KK}$}
            
            \label{fig:SW3K}
        \end{subfigure}
         \vskip\baselineskip    
        \begin{subfigure}[b]{0.48\textwidth}  
            \centering 
            \includegraphics[width=\textwidth]{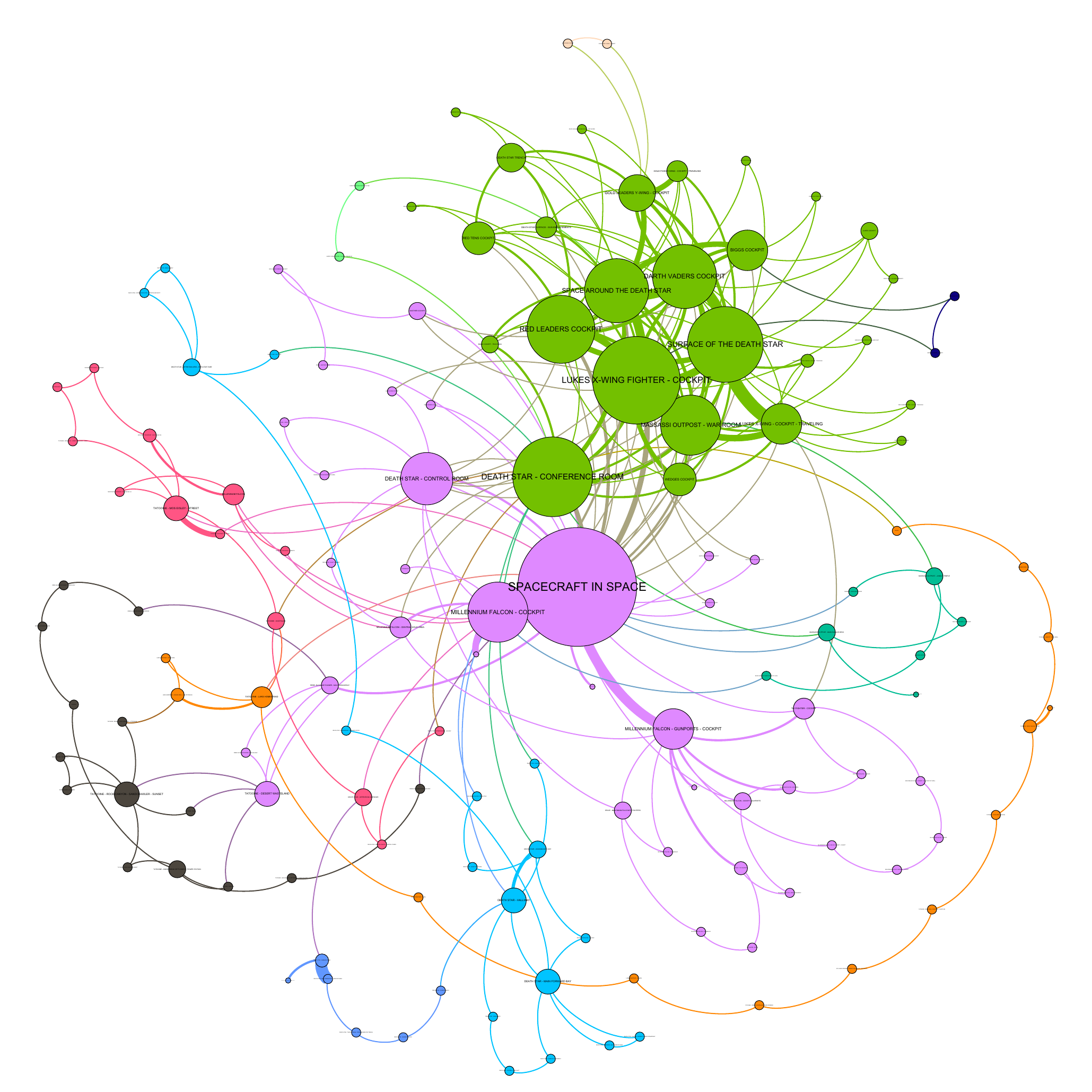}
            \caption[  $G_{LL}$]{$G_{LL}$} 
            
            \label{fig:SW3L}

        \end{subfigure}

\caption{The networks are better seen zoomed on the digital version of this \newline document. Visualization of communities in different layers of Episode IV - A New \newline Hope (1977)~\cite{starwars1977episode}. The size of each node corresponds to its degree. (a) The character \newline layer $G_{CC}$. (b) The keyword layer $G_{KK}$. (c) The location layer $G_{LL}$.}
\label{fig:communities}
\end{figure}

\begin{figure}[t!]
\centering
        \begin{subfigure}[b]{0.49\textwidth}   
            \centering 
            \includegraphics[width=\textwidth]{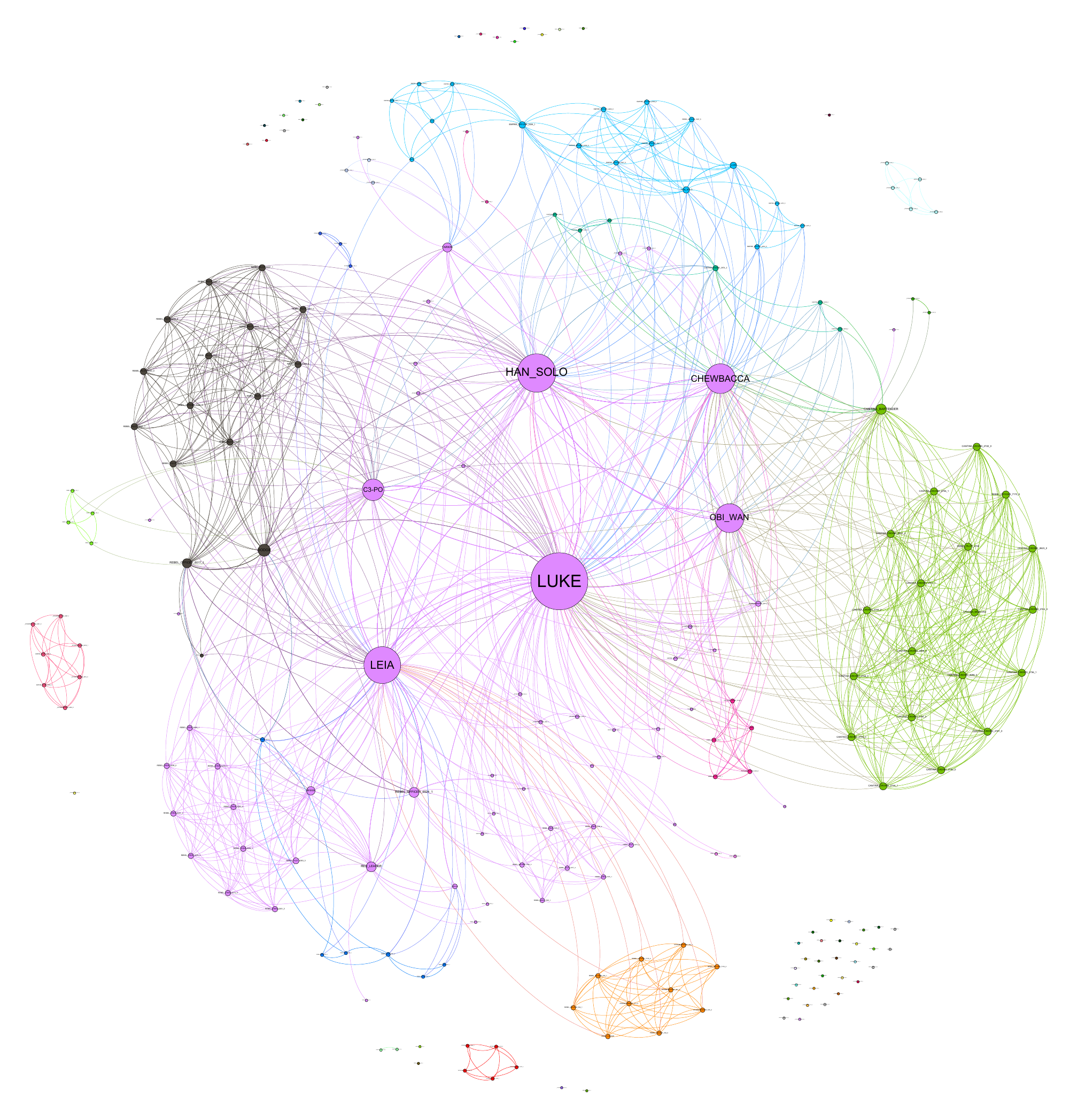}
            \caption[  $G_{FF}$]{$G_{FF}$}
            
            \label{fig:SW3F}
        \end{subfigure}
        \vskip\baselineskip  
        \begin{subfigure}[b]{0.48\textwidth}   
            \centering 
            \includegraphics[width=\textwidth]{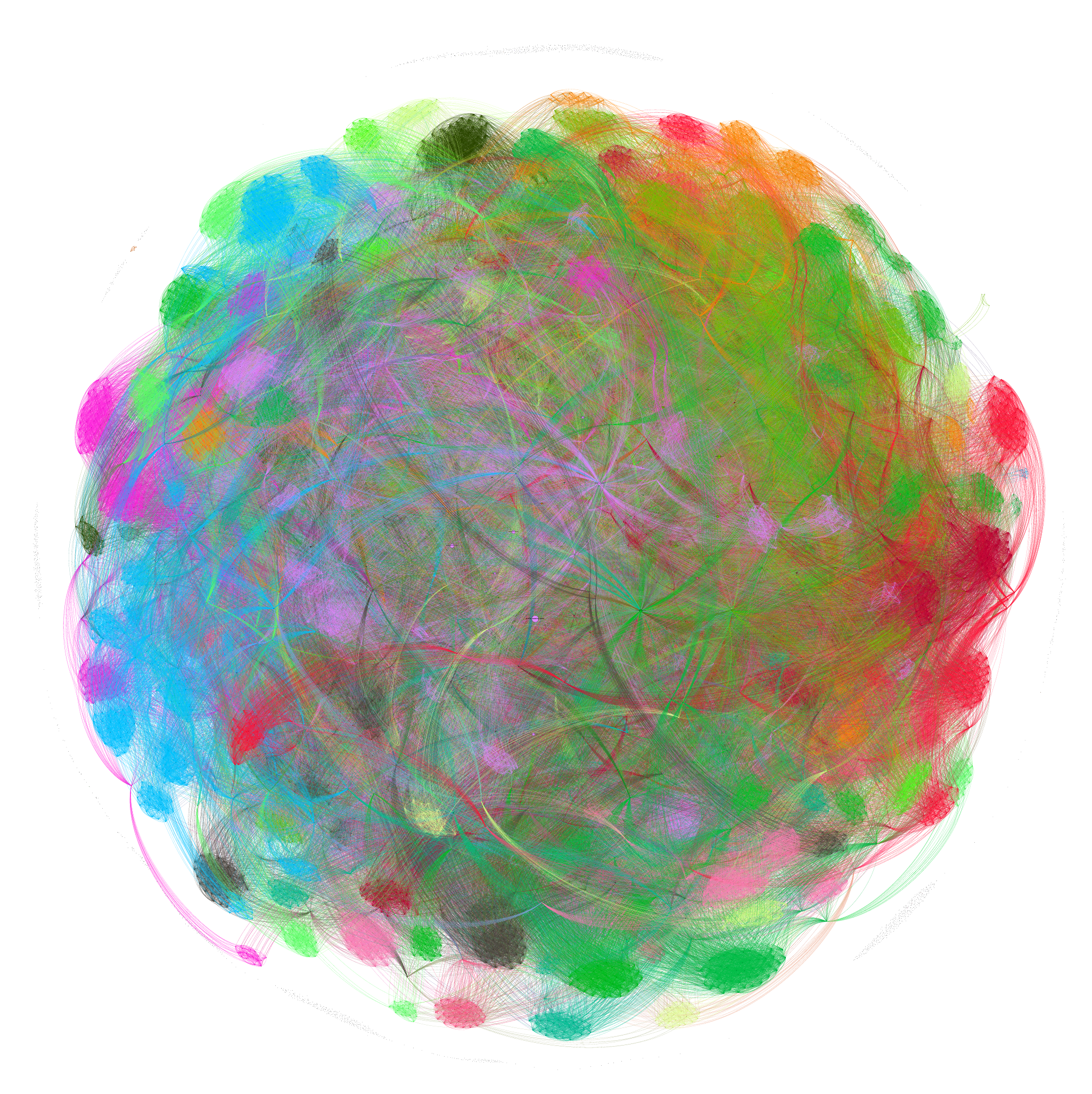}
            \caption[  $G_{CaCa}$]{$G_{CaCa}$}
            
            \label{fig:SW3CA}
        \end{subfigure}
        \qquad 
        \begin{subfigure}[b]{0.48\textwidth}   
            \centering 
            \includegraphics[width=\textwidth]{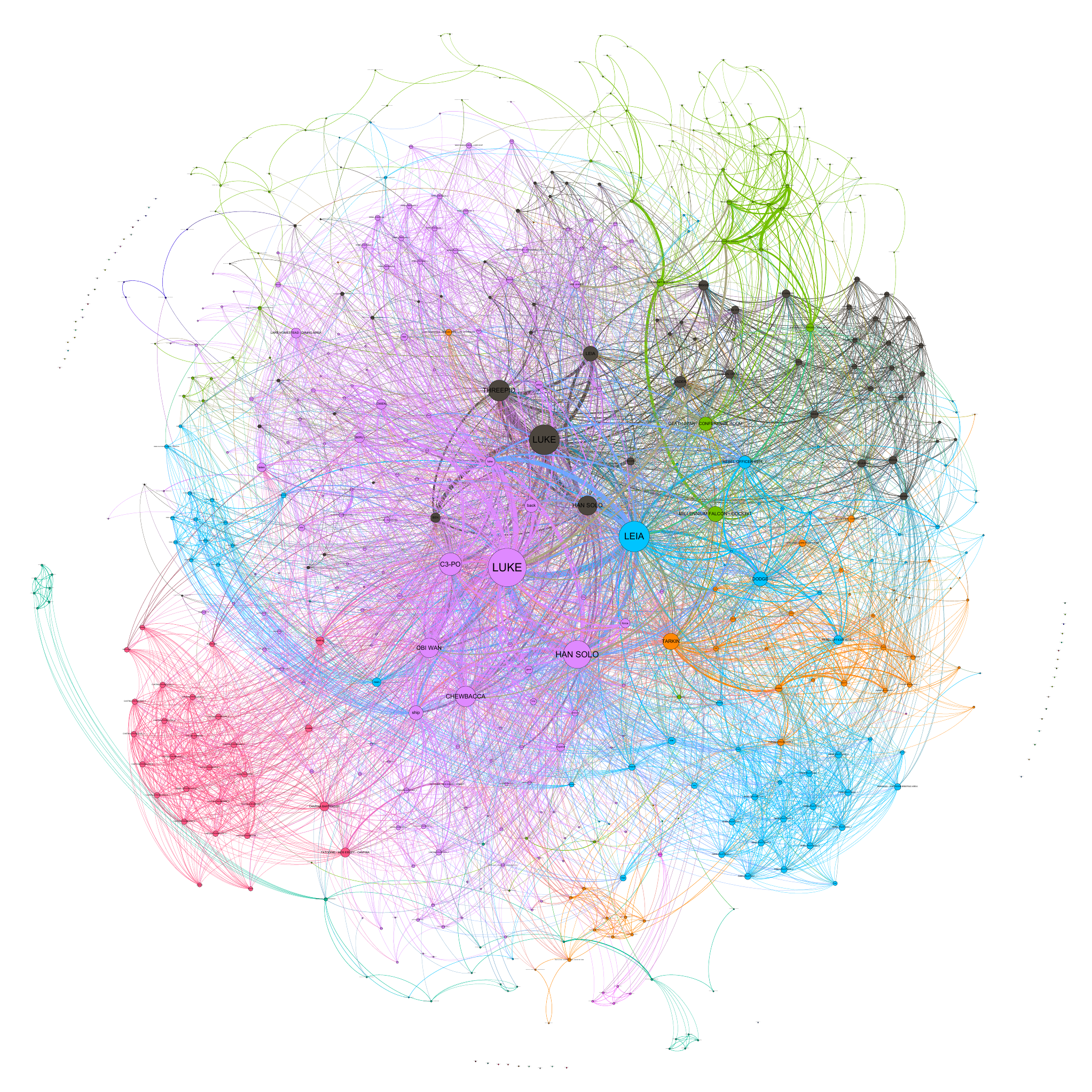}
            \caption[  $\mathbb G'$] {$\mathbb G'$}
            
            \label{fig:SW3M}
        \end{subfigure}

\caption{The networks are better seen zoomed on the digital version of this \newline document. Visualization of communities in different layers of Episode IV - A New \newline Hope (1977)~\cite{starwars1977episode}. The size of each node corresponds to its degree. (a) The face \newline layer $G_{FF}$. (b) The caption layer $G_{CaCa}$. (c) The multilayer without captions $\mathbb G'$, with \newline the node label encoding: CHARACTER(C), FACE(F), keyword,  and LOCATION-.}
\label{fig:communities}
\end{figure}

\begin{figure}[t!]
\centering
        \begin{subfigure}[b]{0.49\textwidth}
            \centering
            \includegraphics[width=\textwidth]{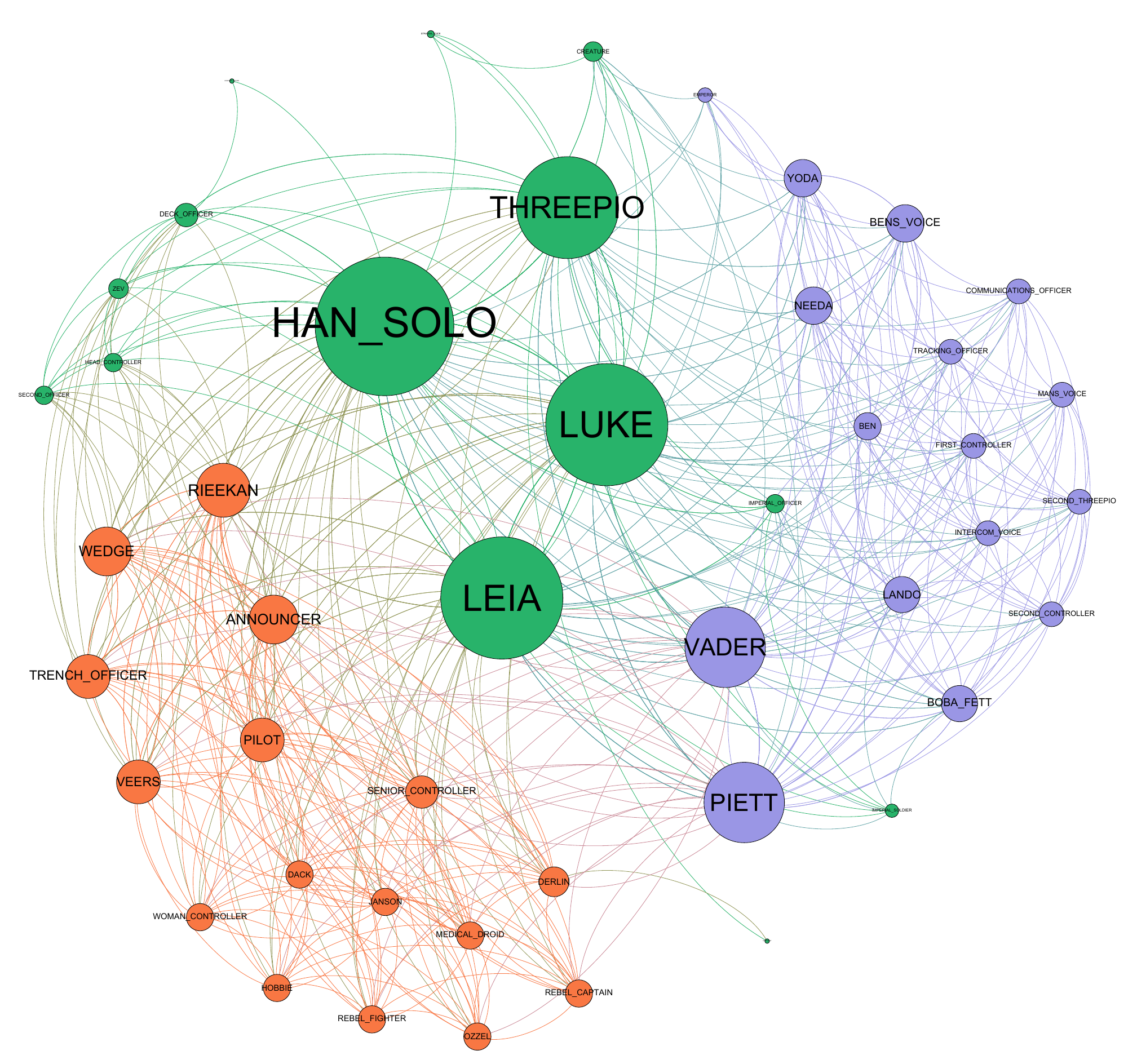}
            \caption[  $G_{CC}$] {$G_{CC}$}
            
            \label{fig:SW3C}
        \end{subfigure}
        \qquad        
        \begin{subfigure}[b]{0.48\textwidth}  
            \centering 
            \includegraphics[width=\textwidth]{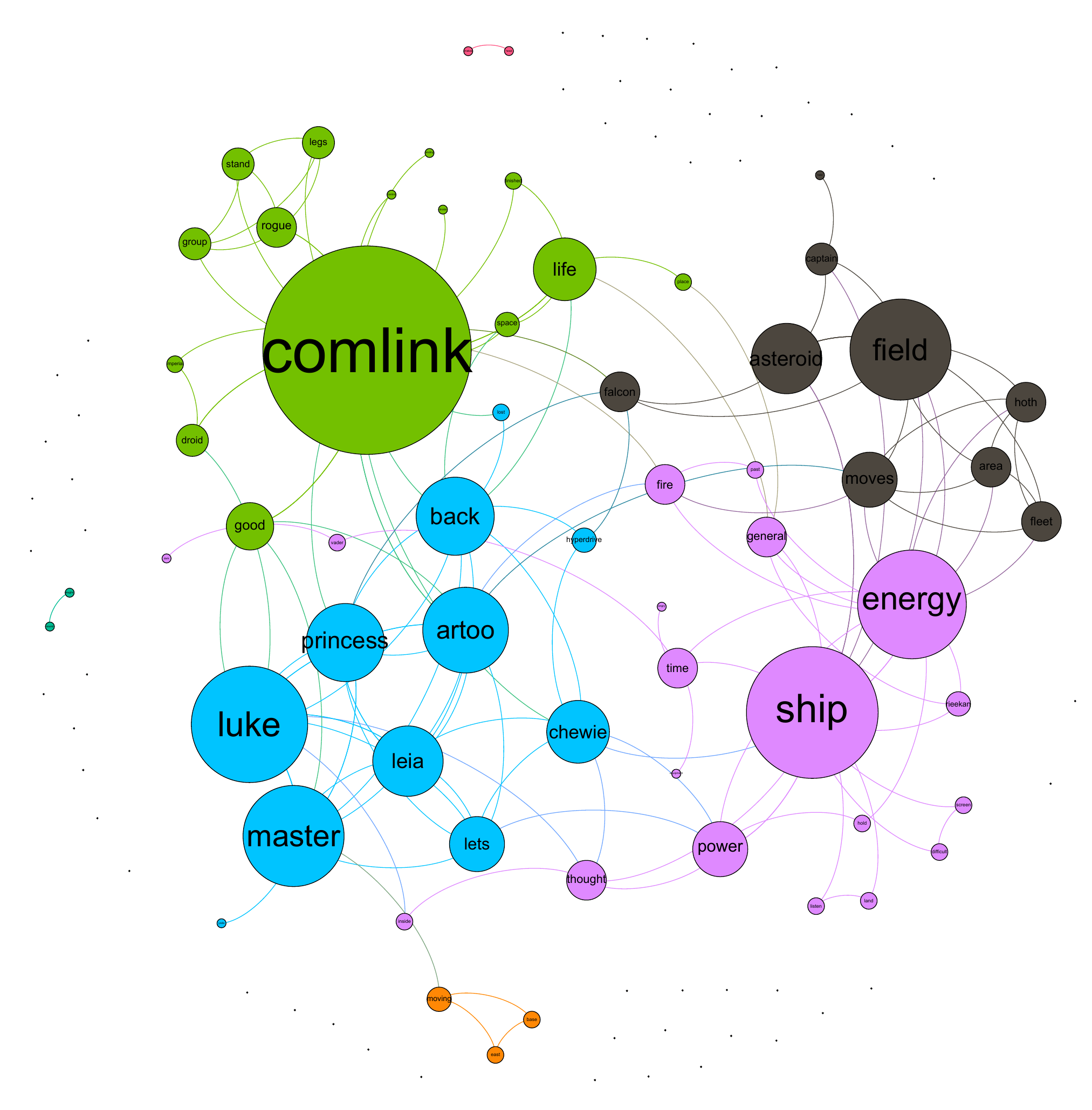}
            \caption[  $G_{KK}$]{$G_{KK}$}
            
            \label{fig:SW3K}
        \end{subfigure}
         \vskip\baselineskip    
        \begin{subfigure}[b]{0.48\textwidth}  
            \centering 
            \includegraphics[width=\textwidth]{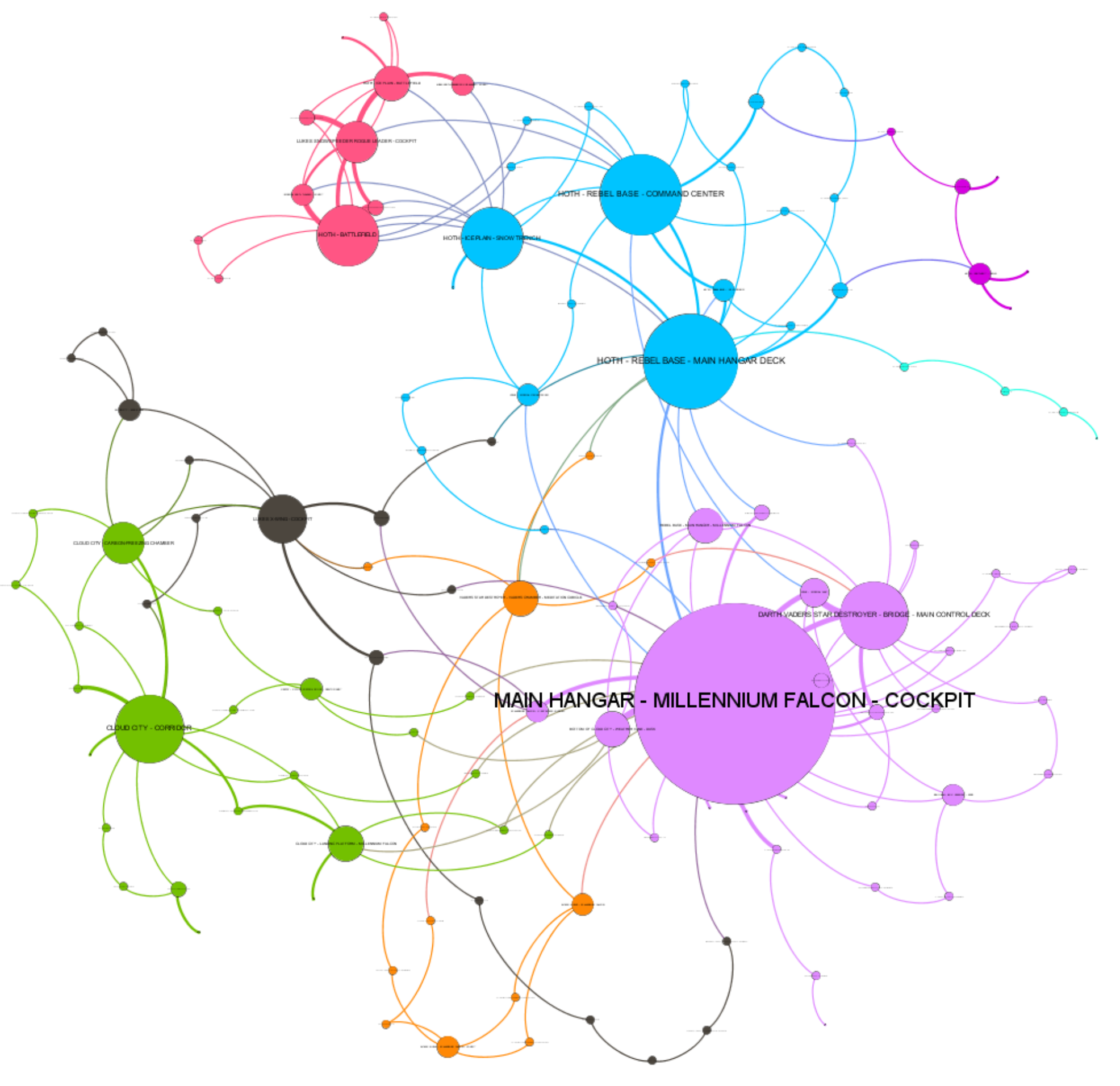}
            \caption[  $G_{LL}$]{$G_{LL}$} 
            
            \label{fig:SW3L}

        \end{subfigure}

\caption{The networks are better seen zoomed on the digital version of this \newline document. Visualization of communities in different layers of Episode V - The Empire \newline Strikes Back (1980)~\cite{starwars1980episode}. The size of each node corresponds to its degree. (a) The \newline character layer $G_{CC}$. (b) The keyword layer $G_{KK}$. (c) The location layer $G_{LL}$.}
\label{fig:communities}
\end{figure}

\begin{figure}[t!]
\centering
        \begin{subfigure}[b]{0.49\textwidth}   
            \centering 
            \includegraphics[width=\textwidth]{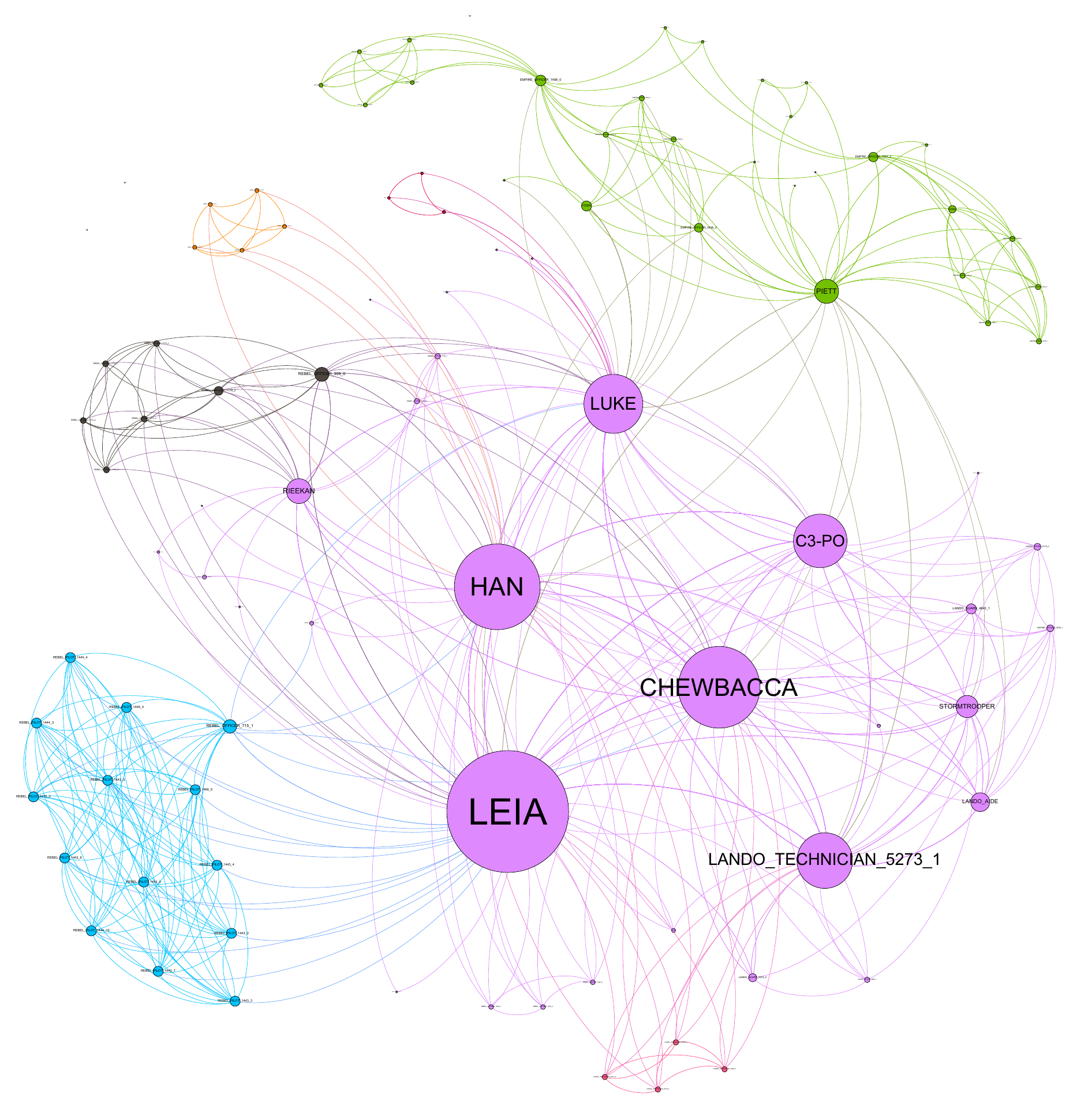}
            \caption[  $G_{FF}$]{$G_{FF}$}
            
            \label{fig:SW3F}
        \end{subfigure}
        \vskip\baselineskip  
        \begin{subfigure}[b]{0.48\textwidth}   
            \centering 
            \includegraphics[width=\textwidth]{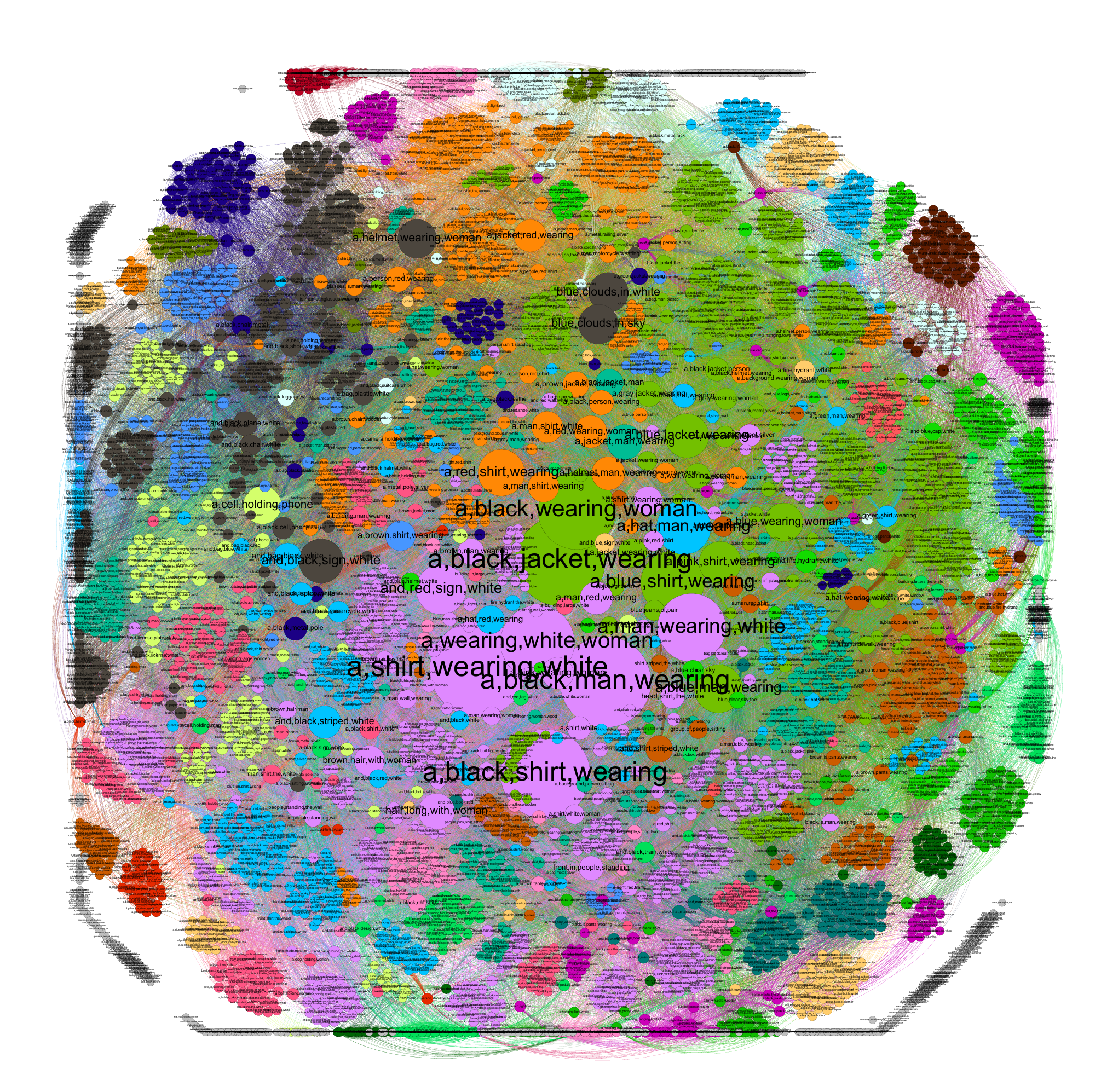}
            \caption[  $G_{CaCa}$]{$G_{CaCa}$}
            
            \label{fig:SW3CA}
        \end{subfigure}
        \qquad 
        \begin{subfigure}[b]{0.48\textwidth}   
            \centering 
            \includegraphics[width=\textwidth]{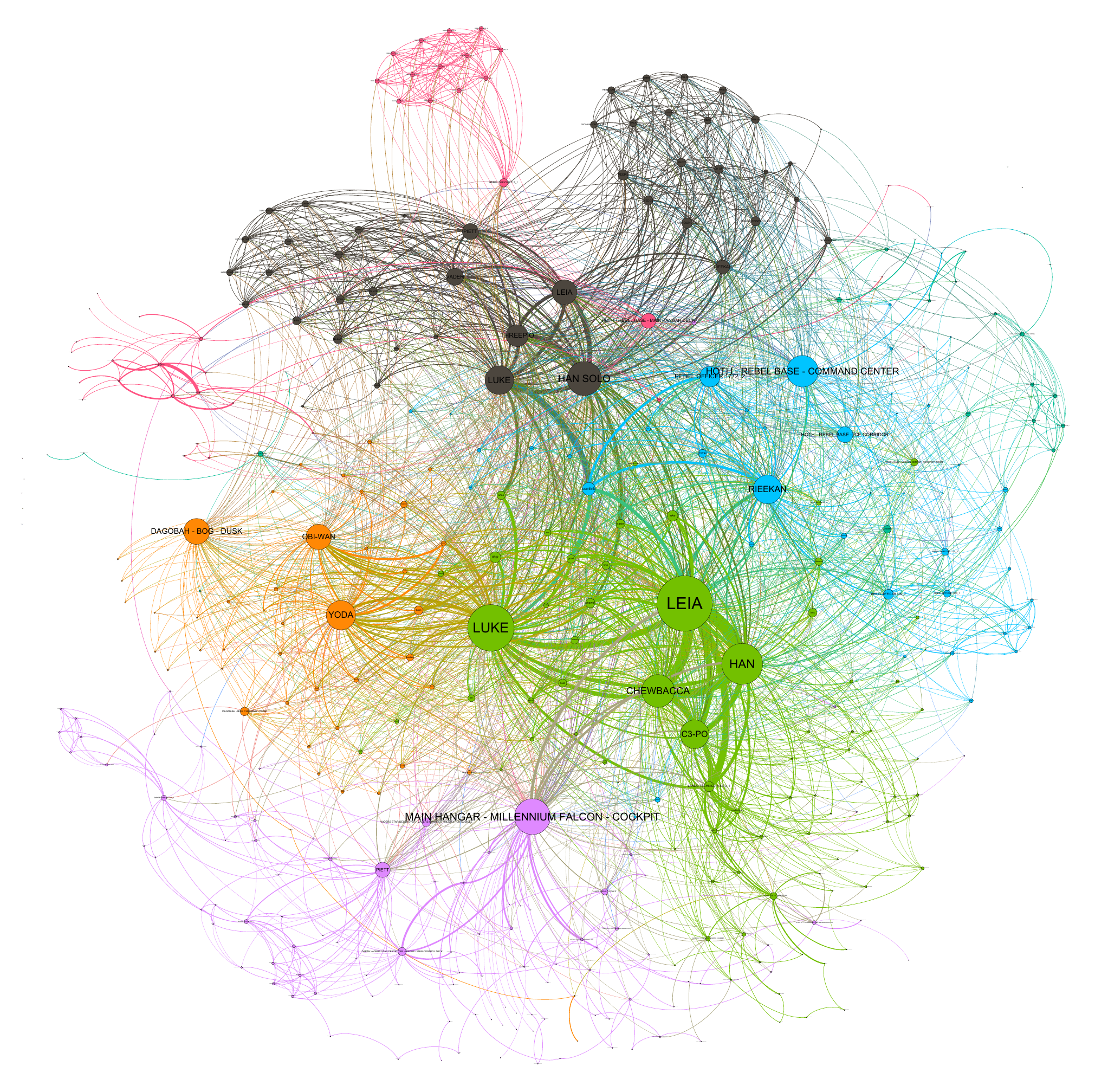}
            \caption[  $\mathbb G'$] {$\mathbb G'$}
            
            \label{fig:SW3M}
        \end{subfigure}

\caption{The networks are better seen zoomed on the digital version of this \newline document. Visualization of communities in different layers of Episode V - The Empire \newline Strikes Back (1980)~\cite{starwars1980episode}. The size of each node corresponds to its degree. (a) The face \newline layer $G_{FF}$. (b) The caption layer $G_{CaCa}$. (c) The multilayer without captions $\mathbb G'$, with \newline the node label encoding: CHARACTER(C), FACE(F), keyword,  and LOCATION-.}
\label{fig:communities}
\end{figure}

\begin{figure}[t!]
\centering
        \begin{subfigure}[b]{0.49\textwidth}
            \centering
            \includegraphics[width=\textwidth]{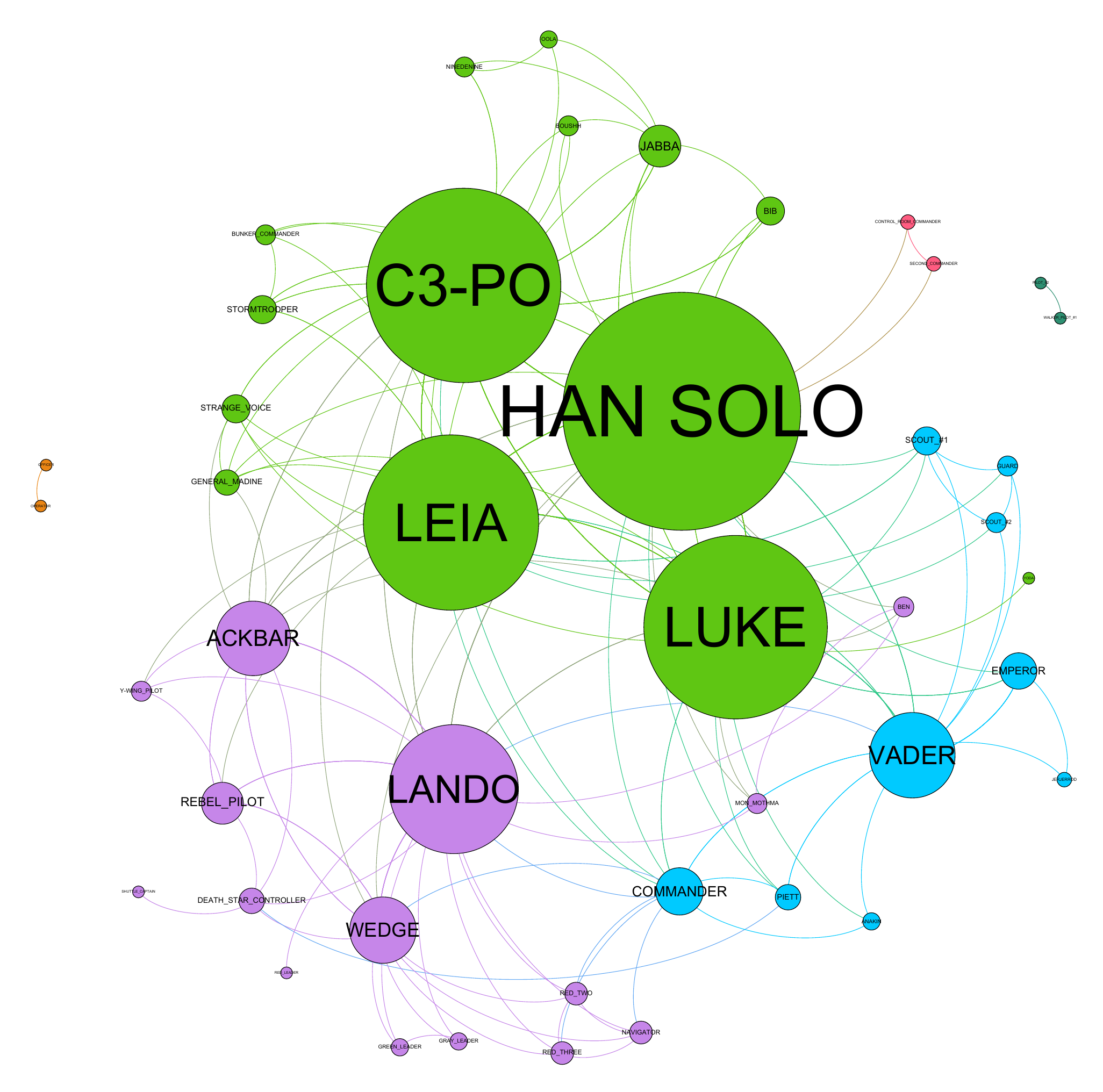}
            \caption[  $G_{CC}$] {$G_{CC}$}
            
            \label{fig:SW3C}
        \end{subfigure}
        \qquad        
        \begin{subfigure}[b]{0.48\textwidth}  
            \centering 
            \includegraphics[width=\textwidth]{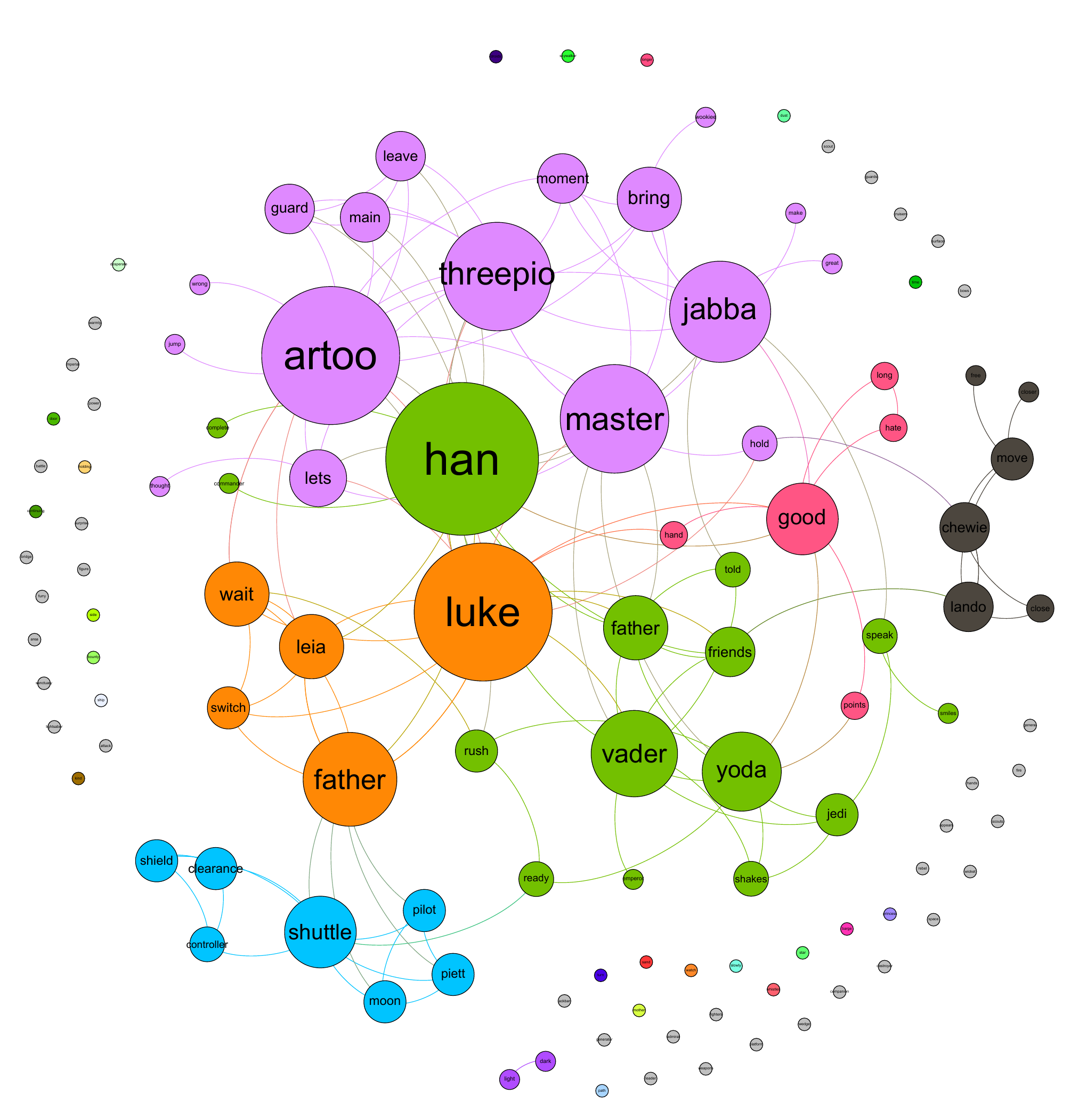}
            \caption[  $G_{KK}$]{$G_{KK}$}
            
            \label{fig:SW3K}
        \end{subfigure}
         \vskip\baselineskip    
        \begin{subfigure}[b]{0.48\textwidth}  
            \centering 
            \includegraphics[width=\textwidth]{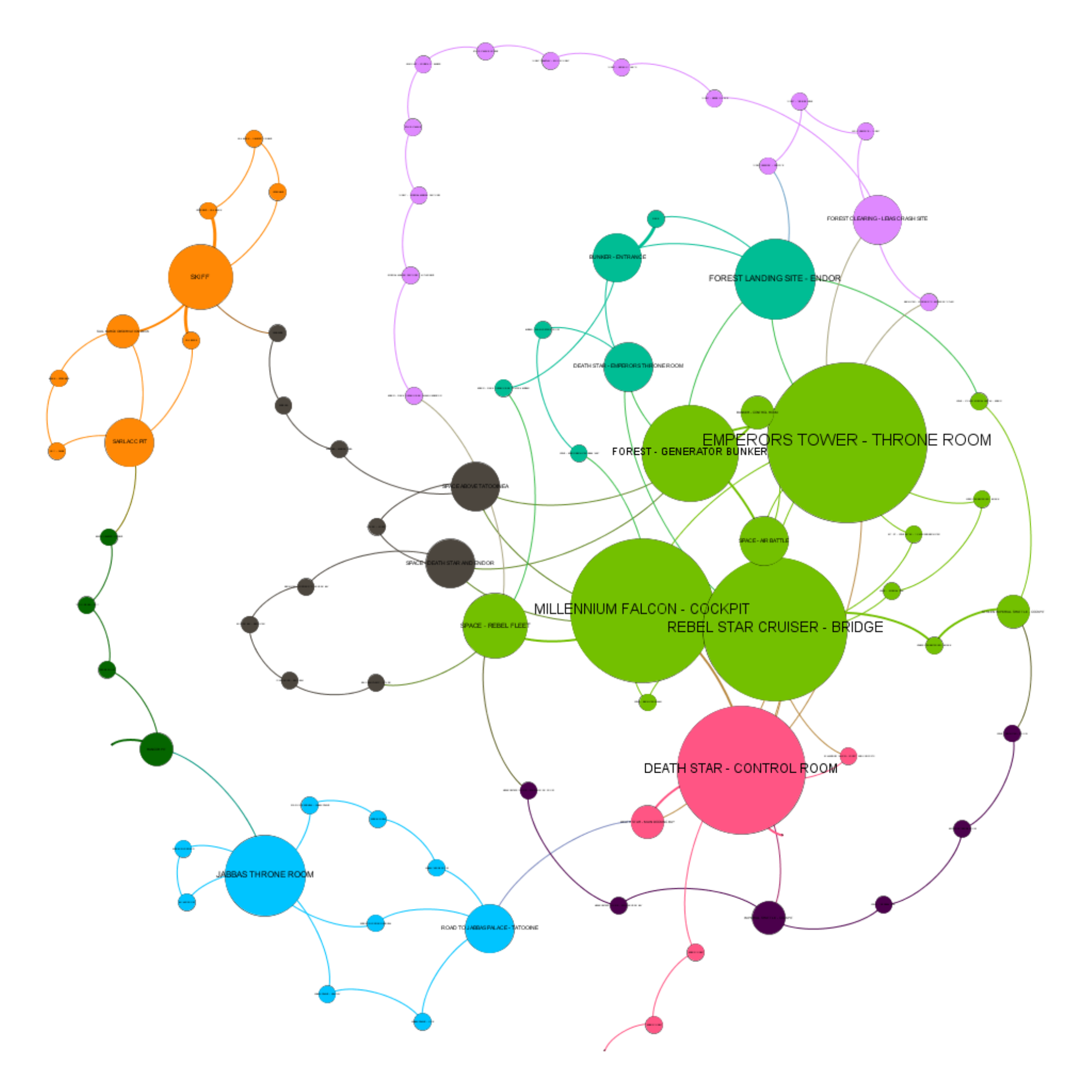}
            \caption[  $G_{LL}$]{$G_{LL}$} 
            
            \label{fig:SW3L}

        \end{subfigure}

\caption{The networks are better seen zoomed on the digital version of this \newline document. Visualization of communities in different layers of Episode VI - Return \newline of the Jedi (1983)~\cite{starwars1983episode}. The size of each node corresponds to its degree. (a) The \newline character layer $G_{CC}$. (b) The keyword layer $G_{KK}$. (c) The location layer $G_{LL}$.}
\label{fig:communities}
\end{figure}

\begin{figure}[t!]
\centering
        \begin{subfigure}[b]{0.49\textwidth}   
            \centering 
            \includegraphics[width=\textwidth]{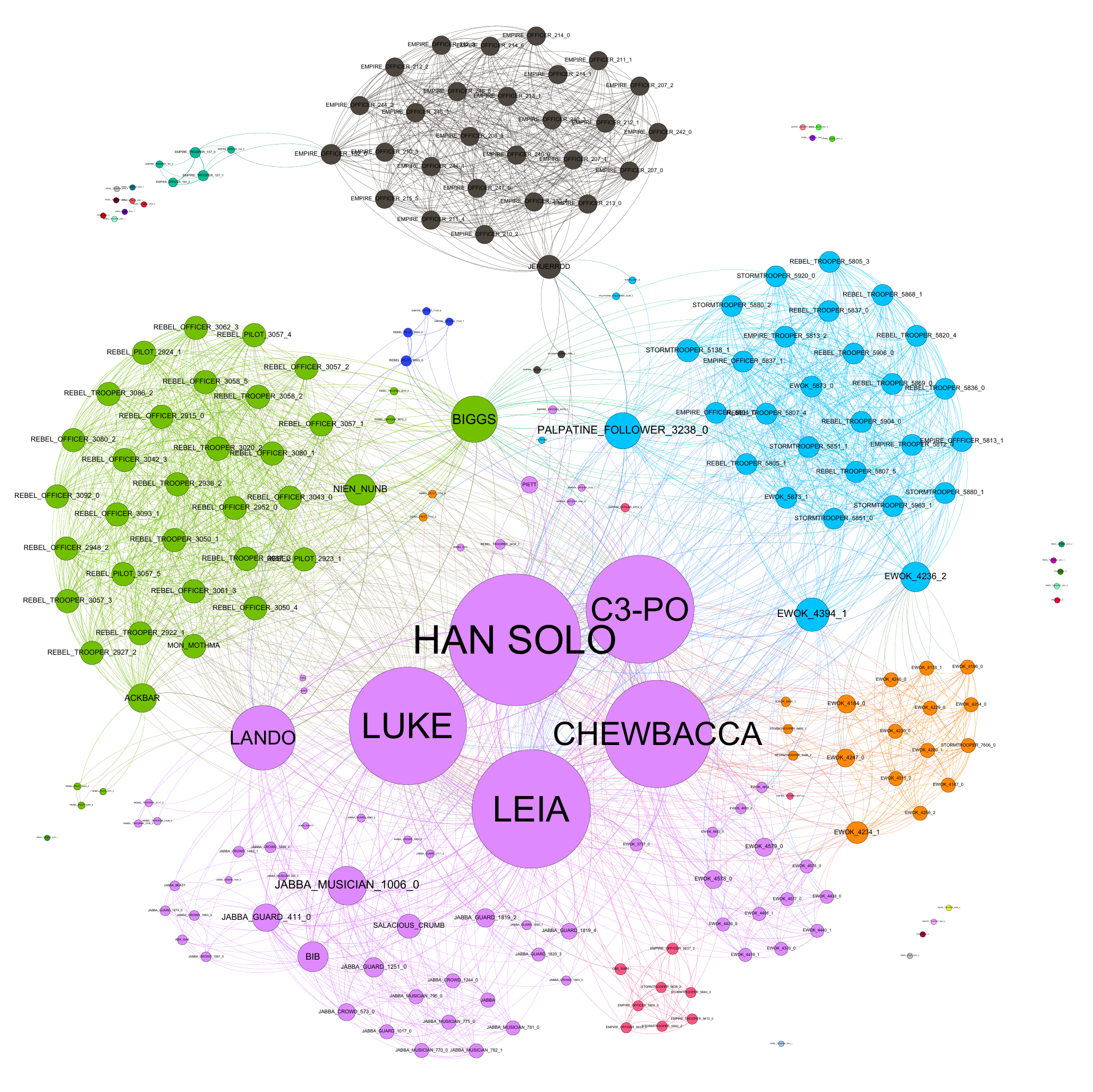}
            \caption[  $G_{FF}$]{$G_{FF}$}
            
            \label{fig:SW3F}
        \end{subfigure}
        \vskip\baselineskip  
        \begin{subfigure}[b]{0.48\textwidth}   
            \centering 
            \includegraphics[width=\textwidth]{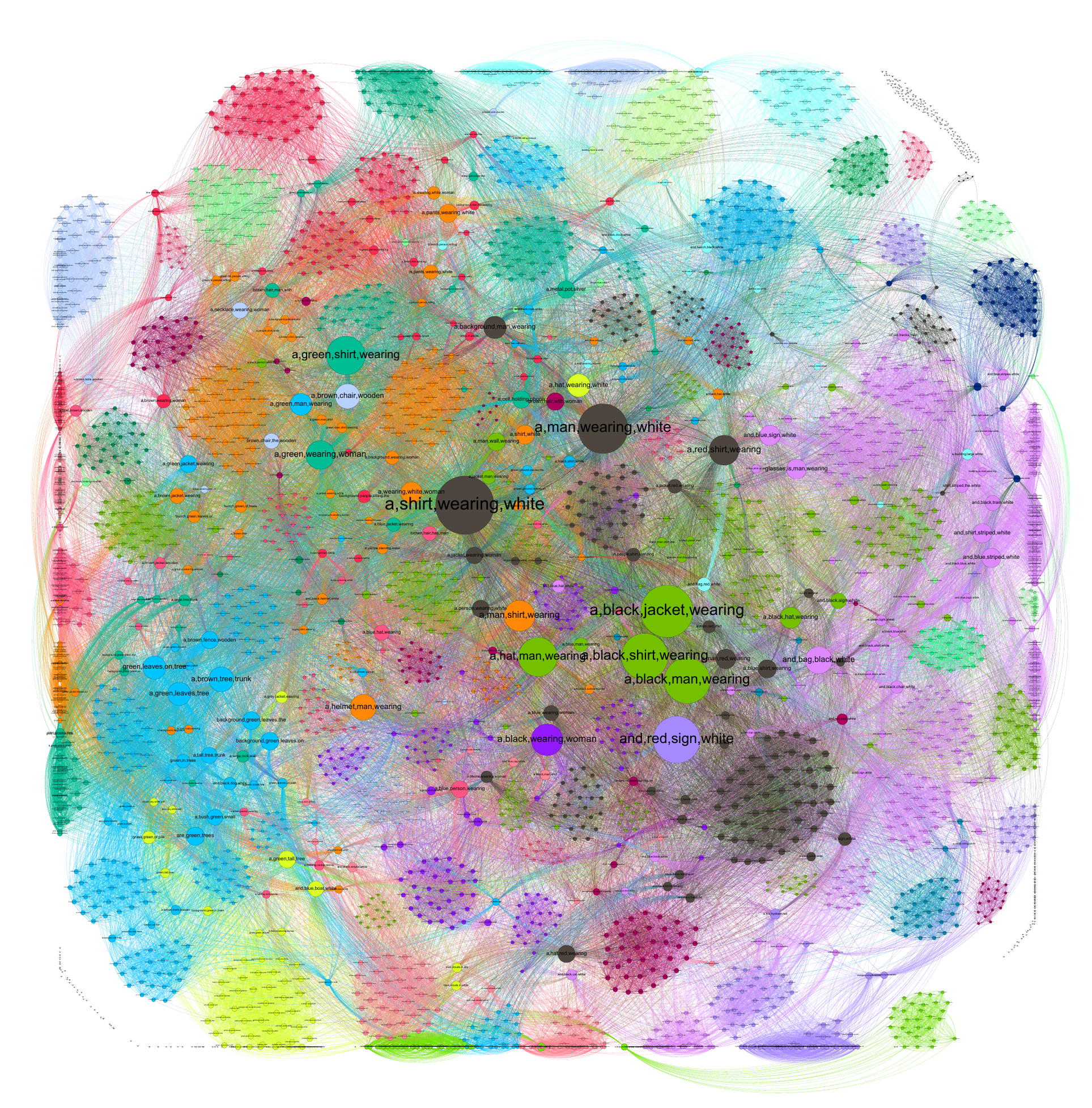}
            \caption[  $G_{CaCa}$]{$G_{CaCa}$}
            
            \label{fig:SW3CA}
        \end{subfigure}
        \qquad 
        \begin{subfigure}[b]{0.48\textwidth}   
            \centering 
            \includegraphics[width=\textwidth]{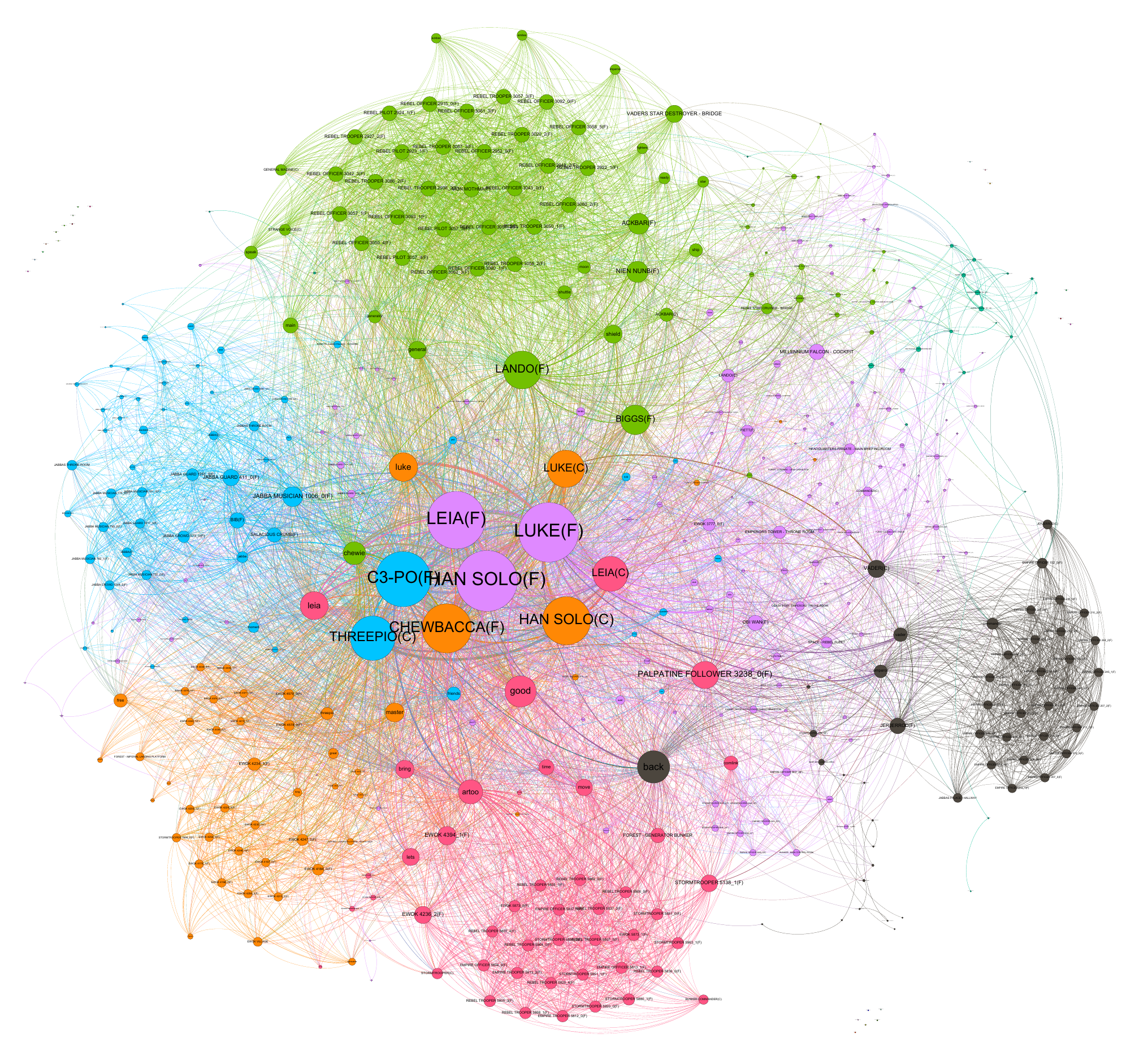}
            \caption[  $\mathbb G'$] {$\mathbb G'$}
            
            \label{fig:SW3M}
        \end{subfigure}

\caption{The networks are better seen zoomed on the digital version of this \newline document. Visualization of communities in different layers of Episode VI - Return \newline of the Jedi (1983)~\cite{starwars1983episode}. The size of each node corresponds to its degree. (a) The face \newline layer $G_{FF}$. (b) The caption layer $G_{CaCa}$. (c) The multilayer without captions $\mathbb G'$, with \newline the node label encoding: CHARACTER(C), FACE(F), keyword,  and LOCATION-.}
\label{fig:communities}
\end{figure}

\begin{table}[ht]
\begin{adjustbox}{max width=1.2\textwidth}
\begin{tabular}{c cccccccc}
\toprule
\multirow{2}{*}{} & \multicolumn{8}{c}{\textbf{CHARACTERS} $G_{CC}$}                                                                                                                                                     \\  
                         & \multicolumn{2}{c}{INFLUENCE SCORE} & \multicolumn{2}{c}{BETWEENNESS CENTRALITY} & \multicolumn{2}{c}{DEGREE} & \multicolumn{2}{c}{EIGEN CENTRALITY}\\ \midrule
\multirow{10}{*}{SW1}    & Q.GON     & 1,00  & Q.GON          & 365,29 & Q.GON     & 188   & Q.GON     & 1,00  \\ 
                         & ANAKIN    & 2,33  & J.JAR          & 262,28 & ANAKIN    & 151   & ANAKIN    & 0,77  \\  
                         & J.JAR     & 2,67  & ANAKIN         & 256,84 & J.JAR     & 131   & J.JAR     & 0,74  \\ 
                         & O.WAN     & 4,67  & AMIDALA        & 243,94 & PADME     & 110   & O.WAN     & 0,70  \\ 
                         & PADME     & 5,00  & O.WAN          & 145,97 & O.WAN     & 109   & PADME     & 0,63  \\ 
                         & AMIDALA   & 5,67  & PADME          & 124,22 & AMIDALA   & 82    & PANAKA    & 0,46  \\
\bottomrule
\end{tabular}

\end{adjustbox}

\captionof{table}{Top 10 nodes sorted and their different metrics of the character layer $G_{CC}$ for each of the 6 SW movies. }

\label{tab:character2}
\end{table}

\begin{table}[ht]
\begin{adjustbox}{max width=1.2\textwidth}

\begin{tabular}{c cccccccc}
\toprule
\multirow{2}{*}{} & \multicolumn{8}{c}{\textbf{KEYWORDS} $G_{KK}$}                                                                                                                                                     \\  
                         & \multicolumn{2}{c}{INFLUENCE SCORE} & \multicolumn{2}{c}{BETWEENNESS CENTRALITY} & \multicolumn{2}{c}{DEGREE} & \multicolumn{2}{c}{EIGEN CENTRALITY}\\ \midrule
\multirow{10}{*}{SW1}    & federation & 2,00  & jedi       & 1031,44 & federation & 40    & queen      & 1,00  \\  
                         & jedi       & 3,00  & federation & 780,77  & queen      & 40    & senate     & 0,89  \\  
                         & queen      & 3,67  & master     & 663,90  & jedi       & 39    & federation & 0,86  \\  
                         & senate     & 5,33  & naboo      & 608,94  & people     & 35    & people     & 0,81  \\  
                         & time       & 6,33  & time       & 591,96  & senate     & 34    & jedi       & 0,74  \\  
                         & people     & 6,33  & anakin     & 543,92  & chancellor & 32    & chancellor & 0,73  \\
\bottomrule
\end{tabular}
\end{adjustbox}
\captionof{table}{Top 10 nodes sorted and their different metrics of the keyword layer $G_{KK}$ for each of the 6 SW movies. }
\label{tab:keyword2}
\end{table}

\begin{table}[ht]
\begin{adjustbox}{max width=1.2\textwidth}
\begin{tabular}{c cccccccc}
\toprule
\multirow{2}{*}{} & \multicolumn{8}{c}{\textbf{LOCATIONS} $G_{FF}$}                                                                                                                                                     \\  
                         & \multicolumn{2}{c}{INFLUENCE SCORE} & \multicolumn{2}{c}{BETWEENNESS CENTRALITY} & \multicolumn{2}{c}{DEGREE} & \multicolumn{2}{c}{EIGEN CENTRALITY}\\ \midrule
\multirow{10}{*}{SW1}    & FBB      & 1,00  & FBB     & 3250,47 & FBB      & 14    & FBB      & 1,00  \\  
                         & NGP      & 4,00  & FBCR    & 2898,56 & TCH      & 10    & FBHOB    & 0,69  \\  
                         & TCH      & 4,67  & NSQC    & 1656,26 & NGP      & 8     & TCH      & 0,48  \\  
                         & TDNS     & 6,67  & TDNS    & 1649,20 & TDNS     & 8     & NGP      & 0,38  \\  
                         & FBCR     & 9,33  & NGP     & 1448,21 & AHMR     & 8     & SCU      & 0,38  \\  
                         & AHMR     & 11,33 & NSMA    & 1421,79 & NSCS     & 6     & FBH      & 0,34  \\
\bottomrule
\end{tabular}

\end{adjustbox}
\captionof{table}{Top 10 nodes sorted and their different metrics of the location layer $G_{LL}$ for each of the 6 SW movies. }

\label{tab:location2}
\end{table}

\begin{table}[ht]
\begin{adjustbox}{max width=1.2\textwidth}

\begin{tabular}{ c cccccccc }
\toprule
\multirow{2}{*}{} & \multicolumn{8}{c}{\textbf{FACES} $G_{FF}$}                                                                                                                                                     \\  
                         & \multicolumn{2}{c}{INFLUENCE SCORE} & \multicolumn{2}{c}{BETWEENNESS CENTRALITY} & \multicolumn{2}{c}{DEGREE} & \multicolumn{2}{c}{EIGEN CENTRALITY}\\ \midrule
                         
\multirow{10}{*}{SW1}    & Q.GON          & 1,00  & Q.GON          & 6911,14 & Q.GON          & 561   & Q.GON          & 1,00  \\  
                         & A.DOPPELGANGER & 2,00  & A.DOPPELGANGER & 4414,32 & A.DOPPELGANGER & 465   & A.DOPPELGANGER & 0,87  \\  
                         & ANAKIN         & 3,00  & ANAKIN         & 4020,27 & ANAKIN         & 442   & ANAKIN         & 0,79  \\  
                         & J.JAR          & 4,00  & J.JAR          & 3519,72 & J.JAR          & 424   & J.JAR          & 0,77  \\  
                         & O.WAN          & 5,00  & O.WAN          & 3097,53 & O.WAN          & 372   & O.WAN          & 0,71  \\  
                         & PANAKA         & 6,33  & SHMI           & 770,78  & PANAKA         & 244   & PANAKA         & 0,49  \\
\bottomrule
\end{tabular}

\end{adjustbox}

\captionof{table}{Top 10 nodes sorted and their different metrics of the face layer $G_{FF}$ for each of the 6 SW movies. }
\label{tab:face2}

\end{table}

\begin{table}[ht]
\begin{adjustbox}{max width=1.35\textwidth}
\centering

\begin{tabular}{ c cccccccc }
\toprule
\multirow{2}{*}{} & \multicolumn{8}{c}{\textbf{CAPTIONS} $G_{CaCa}$}                                                                                                                                                     \\  
                         & \multicolumn{2}{c}{INFLUENCE SCORE} & \multicolumn{2}{c}{BETWEENNESS CENTRALITY} & \multicolumn{2}{c}{DEGREE} & \multicolumn{2}{c}{EIGEN CENTRALITY}\\ \midrule
\multirow{10}{*}{SW1}   &  a,black,jacket,wearing & 1,67  & a,black,wearing,woman  & 606832,93  & a,black,jacket,wearing & 1244  & a,black,jacket,wearing & 1,00  \\
        & a,black,wearing,woman  & 2,00  & a,shirt,wearing,white  & 582836,63  & a,black,wearing,woman  & 1198  & a,red,wearing,woman    & 0,96  \\
        & a,red,wearing,woman    & 3,33  & a,black,jacket,wearing & 526258,21  & a,red,wearing,woman    & 1172  & a,black,wearing,woman  & 0,94  \\
        & a,red,shirt,wearing    & 4,33  & a,red,shirt,wearing    & 464385,84  & a,red,shirt,wearing    & 1077  & a,jacket,red,wearing   & 0,85  \\
        & a,shirt,wearing,white  & 4,67  & a,red,wearing,woman    & 407224,11  & a,shirt,wearing,white  & 1026  & a,red,shirt,wearing    & 0,83  \\
        & a,jacket,red,wearing   & 6,67  & a,brown,shirt,wearing  & 375031,25  & a,jacket,red,wearing   & 1004  & a,black,man,wearing    & 0,78  \\
\bottomrule
\end{tabular}
\end{adjustbox}
\captionof{table}{Top 10 nodes sorted and their different metrics of the caption layer $G_{CaCa}$ for each of the 6 SW movies. }

\label{tab:caption2}
\end{table}

\begin{table}[ht]
\begin{adjustbox}{max width=1.3\textwidth}
\begin{tabular}{c cccccccccccc}
\toprule
\multirow{2}{*}{} & \multicolumn{12}{c}{\textbf{MULTILAYER, ALL LAYERS} $\mathbb G$}                                                                                                                                                     \\  
                         & \multicolumn{3}{c}{INFLUENCE SCORE} & \multicolumn{3}{c}{BETWEENNESS CENTRALITY} & \multicolumn{3}{c}{DEGREE} & \multicolumn{3}{c}{EIGEN CENTRALITY}\\ \midrule
\multirow{10}{*}{SW1}     & Q.GON&$G_{FF}$                   & 1,00  & Q.GON&$G_{FF}$                   & 2029399,14 & Q.GON&$G_{FF}$                   & 5200  & Q.GON&$G_{FF}$                   & 1,00  \\
        & A.DOPPELGANGER&$G_{FF}$          & 2,33  & ANAKIN&$G_{FF}$                  & 1086600,09 & A.DOPPELGANGER&$G_{FF}$          & 4036  & A.DOPPELGANGER&$G_{FF}$          & 0,83  \\
        & ANAKIN&$G_{FF}$                  & 2,67  & A.DOPPELGANGER&$G_{FF}$          & 1020904,25 & ANAKIN&$G_{FF}$                  & 3602  & ANAKIN&$G_{FF}$                  & 0,72  \\
        & O.WAN&$G_{FF}$                   & 4,00  & O.WAN&$G_{FF}$                   & 627969,35  & O.WAN&$G_{FF}$                   & 3345  & O.WAN&$G_{FF}$                   & 0,71  \\
        & Q.GON&$G_{CC}$                   & 5,33  & Q.GON&$G_{CC}$                   & 619019,91  & Q.GON&$G_{CC}$                   & 3153  & J.JAR&$G_{FF}$                   & 0,66  \\
        & J.JAR&$G_{FF}$                   & 5,67  & J.JAR&$G_{FF}$                   & 588256,19  & J.JAR&$G_{FF}$                   & 3102  & Q.GON&$G_{CC}$                   & 0,64  \\
\bottomrule
\end{tabular}
\end{adjustbox}
\captionof{table}{Top 10 nodes sorted and their different metrics of the multilayer network with all layers $\mathbb G$ for each of the 6 SW movies.}

\label{tab:multilayer2}

\end{table}

\begin{table}[ht]
\begin{adjustbox}{max width=1.3\textwidth}
\begin{tabular}{c cccccccccccc}
\multirow{2}{*}{} & \multicolumn{12}{c}{\textbf{MULTILAYER, WITHOUT CAPTIONS} $\mathbb G'$}                                                                                                                                                     \\  
                         & \multicolumn{3}{c}{INFLUENCE SCORE} & \multicolumn{3}{c}{BETWEENNESS CENTRALITY} & \multicolumn{3}{c}{DEGREE} & \multicolumn{3}{c}{EIGEN CENTRALITY}\\ \midrule
\multirow{10}{*}{SW1}     & Q.GON&$G_{FF}$          & 1,00  & Q.GON&$G_{FF}$                  & 22399,29 & Q.GON&$G_{FF}$          & 463   & Q.GON&$G_{FF}$          & 1,00     \\
                         & A.DOPPELGANGER&$G_{FF}$ & 2,33  & A.DOPPELGANGER&$G_{FF}$         & 18254,37 & A.DOPPELGANGER&$G_{FF}$ & 412   & ANAKIN&$G_{FF}$         & 0,92     \\
                         & ANAKIN&$G_{FF}$         & 2,67  & ANAKIN&$G_{FF}$                 & 16702,12 & ANAKIN&$G_{FF}$         & 404   & A.DOPPELGANGER&$G_{FF}$ & 0,90     \\
                         & J.JAR&$G_{FF}$          & 4,33  & O.WAN&$G_{FF}$                  & 11760,71 & J.JAR&$G_{FF}$          & 370   & J.JAR&$G_{FF}$          & 0,89     \\
                         & O.WAN&$G_{FF}$          & 4,67  & J.JAR&$G_{FF}$                  & 11071,31 & O.WAN&$G_{FF}$          & 360   & O.WAN&$G_{FF}$          & 0,84     \\
                         & Q.GON&$G_{CC}$     & 6,00  & Q.GON&$G_{CC}$             & 8245,34  & Q.GON&$G_{CC}$     & 311   & Q.GON&$G_{CC}$     & 0,76     \\
\bottomrule
\end{tabular}
\end{adjustbox}
\captionof{table}{Top 10 nodes sorted and their different metrics of the multilayer network without the caption layer $\mathbb G'$ for each of the 6 SW movies. }

\label{tab:multilayerbis2}

\end{table}

\begin{table}[ht]

\begin{adjustbox}{max width=\textwidth}

\begin{tabular}{l l}
\toprule
\textbf{ABBREVIATION} & \textbf{NAME}          \\  \midrule
A.DOPPELGANGER                         & AMIDALA DOPPELGANGER 1         \\
B.B.ORGANA                             & BREHA BAIL ORGANA              \\
B.VOICE                                & BENS VOICE                     \\
B.FETT                                 & BOBA FETT                      \\
B.ORGANA                               & BAIL ORGANA                    \\
C.C.CODY                               & CLONE COMMANDER CODY           \\
C.CROWD 1                              & CORUSCANT CROWD 1              \\
\bottomrule
\end{tabular}
\end{adjustbox}
\captionof{table}{Abbreviation table for character and face names.}

\end{table}

\begin{table}[ht]

\begin{adjustbox}{max width=0.7\textwidth}

\begin{tabular}{l l l}

\toprule
\textbf{MOVIE}                 & \textbf{ABBREVIATION} & \textbf{NAME}                                                     \\ \midrule
\multirow{18}{*}{SW1} & AHMR         & ANAKINS HOVEL - MAIN ROOM                                \\ 
                      & FBB          & FEDERATION BATTLESHIP - BRIDGE                           \\  
                      & FBCR         & FEDERATION BATTLESHIP - CONFERENCE ROOM                  \\
                      & FBH          & FEDERATION BATTLESHIP - HALLWAY                          \\ 
                      & FBHOB        & FEDERATION BATTLESHIP - HALLWAY - OUTSIDE BRIDGE-        \\ 
                      & FBMB         & FEDERATION BATTLESHIP - MAIN BAY                         \\  
                      & MSAVP        & MOS ESPA - ARENA - VIEWING PLATFORM                      \\
\bottomrule
\end{tabular}

\end{adjustbox}

\captionof{table}{Abbreviation table for location names.}
\end{table}

\end{document}